%% file: main.tex
\documentclass[twoside,11pt]{article}
\usepackage[preprint]{jmlr2e}
\usepackage{microtype,booktabs,xurl,array,adjustbox}
\usepackage{amsmath,amssymb,mathtools}
\usepackage{tikz}
\newif\ifjmlrlayout\jmlrlayouttrue

\providecommand{\tightlist}{\setlength{\itemsep}{0pt}\setlength{\parskip}{0pt}}

\ShortHeadings{Hessian-Similar Optimization with Capped Participation}{Rui Wang}
\firstpageno{1}

\begin{document}

\title{Hessian-Similar Optimization with Capped Participation:\\
Lower Bounds and Virtual Masters}

\author{\name Rui Wang \email wangrui@microsoft.com\\
       \addr Microsoft, Vancouver, BC, Canada}

\maketitle
\thispagestyle{plain}

\begin{abstract}
\input{abstract}
\end{abstract}


\input{body}

\clearpage
\appendix
\numberwithin{equation}{section}
\input{appendix}

\vskip 0.2in
\bibliography{references}

\end{document}

%% file: abstract.tex
What is the round complexity of Hessian-similar optimization when each round
can contact at most \(s\) of \(m\) clients? Prior rates permit a free master
or full participation and do not characterize strict caps. We count every
one-vector query and reply. A hidden-client chain gives a lower bound for
zero-respecting local algorithms with adaptive cohorts and random stopping:
on strongly convex quadratics, the required rounds are
\(\Omega((m/s)\sqrt{\min\{\delta_a,\bar\delta/\sqrt m\}/\mu}\,
\log(1/\epsilon))\) for \(s\leq(m-2)/2\) in the high-accuracy,
high-dissimilarity regime. Here \(\mu\) is strong convexity, while
\(\delta_a\) and \(\bar\delta\) bound anchor and average Hessian variation.
To remove dependence on a poor anchor, virtual-master SVRS uses independent
estimator and proximal cohorts and averages the proximal replies. With Catalyst,
it reaches relative gap \(\epsilon\) in
\(\tilde O(m/s+(\sqrt{m/s}+m^{3/4}/s)\sqrt{\bar\delta/\mu})\) expected
rounds on quadratics with strongly convex clients, independent of
\(\delta_a\). Combining this method with charged analyses of existing
algorithms matches the lower bound's polynomial dependence for
\(s\leq\sqrt m\) and \(s=\Omega(m)\); the worst gap is \(O(m^{1/8})\).
A held-out 12-seed synthetic study finds fewer rounds than tuned SVRP in nine
of ten dip-family incomplete-cohort cells, versus two of 30 other cells.
The lower bounds concern zero-respecting algorithms and do not cover
unrestricted messages.

%% file: body.tex
\section{Introduction}\label{introduction}

\begin{figure*}[!t]
\centering
\includegraphics[width=0.98\textwidth]{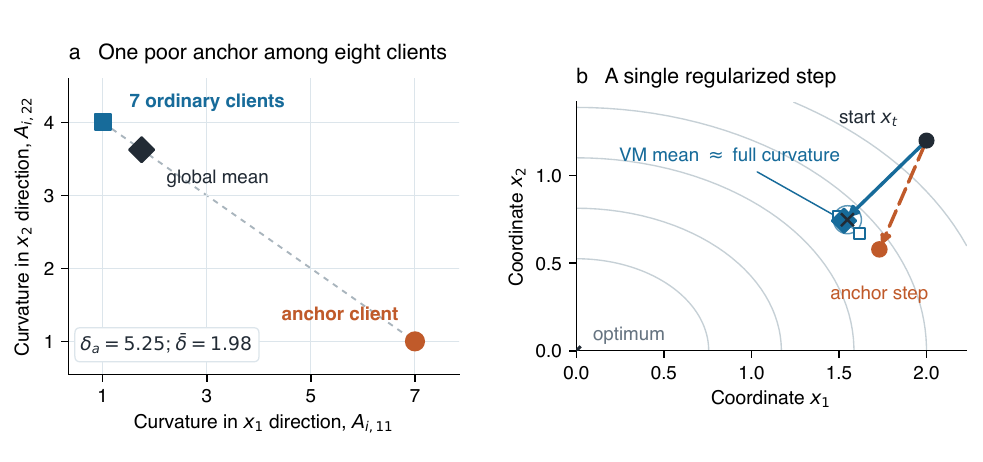}
\caption{One atypical anchor can deviate much more from global
curvature than the population-average second-order scale. (a) For eight quadratic clients
\(f_i(x)=\frac12x^\top A_i x\), seven have
\(A_i=\operatorname{diag}(1,4)\), one designated anchor has
\(A_a=\operatorname{diag}(7,1)\), and the mean is
\(A=\operatorname{diag}(1.75,3.625)\). Thus
\(\delta_a=5.25=\sqrt7\,\bar\delta\). (b) From
\(x_t=(2,1.2)\), with exact global gradient and \(\lambda=6\), the
anchor step uses \((A_a+\lambda I)^{-1}\), while the size-two
virtual-master step averages sampled client inverses. Its expected
endpoint lies near the full-curvature reference using
\((A+\lambda I)^{-1}\); hollow squares mark the two possible cohort
endpoints. This single-step illustration is not a convergence or
runtime comparison.}
\label{fig:quadratic-geometry}
\end{figure*}

Second-order similarity can reduce communication in distributed
optimization, but published round counts use different units. A
full-gradient refresh, a proximal reply, and a multistage exchange may
each count as one round. Here each synchronous round contacts at most
\(s\) of \(m\) clients. The server sends one \(d\)-vector to each
contacted client and receives one \(d\)-vector in return. We count
rounds, contacts, and transmitted vectors separately. The server holds
no client data, every contact is charged, and local computation is
unpriced.
For example, with eight clients and a two-client cap, collecting a full
gradient takes four rounds. An anchor proximal query that uses this
gradient occupies a fifth round because its input depends on the fourth
round's replies.

One designated client satisfies an anchor similarity bound
\(\delta_a\). From Section~\ref{two-parameter-main} onward, we also
use the average bound \(\bar\delta\) of \citet{lin2023}. Methods whose
parameters depend on these bounds receive certified values as public
inputs. We ask how
the required rounds depend on \(m/s\), \(\delta_a\),
\(\bar\delta\), \(\mu\), and \(\epsilon\). Existing upper bounds use an
uncharged master or occasional full participation
\citep{khaled2023,lin2023,jiang2024}. The lower bound of
\citet{lin2023} samples one client uniformly at each step and leaves
the master uncharged. It does not cover adaptive cohort selection
under a participation cap.

Figure~\ref{fig:quadratic-geometry} shows the distinction between a
designated client's curvature and the population average. In its
eight-client quadratic example, the anchor deviation is
\(\sqrt7\) times the average second-order deviation. A cohort-averaged
proximal step then lies close to the full-curvature reference for the
shown starting point, while the anchor step takes a different direction.
The figure illustrates one update; the round guarantees below require
the separate analysis.

\textbf{Results.}
\begin{itemize}\tightlist
\item \emph{Lower bounds for adaptive cohorts (main results).} In a
ZR local oracle, any algorithm with adaptive cohorts and random
stopping needs
\(\Omega\bigl((m/s)\sqrt{\min\{\delta_a,\bar\delta/\sqrt m\}/\mu}\)
\(\log(1/\epsilon)\bigr)\)
expected rounds for \(s\leq(m-2)/2\) (Theorem~7, Corollary~11) and
\(\Omega\bigl(\sqrt{\min\{\delta_a,\bar\delta\}/\mu}\)
\(\log(1/\epsilon)\bigr)\) for
every \(s\leq m\) (Corollary~12). The key step spreads a chain's
links over hidden clients so that no client holds two adjacent links.
This keeps \(\bar\delta\) at \(O(\delta\sqrt m)\) or \(O(\delta)\) and
places the hard instances inside the two-parameter class.
\item \emph{Charged upper bounds.} A preconditioned Chebyshev method
(Theorem~8), cohort SVRS and AccSVRS with the anchor charged as a
client (Theorems~9--10), and batched Acc-S-DANE (Proposition~13) are ZR
on quadratics. Theorems~9--10 adapt the proofs of \citet{lin2023}, and
Proposition~13 is an accounting corollary of \citet{jiang2024}.
\item \emph{A new anchor-free algorithm.} Virtual-master cohort SVRS
(VM-SVRS) replaces the anchor's proximal reply with the average of a
fresh cohort's proximal replies. With Catalyst it needs
\(\tilde O\bigl(m/s+(\sqrt{m/s}+m^{3/4}/s)\sqrt{\bar\delta/\mu}\bigr)\)
rounds on quadratics with strongly convex clients, independently of
\(\delta_a\) (Theorem~14, Corollary~15).
\item \emph{Complexity map.} The polynomial parts of the bounds match
when \(\delta_a\leq\sqrt m\,\bar\delta/s\), and also when
\(s\leq\sqrt m\) or \(s=\Omega(m)\) (Table~\ref{tab:regimes}).
Corollary~15 additionally pays \(\ell(\ell+\log(1/\epsilon))\), where
\(\ell=\log(mL_{\max}/\mu)\), while the lower bound pays
\(\log(1/\epsilon)\). In
the remaining poor-anchor range the gap is
\(O(1+\min\{\sqrt s/m^{1/4},\sqrt{m/s}\})\leq O(m^{1/8})\); without
VM-SVRS it is up to \(m^{1/4}\).
\item \emph{Boundaries.} The ZR restriction is essential
(Proposition~5). The anchor bound alone cannot control a fresh sampled
estimator (Theorem~3). Hidden clients force \(\Omega(m/s)\) rounds for
every algorithm (Theorem~2).
\item \emph{Experiments.} A 16-seed GPU study on exact quadratics tests
the predicted cohort-size scaling. A separate 12-seed study
compares VM-SVRS with tuned baselines, selecting parameters on four
seeds and evaluating on eight fresh paired seeds.
\end{itemize}

\section{Related work and scope}\label{prior-work-and-scope}

\paragraph{Similarity-based methods.}
Approximate Newton and statistically preconditioned methods exploit
related local curvature \citep{shamir2014dane,hendrikx2020statistically}.
\citet{khaled2023} analyze SVRP under average squared gradient-difference
similarity, with occasional full refreshes. \citet{lin2023} prove
SVRS and AccSVRS rates under average second-order similarity with a free
master; their lower bound samples one client per step.
\citet{kovalev2022} use gradient sliding with full participation, and
SVOGS \citep{zhou2024svogs} samples client mini-batches for minimax
problems. S-DANE and Acc-S-DANE \citep{jiang2024} stabilize local
proximal steps; their partial-participation analysis assumes subset
similarity and bounded gradient variance, and their full-participation
accelerated rate is what Proposition~13 batches. SPAM
\citep{karagulyan2024spam} handles cohorts of clients with per-client
Hessian similarity, bounded gradient variance, inexact proximal steps,
and nonconvex losses. ASEG \citep{bylinkin2024aseg} is an accelerated
client-sampled extragradient method for strongly convex problems that
mixes Hessian and gradient similarity, with a neighborhood term set by
the gradient dissimilarity. These results use different assumptions;
none gives a lower bound for adaptive cohorts. Corollary~15 uses the
outer loop of Catalyst \citep{lin2018catalyst} with our round
accounting.
Averaging local proximal or Newton-type solutions is the core of DANE
and FedDANE \citep{li2019feddane}, and AIDE \citep{reddi2016aide} wraps
inexact DANE in Catalyst, as Corollary~15 does for VM-SVRS. DANE and AIDE
average over all clients in every exchange. FedDANE already samples
one device set for the gradient and another for the local solves, but
its gradient is a cohort estimate at the current point, not a
variance-reduced snapshot correction. VM-SVRS uses independent cohorts
for the snapshot correction and local solves, with SVRS variance
reduction. We analyze the cohort average as a random preconditioner;
this yields a rate independent of \(\delta_a\).

\paragraph{Lower bounds and oracles.}
\citet{arjevani2015communication} prove communication lower bounds
for \(\delta\)-related functions, and \citet{scaman2017optimal} for
networks. Their oracle and participation rules differ from the capped
one-vector protocol. A zero-respecting restriction makes our lower
bound possible: a dense probe can activate several chain links, while
an unrestricted vector reply can encode a local objective. The bound
therefore does not cover the unrestricted messages allowed by S-DANE
or SPAM.

\paragraph{Participation and variance reduction.}
FedAvg \citep{mcmahan2017fedavg}, ProxSkip
\citep{mishchenko2022proxskip}, and SCAFFOLD
\citep{karimireddy2020scaffold} change exchange frequency or control
drift. SVRG \citep{johnson2013svrg}, SAGA \citep{defazio2014saga}, and
loopless variance reduction \citep{kovalev2020loopless} rely on
reference gradients or stored client tables. Our sampled-estimator
obstruction (Theorem~3) concerns one estimator, not these methods.

\begin{table*}[t]
\centering\footnotesize
\caption{Assumptions and accounting for nearby analyses. This compares
hypotheses and accounting units, not theorem rates. ``Full'' means all
clients may join a nominal exchange; under our cap a full pass is
batched.}
\label{tab:scope-main}
\setlength{\tabcolsep}{4pt}
\newcolumntype{P}[1]{>{\raggedright\arraybackslash}p{#1}}
\begin{tabular}{P{0.14\textwidth}P{0.29\textwidth}P{0.15\textwidth}P{0.31\textwidth}}
\toprule
Work & Similarity information & Participation & Relevant accounting distinction\\
\midrule
SVRP & Average squared differences & Full refreshes & Refresh cost batched under cap\\
SVRS/\hspace{0pt}AccSVRS & Average second-order similarity & One client and full refreshes & Master uncharged; lower bound for one sampled client\\
Gradient sliding & Componentwise similarity & Full & Full exchange exceeds cap\\
S-DANE/Acc-S-DANE & Subset similarity and gradient variance & Partial or full & Multistage transcript spans rounds\\
SPAM & Per-client Hessian similarity, gradient variance & Cohorts & Nonconvex stationarity; inexact prox\\
ASEG & Hessian and gradient similarity & Sampled clients & Neighborhood set by gradient dissimilarity\\
This paper & Anchor \(\delta_a\) and average \(\bar\delta\) & At most \(s\) & Every query and reply charged; ZR lower bounds for adaptive cohorts\\
\bottomrule
\end{tabular}
\end{table*}

Appendix~\ref{app:oracle} gives the cost of every operation used below
and transcribes an S-DANE iteration into three strict rounds.
Appendix~\ref{app:deferred} reports a charged S-DANE execution.

\section{Strict participation model and
accounting}\label{strict-participation-model-and-accounting}

Let

\[f(x)=\frac1m\sum_{i=1}^m f_i(x),\qquad x\in\mathbb R^d ,\]

where each differentiable \(f_i\) is \(\mu\)-strongly convex. Fix an
anchor client \(a\). Assume \(f\) and \(f_a\) are twice differentiable
and

\[\sup_x\|\nabla^2 f_a(x)-\nabla^2 f(x)\|_{\mathrm{op}}\leq\delta .
\tag{1}\]

There is no bound on the local gradient offsets or on the average
squared deviation of the local Hessians. The server initially knows
\(x_0,m,s,\mu,\delta\) and the anchor identity, but has no stored client
data. It chooses the clients and query vector in each strict round. We
count a query and its reply as two vector transmissions per contact.
Local arithmetic and numerical precision are not priced by this oracle.

Write \(r=f-f_a\). Condition (1) makes \(\nabla r\)
\(\delta\)-Lipschitz, although \(r\) need not be convex. For a parameter
\(\lambda\geq\delta\), \(\lambda>0\), define

\[Q_y(x)= f_a(x)+r(y)+\langle\nabla r(y),x-y\rangle+
              \frac{\lambda}{2}\|x-y\|^2 .
\tag{2}\]

\textbf{Batched anchored proximal gradient (BAPG).} At \(y=x_t\),
partition all \(m\) clients into \(\lceil m/s\rceil\) capped batches and
obtain their gradients at \(y\). The anchor gradient from its batch
gives \(\nabla r(y)=\nabla f(y)-\nabla f_a(y)\). After all replies are
known, contact the anchor once more for

\[x_{t+1}=\arg\min_x Q_y(x)
       =\operatorname{prox}_{f_a/\lambda}
          \!\left(y-\frac{\nabla r(y)}{\lambda}\right).
\tag{3}\]

The proximal input cannot be formed before the final gradient batch, so
its anchor contact occupies a separate round. One step costs exactly
\(\lceil m/s\rceil+1\) rounds, \(m+1\) contacts, and \(2(m+1)\)
transmitted vectors.

\section{Baseline and barriers}\label{a-charged-baseline-and-its-proof}
\label{participation-and-sampling-barriers}

\textbf{Theorem 1 (BAPG).} Let \(x^\star=\arg\min f\) and
\(\Delta_t=f(x_t)-f(x^\star)\). Exact BAPG satisfies

\[\Delta_{t+1}\leq q\Delta_t,\qquad
q=\frac{\lambda+\delta}{\mu+\lambda+\delta}<1.
\tag{4}\]

Hence \(T=\lceil\log(\Delta_0/\varepsilon)/\log(1+\mu/(\lambda+\delta))\rceil\)
steps and \(R\leq T(\lceil m/s\rceil+1)\) rounds suffice (5). A local
solve with model error \(e_t\) adds \(\sum_tq^{T-1-t}e_t\) to the final
gap (6). With \(\lambda=\delta\), \(q=2\delta/(\mu+2\delta)\).

\textbf{Theorem 2 (hidden-client discovery).} For any \(m\geq2\),
\(1\leq s\leq m\), \(\mu>0\), and \(\theta>0\), every possibly
randomized \(R\)-round algorithm in this oracle has a one-dimensional
quadratic instance with \(\delta=0\) for which

\[\mathbb E[f(x_R)-f(x^\star)]
\geq (1-Rs/m)_+\frac{\mu\theta^2}{2}.
\tag{7}\]

The instance hides a large linear term at a uniformly random client.
The bound forces a participation term for any fixed relative reduction,
but no curvature term.

\textbf{Theorem 3 (unbounded variance at any prescribed anchor
dissimilarity).} Fix \(m\geq3\), \(1\leq s<m\), \(\mu>0\), and any
\(\delta\geq0\). For any \(M\geq\delta/2\), the one-dimensional
quadratic clients with zero linear terms and Hessians
\(A_0=\mu+M+\delta\), \(A_1=\mu\), \(A_2=\mu+2M-\delta\), and
\(A_i=\mu+M\) for \(i\geq3\) are \(\mu\)-strongly convex, have anchor
gap exactly \(\delta\), and for a uniform size-\(s\) cohort \(S\) satisfy
\[\begin{aligned}
&\mathbb E\Bigl|
\frac1s\sum_{i\in S}[\nabla f_i(x)-\nabla f_i(w)]
-[\nabla f(x)-\nabla f(w)]
\Bigr|^2\\
&=\frac{m-s}{s(m-1)}
 \frac{\delta^2+M^2+(M-\delta)^2}{m}|x-w|^2.
\end{aligned}
\tag{8}\]

So a fresh sampled correction cannot use the anchor \(\delta\) as a
variance proxy. This is an estimator obstruction, not a lower bound for
methods that learn client deviations. Proofs of Theorems~1--3 are in
Appendix~\ref{app:proofs}.

\textbf{Zero-similarity controls.} At \(\delta=0\), the residual
gradient \(\nabla r\) is constant. One capped full pass then lets the
anchor contract the gap by half per contact (Proposition~4), and one
pass plus one target-dependent prox reaches any relative reduction
\(\tau\) (Proposition~6). Together with Theorem~2, this gives a
\(\Theta(m/s)\) round law for a fixed relative reduction at
\(\delta=0\).

\textbf{Proposition 5 (quadratic identification).} If every client is
quadratic with \(A_i\succeq\mu I\), the server can compute the exact
minimizer in \((d+1)\lceil m/s\rceil\) rounds for any \(\delta\), by
probing complete gradients at \(x_0\) and \(x_0+e_j\). Hence no
accuracy-dependent lower bound holds for fixed-dimensional quadratics
in the unrestricted oracle. Appendix~\ref{app:controls-proof} proves
Propositions~4--6.

\section{A quadratic zero-respecting round law}
\label{positive-chain-main}
For the lower bounds, the server's initial linear memory is
\(\{0\}\). Every sent vector and final output must be in the span
of prior sent vectors and replies. Each client reply is a requested
local gradient, Hessian-vector product, shifted inverse, or proximal
point of its quadratic; arbitrary objective encodings are excluded.
Client selection and scalar coefficients remain adaptive. This
\emph{ZR local oracle} is a restriction of the strict one-vector
protocol. Appendix~\ref{app:zr-chain} gives the full construction and
proofs.

\ifjmlrlayout
\begin{figure*}[t]
\input{hidden_chain_figure}
\end{figure*}
\fi

\textbf{Theorem 7 (positive-\(\delta\) ZR lower bound).} Fix
\(\mu,\delta>0\), \(m\geq3\), and
\[
q=\frac{\sqrt{1+\delta/\mu}-1}{\sqrt{1+\delta/\mu}+1},\qquad
K=\left\lceil\frac{\log((1-q)^2/\epsilon)}
                         {2\log(1/q)}\right\rceil
\]
for \(0<\epsilon<(1-q)^2\). In dimension \(d\geq K+1\), a family
of locally \(\mu\)-strongly convex quadratics has anchor-to-average
Hessian deviation at most \(\delta\) and initial point zero, yet
every \(R<K\) round ZR algorithm has
\(\Delta_R>\epsilon\Delta_0\) on every member of this family. If
\(s\leq(m-2)/2\), \(K\geq2\), and
\(j_0=\lfloor(m-2)/(2s)\rfloor\), then for
\(R<1+(K-1)(j_0+1)/4\) some fixed member satisfies
\[
\Pr(\Delta_R>\epsilon\Delta_0)
\geq1-\exp(-(K-1)/8).
\tag{11}
\]
The probability is over the algorithm's coins. The construction
distributes nonadjacent links of a tridiagonal chain over clients;
each discovery requires contacting its still-hidden holder. In the
high-accuracy regime \(\epsilon\leq(1-q)^4\) with
\(\delta\geq3\mu\), (11) supplies the
\(m/s\) and \(\sqrt{\delta/\mu}\log(1/\epsilon)\) factors.

\textbf{Proof idea.}
\ifjmlrlayout
Figure~\ref{fig:hidden-chain}
\else
Figure~\ref{fig:hidden-chain} in Appendix~\ref{app:zr-chain}
\fi
shows the two
obstacles: the current link holder must be found, and each reply can
extend the visible support by at most one coordinate. The hard objective is
\(f(x)=\frac\mu2\|x\|^2+\frac\alpha2Q(x)-\alpha x_1\) with
\(\alpha=\delta/4\), where \(Q\) is a tridiagonal chain whose
minimizer \(x^\star_j=q^j\) decays geometrically. Every client keeps
\(\frac\mu2\|x\|^2\). The anchor holds the linear term, and link
\(\ell\), the term \((x_\ell-x_{\ell+1})^2\), is held by a random
non-anchor client \(h_\ell\ne h_{\ell-1}\). Since no client holds two
adjacent links, every ZR reply maps
\(V_J=\operatorname{span}(e_1,\ldots,e_J)\) into \(V_J\) unless it
comes from \(h_J\), and then only into \(V_{J+1}\). So the frontier
advances at most one coordinate per round, and only when the cohort
contains the still-hidden holder of the current link. When frontier
\(J\) is first reached, \(h_J\) is uniform over \(m-2\) clients given
the transcript. Hence the advance within \(j\) rounds has probability
at most \(js/(m-2)\), for any adaptive cohort. A Hoeffding bound over
the waits gives (11), and any point in \(V_J\) has relative gap at
least \((1-q)^2q^{2J}\). The anchor sees deviation
\(\alpha\|\nabla^2Q/2\|\leq\delta\). Spreading the links also keeps the
average similarity small (Lemma~C.6), which Section~\ref{two-parameter-main}
exploits.

\textbf{Theorem 8 (charged quadratic upper bound).} For any quadratic
family with the anchor condition and \(\delta>0\), put \(P=A_a+\delta I\) and
\(\rho=(\sqrt{1+2\delta/\mu}-1)/(\sqrt{1+2\delta/\mu}+1)\).
Preconditioned Chebyshev semi-iteration from \(x_0=0\) is ZR and satisfies
\[
 \Delta_k\leq4\rho^{2k}\Delta_0.
 \tag{12}
\]
Each step costs exactly \(\lceil m/s\rceil+1\) strict rounds,
\(m+1\) contacts, and \(2(m+1)\) vector transmissions: one capped
full-gradient pass and a subsequent anchor prox reply.

The proof uses \(\frac{\mu}{\mu+2\delta}P\preceq A\preceq P\), and the
anchor prox computes \(x-P^{-1}\nabla f(x)\). Theorems~7 and~8 agree
up to constants in the high-accuracy, high-dissimilarity ZR regime.
The span restriction is essential: the hard family's global objective
does not depend on its random client assignment, so an unrestricted
server that knows the family can output the optimum directly.

\section{Adding average similarity}
\label{two-parameter-main}
Theorem~3 shows that the anchor bound alone cannot control a sampled
estimator. We therefore add the \emph{average} second-order similarity
of \citet{lin2023}, keeping~(1) as a separate anchor bound
\(\delta_a:=\delta\):
\[
 \frac1m\sum_{i=1}^m\|\nabla(f_i-f)(x)-\nabla(f_i-f)(y)\|^2
 \leq\bar\delta^2\|x-y\|^2 .
 \tag{13}
\]
For quadratics, \(\bar\delta^2=\|\frac1m\sum_i(A_i-A)^2\|_{\rm op}\).
Condition~(13) implies~(1) with \(\delta_a=\sqrt{m-1}\,\bar\delta\),
which is tight (Lemma~C.1), so we take
\(\delta_a\leq\sqrt{m-1}\,\bar\delta\). Theorems~9--10 need only \(f\)
to be \(\mu\)-strongly convex; the lower bounds, Proposition~13, and
Theorem~14 use \(\mu\)-strongly convex clients. Appendices~\ref{app:cohort}
and~\ref{app:anchor-free} have all proofs.

\textbf{Cohort SVRS.} Fix a cohort size \(b\) and put \(E=\lceil m/b\rceil\),
\(p=1/E\), \(\xi=\frac{m-b}{b(m-1)}\), and
\[
 \lambda=4\max\{\mu,\ \delta_a,\ \bar\delta\sqrt{\xi E}\}
 \leq4\max\{\mu,\ \delta_a,\ \sqrt m\,\bar\delta/b\}.
 \tag{14}
\]
An epoch from \(w_0\) first collects \(\nabla f(w_0)\) in
\(\lceil m/s\rceil\) rounds. It then runs \(T\sim\mathrm{Geom}(p)\)
steps. Step \(t\geq1\) draws a uniform \(b\)-cohort \(S_t\) and forms
\(\hat g_t=\nabla f(w_0)+\frac1b\sum_{i\in S_t}
[\nabla f_i(x_t)-\nabla f_i(w_0)]\); step~\(0\) uses
\(\hat g_0=\nabla f(w_0)\). Every step contacts the anchor once for
\(x_{t+1}=\arg\min_x f_a(x)+\langle\hat g_t-\nabla f_a(x_t),x\rangle
+\frac\lambda2\|x-x_t\|^2\), which for a quadratic anchor is one
shifted-inverse reply. Inexact anchor solves are allowed. At
\(b=s=1\) with a free master, this is SVRS \citep[Algorithm~1]{lin2023}.

\textbf{Theorem 9 (cohort SVRS).} Let
\(\tilde\kappa=\mu E/\lambda\). Successive epochs contract an expected
potential, which dominates the gap and is at most
\((\frac54+\frac5{8\tilde\kappa})\Delta_0\) initially, by the factor
\(1-\min\{\frac12,\tilde\kappa/5\}\). An epoch costs
\(\lceil m/s\rceil+1+(E-1)(\lceil b/s\rceil+1)\) strict rounds in
expectation, or \(3\lceil m/s\rceil-1\) at \(b=s\). The total is
\(O((m/s+\lambda/\mu)\log(\Delta_0/\epsilon))\) expected rounds.

\textbf{Theorem 10 (cohort AccSVRS).} The accelerated outer loop of
\citet[Algorithm~2]{lin2023} wraps one cohort-SVRS epoch per step, with
\(\tau=\frac14\min\{1,\sqrt{\tilde\kappa}\}\) and a fresh \(b\)-cohort
for the gradient mapping. It reaches
\(\mathbb E[f(y_K)-f^\star]\leq\epsilon\) with
\(K=\lceil\max\{4,4/\sqrt{\tilde\kappa}\}\log(1.2\Delta_0/\epsilon)\rceil\)
outer steps. With \(b=s\), the expected number of strict rounds is
\[
 \bar R
 =O\Bigl(\Bigl[\frac ms+\sqrt{\frac ms}\sqrt{\frac{\delta_a}\mu}
 +\frac{m^{3/4}}s\sqrt{\frac{\bar\delta}\mu}\Bigr]
 \log\frac{\Delta_0}\epsilon\Bigr),
 \tag{15}
\]
with explicit constants in \eqref{eq:cohort-total}. At \(s=1\) and
\(\delta_a\leq\sqrt m\,\bar\delta\), (15) is the rate of
\citet{lin2023}. For quadratic clients, exact anchor solves, and
\(x_0=0\), the method is ZR. The proof follows Lin et al.'s Lemma~D.1,
Lemmas~3.1 and~3.5, and Theorems~3.3 and~3.6 step by step, with three
changes: cohort sampling contributes the finite-population factor
\(\xi\), the anchor deviation is the separate \(\delta_a\) rather than
\(\sqrt m\,\bar\delta\), and each cost is converted to strict rounds.

\textbf{Corollary 11 (two-parameter ZR lower bound).} Let
\(\delta=\min\{\delta_a,2\bar\delta/\sqrt{3m}\}\geq3\mu\),
\(s\leq(m-2)/2\), \(4\epsilon\leq(1-q)^4\) with \(q\) of Theorem~7 at
\(\delta\), \(K\geq7\) at accuracy \(4\epsilon\), and \(d\geq K+1\).
Consider any ZR algorithm with adaptive cohorts and a possibly random
number of rounds that guarantees
\(\mathbb E[f(\hat x)-f^\star]\leq\epsilon\Delta_0\) on every
quadratic family obeying~(1), (13), and local \(\mu\)-strong
convexity. On some family it uses expected rounds
\[
 \bar R\geq\max\Bigl\{\frac m{16s},\
 \frac m{1792\,s}\sqrt{\frac{\min\{\delta_a,\bar\delta/\sqrt m\}}\mu}
 \log\frac1{4\epsilon}\Bigr\}.
\]
Theorem~7's chain has \(\bar\delta\leq\frac{\sqrt3}2\delta\sqrt m\)
(Lemma~C.6), because no client holds two adjacent links, so it embeds
in the two-parameter class. Markov's inequality and truncation turn the
expected-gap guarantee into the event that Theorem~7 excludes.

\textbf{Corollary 12 (every cohort size).} Let
\(\delta=\min\{\delta_a,\bar\delta/\sqrt3\}\geq3\mu\),
\(4\epsilon\leq(1-q)^4\), and \(d\geq K'+1\), where \(K'\) is the
smallest \(R\) with \((1-q)^2q^{2R}\leq4\epsilon\). For every
\(s\leq m\), including \(s=m\), any ZR algorithm as in Corollary~11
needs
\(\mathbb E[R]\geq\frac3{64}\sqrt{\min\{\delta_a,\bar\delta/\sqrt3\}/\mu}
\,\log\frac1{4\epsilon}\) on one fixed instance. The instance
(Lemma~C.7) gives the odd links to one half of the non-anchor clients
and the even links to the other half. No client holds adjacent links,
so every reply extends the support by at most one coordinate per
round, whatever the cohort, while \(\bar\delta\leq\sqrt3\,\delta\).
This follows \citet{arjevani2015communication}; what is new is that the
instance lies in the two-parameter class with a charged anchor.

\textbf{Proposition 13 (batched Acc-S-DANE).} If every \(f_i\) is
\(\mu\)-strongly convex and \(\bar\delta\geq\mu/8\), Acc-S-DANE
\citep[Algorithm~2]{jiang2024} with full participation,
\(\lambda=2\bar\delta\), and exact local steps reaches
\(f(x_R)-f^\star\leq\epsilon\Delta_0\) after
\(R^\ast=\lceil2\sqrt{8\bar\delta/\mu}\log(1+2/\sqrt\epsilon)\rceil\)
iterations. Each iteration costs \(2\lceil m/s\rceil\) strict rounds:
one capped pass for gradients at \(y_r\), and one for the local
proximal points. The total is
\(O((m/s)\sqrt{\bar\delta/\mu}\log(1/\epsilon))\) rounds, and the
method is ZR on quadratics. The proof applies Theorem~6 of
\citet{jiang2024}, whose \(\delta\)-SOD condition at full participation
is (13).

\textbf{An anchor-free method.} Theorems~9--10 send every step through
the anchor, which costs \(\sqrt{m/s}\sqrt{\delta_a/\mu}\) when the anchor
is poor. Proposition~13 avoids the anchor but needs full passes. Our
new method, \emph{virtual-master} cohort SVRS (VM-SVRS), keeps the
SVRS epoch and replaces the anchor's proximal reply with the average of
the proximal replies of a fresh uniform \(s\)-cohort. An epoch from \(w\)
collects \(\nabla f(w)\) in \(\lceil m/s\rceil\) rounds. Step \(t\) then
uses two independent cohorts, one round each. An estimator cohort
\(S'_t\) returns Hessian-vector products, so
\(g_t=\nabla f(w)+\frac1s\sum_{i\in S'_t}\nabla^2f_i\,(x_t-w)\)
(at \(t=0\), \(g_0=\nabla f(w)\)). A master cohort \(S_t\) receives
\(g_t\), and each member returns its local proximal point
\(u_i=\arg\min_uf_i(u)+\langle g_t-\nabla f_i(x_t),u\rangle
+\frac\lambda2\|u-x_t\|^2\). The step is
\(x_{t+1}=\frac1s\sum_{i\in S_t}u_i\). On quadratics,
\(x_{t+1}=x_t-P_{S_t}g_t\) with
\(P_S=\frac1s\sum_{i\in S}(A_i+\lambda I)^{-1}\). The mean of \(P_S\) is
the inverse of a virtual master whose Hessian is within
\(\bar\delta^2/(\mu+\lambda)\) of \(A\) (Lemma~C.8), whereas a single
client can deviate by \(\sqrt{m-1}\,\bar\delta\). The randomness of
\(P_S\) is controlled by sampling without replacement. Unlike S-DANE
with partial participation \citep{jiang2024}, the correction uses the
full snapshot gradient, so there is no gradient-dissimilarity floor.
Unlike SVRP \citep{khaled2023}, the master and estimator cohorts are
independent.

\begin{figure*}[t]
\centering
\includegraphics[width=0.98\textwidth]{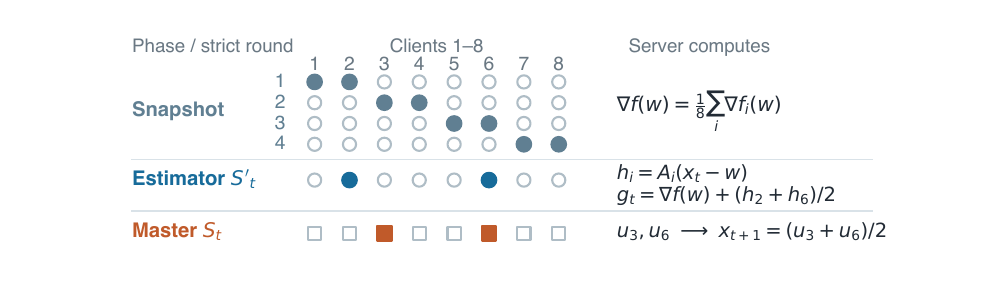}
\caption{Independent cohorts have different jobs in VM-SVRS: one
corrects the gradient, and the other averages local proximal points.
The schematic quadratic case shows \(m=8,s=2\). Four capped snapshot rounds
cover all clients. For \(t\geq1\), the estimator
\(S'_t=\{2,6\}\) supplies Hessian-vector products
\(h_i=A_i(x_t-w)\), where \(A_i=\nabla^2f_i\); the independently drawn master
\(S_t=\{3,6\}\) supplies proximal points \(u_i\), whose mean is
\(x_{t+1}\). Client 6 appears in both cohorts because independence
does not require disjointness. At \(t=0\), the estimator row is skipped
and \(g_0=\nabla f(w)\). Each row costs one strict round; filled marks
indicate contacts with one vector sent and one received.}
\label{fig:vm-architecture}
\end{figure*}

\textbf{Theorem 14 (VM-SVRS epoch).} Let every \(f_i\) be quadratic
with \(\nabla^2f_i\succeq\mu I\), \(\xi=\frac{m-s}{s(m-1)}\),
\(\lambda\geq\max\{\mu,4\bar\delta,4\bar\delta\sqrt{\xi T}\}\), and
\(T\geq\max\{\lceil m/s\rceil,2(\mu+\lambda)/\mu\}\). One epoch costs
\(\lceil m/s\rceil+2T-1\) strict rounds and contracts the potential
\(\Phi(x)=(x-x^\star)^\top\bar P^{-1}(x-x^\star)\), with
\(\bar P=\mathbb EP_S\), by \(\mathbb E\Phi(x_T)\leq0.28\,\Phi(w)\). The
potential is equivalent to the gap up to the factor
\(2(\mu+\lambda)/\mu\). It makes no separate anchor contact; the
anchor may still appear in a sampled cohort.

\textbf{Corollary 15 (Catalyst VM-SVRS).} Wrap VM-SVRS in Catalyst
\citep{lin2018catalyst} with \(T=\max\{\lceil m/s\rceil,4\}\),
\(\lambda=4\max\{\mu,\bar\delta,\bar\delta\sqrt{\xi T}\}\), and shift
\(\kappa=\max\{0,2\lambda/(T-2)-\mu\}\). Each inner problem runs
VM-SVRS epochs until a snapshot gradient passes a Catalyst stopping
test, which costs no extra rounds. Then
\(f(x_K)-f^\star\leq\epsilon\Delta_0\) within expected
\[
 \bar R=O\Bigl(\Bigl[\frac ms
 +\Bigl(\sqrt{\frac ms}+\frac{m^{3/4}}s\Bigr)\sqrt{\frac{\bar\delta}\mu}\Bigr]
 \,\ell\,\Bigl(\ell+\log\frac1\epsilon\Bigr)\Bigr)
 \tag{16}
\]
rounds, where \(\ell=\log(mL_{\max}/\mu)\). The bracketed dependence
on \(s\) and similarity has the form of (15) with \(\delta_a\) replaced
by \(\bar\delta\); Catalyst adds the logarithmic factors. The method
is ZR on quadratics. Its replies
are a gradient, Hessian-vector products, and shifted inverses
\((A_i+(\kappa+\lambda)I)^{-1}g\). The proof (Appendix~\ref{app:anchor-free})
combines an AM--HM bound for the averaged shifted inverses with the
exact finite-population variance (Lemmas~C.8--C.9) and Catalyst's
outer analysis.

\begin{table*}[t]
\centering\small
\caption{Polynomial dependence of ZR expected-round bounds for quadratics
with \(\mu\)-strongly convex clients. This table omits constants and
logarithmic factors: Corollary~15 carries
\(\ell(\ell+\log(1/\epsilon))\), whereas the lower bounds carry
\(L_\epsilon=\log(1/\epsilon)\). The comparison uses the high-accuracy
regime of Corollaries~11--12 with
\(\min\{\delta_a,\bar\delta/\sqrt m\}\geq3\mu\). The lower bound is
\(\max\{\frac ms\sqrt{\min\{\delta_a,\bar\delta/\sqrt m\}/\mu},\
\sqrt{\min\{\delta_a,\bar\delta\}/\mu}\}\). The last column is the
ratio of their polynomial parts.}
\label{tab:regimes}
\begin{adjustbox}{max width=\textwidth}
\begin{tabular}{lll}
\toprule
Anchor regime & Best upper bound & Ratio\\
\midrule
\(\delta_a\leq\bar\delta/\sqrt m\) &
Thm.~8: \(\frac ms\sqrt{\delta_a/\mu}\) & \(\Theta(1)\)\\
\(\bar\delta/\sqrt m\leq\delta_a\leq\sqrt m\,\bar\delta/s\) &
Thm.~10: \(\frac{m^{3/4}}s\sqrt{\bar\delta/\mu}\) & \(\Theta(1)\)\\
\(\delta_a>\sqrt m\,\bar\delta/s\) &
Thm.~10 or Cor.~15: \(\frac{m^{3/4}}s\sqrt{\frac{\bar\delta}\mu}
+\sqrt{\frac ms}\sqrt{\frac{\min\{\delta_a,\bar\delta\}}\mu}\) &
\(O\bigl(1+\min\{\frac{\sqrt s}{m^{1/4}},\ \sqrt{\frac ms}\}\bigr)\leq O(m^{1/8})\)\\
\bottomrule
\end{tabular}
\end{adjustbox}
\end{table*}

\textbf{Complexity map.} Table~\ref{tab:regimes} combines the bounds
(Appendices~\ref{app:large-cohorts} and~\ref{app:vm-gap}). In the first
two rows, the ZR round complexity is
\(\Theta((m/s)\sqrt{\min\{\delta_a,\bar\delta/\sqrt m\}/\mu}\,L_\epsilon)\)
for every \(s\). A good anchor favors the full-pass Chebyshev method,
and otherwise sampled cohorts attain the order. With a poor anchor,
Corollary~15 matches the polynomial dependence when \(s\leq\sqrt m\)
or \(s=\Omega(m)\). Its full logarithmic overhead relative to the
lower bound is \(\ell(\ell+L_\epsilon)/L_\epsilon\). At full participation
the polynomial dependence is
\(\Theta(\sqrt{\min\{\delta_a,\bar\delta\}/\mu})\). The
largest remaining polynomial gap is \(O(m^{1/8})\), at \(s=m^{3/4}\). Without
Corollary~15 it would be \(O(\min\{\sqrt s,m^{1/4},m/s\})\), which is
\(m^{1/4}\) for \(\sqrt m\leq s\leq m^{3/4}\). If clients may be
nonconvex, Proposition~13 and Corollary~15 are unavailable, and the gap
is \(O(\min\{\sqrt s,m^{3/4}/\sqrt s\})\).

\paragraph{GPU study.}
The two-parameter GPU study runs theorem-parameter cohort SVRS
and AccSVRS (with \(b=s\) and \(b=m\)) beside GD, AGD, BAPG, strict
SAGA, exact S-DANE, SVRP, and source-parameter SVRS/AccSVRS, all on the
same strict ledger. Instances set \(\delta_a\) and \(\bar\delta\)
exactly and keep every client \(\mu\)-strongly convex. The
prespecified grid has \(m\in\{32,128\}\), \(d\in\{64,256\}\),
\(s\in\{1,4,16,m\}\), \(\delta_a,\bar\delta\in\{1,10,100\}\), 16 paired
seeds, and a 20,000-round cap. The constraint
\(\bar\delta\geq\delta_a/\sqrt{m-1}\) leaves 28 feasible cells, hence
112 cell--cohort settings. The endpoint is the first strict round at
which the relative gap reaches \(10^{-4}\). A setting's median is
reported only if all 16 seeds reach the target.
Appendix~\ref{app:gpu-study} gives the full grid.

\begin{table}[t]
\centering\footnotesize
\caption{Median strict rounds to relative gap \(10^{-4}\) for the cell
\(m=128,d=256,\delta_a=\bar\delta=10\); all 16 seeds reach the target
in each entry. ``Best'' is the smallest complete-seed median among
non-weighted outputs; APSVRG\(\times\frac12\) is heuristic. C-Acc and
C-SVRS are the theorem-parameter cohort arms with \(b=s\); Src.\ Acc is
source-parameter AccSVRS.}
\label{tab:gpu-example}
\setlength{\tabcolsep}{2.5pt}
\begin{tabular}{rrrrl}
\toprule
\(s\) & C-Acc & C-SVRS & Src.\ Acc & Best (rounds)\\
\midrule
1   & 2862 & 2846 & 1494 & SVRP\(\times2\) (416)\\
4   & 496  & 601  & 1107 & APSVRG\(\times\frac12\) (130)\\
16  & 159  & 197  & 1012 & APSVRG\(\times\frac12\) (34)\\
128 & 112  & 130  & 983  & APSVRG\(\times\frac12\) (27)\\
\bottomrule
\end{tabular}
\end{table}

Theorem-parameter cohort AccSVRS reaches the target in all
\(112\times16=1792\) trials; cohort SVRS misses it for one seed in each
of two settings. At \(s=16\), cohort AccSVRS beats source-parameter
AccSVRS in 27 of 28 cells, and the median across cells of the ratio of
their medians is 6.44 (0.845 at \(s=1\)). Source AccSVRS still samples
one client per inner step, so increasing \(s\) mainly reduces the cost
of its full-gradient refresh. Across \(s\), the max-to-min ratio of observed rounds to the
theorem's shape has median 2.25 over cells for cohort AccSVRS (maximum
11.48) and 1.54 for cohort SVRS; one constant does not fit every cell.
Neither theorem-parameter arm is fastest in any of the 112 settings:
BAPG wins 36, APSVRG \(\times0.5\) 27, SVRP \(\times2\) 21, S-DANE
\(\times0.5\) 19, AGD eight, and SVRP \(\times1\) one, several with
multipliers outside their theorems. The grid uses fixed dimensions and
includes \(s=m\), so it does not test the lower bounds.

A separate GPU study compares VM-SVRS with SVRP on four quadratic
families: dip, poor anchor, mid anchor, and good anchor. It uses 12
paired seeds per cell, \(m\in\{64,256\}\),
\(s\in\{4,16,64,m\}\), and \(\bar\delta\in\{10,100\}\). Of the 56
cells, 40 have incomplete participation. The poor, mid, and good
families set \(\delta_a\) to \(0.9\sqrt{m-1}\,\bar\delta\),
\(\bar\delta\), and \(\bar\delta/\sqrt m\), respectively; the dip
family gives every client a curvature dip. Appendix~\ref{app:vm-numerics}
plots the paired round ratios (Figure~\ref{fig:vm-ratios}) at relative gap
\(10^{-6}\) after selecting multipliers on seeds 0--3 and evaluating on
fresh seeds 4--11.

\section{Conclusion}\label{discussion-and-limitations}

Strict client caps change the communication cost of Hessian-similar
optimization. Our hidden-client construction gives lower bounds even
when cohorts are chosen adaptively, and the charged analyses of existing
methods put their guarantees in the same round units. VM-SVRS replaces a
designated anchor with independent estimator and proximal cohorts,
removing anchor similarity from its round guarantee. Under the stated
high-accuracy quadratic assumptions, the polynomial upper and lower
bounds match for small or near-full cohorts. On the held-out quadratic
grid, VM-SVRS has lower paired point estimates in nine of ten
incomplete-cohort curvature-dip settings; tuned SVRP remains competitive
in the other families.

The lower bounds concern the zero-respecting local oracle; the VM-SVRS
guarantee assumes strongly convex quadratic clients.

%% file: hidden_chain_figure.tex
\centering
\includegraphics[width=0.98\linewidth]{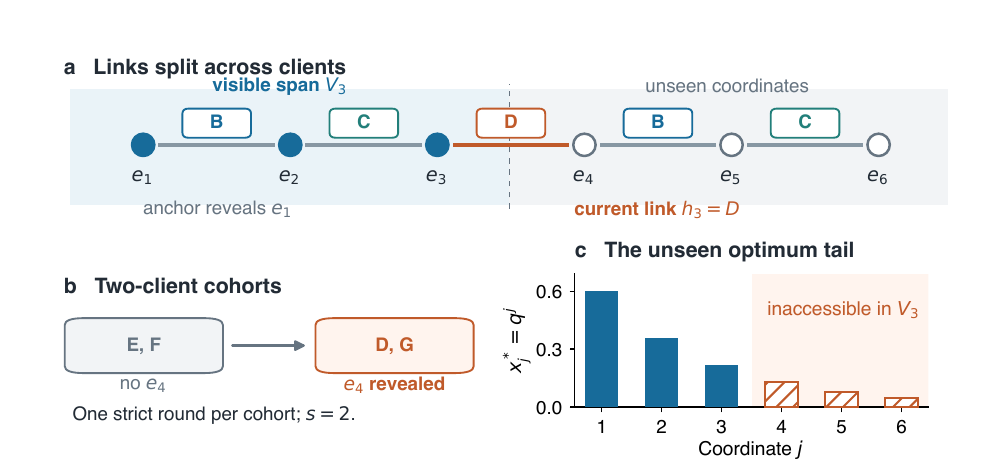}
\caption{Hidden links force both sequential discovery and client
search in the ZR lower bound. (a) In a schematic six-coordinate
assignment, each edge \((x_j-x_{j+1})^2\) has one client holder;
adjacent edges have different holders. At frontier
\(V_3=\operatorname{span}(e_1,e_2,e_3)\), the holder \(h_3=D\) of the
next link remains hidden to the algorithm. (b) With \(s=2\), a cohort
without \(D\) cannot reveal \(e_4\); one containing \(D\) can advance
the frontier by at most one coordinate that round. (c) The optimum has
\(x_j^\star=q^j\); the illustrative choice \(q=0.6\) shows its nonzero
tail outside \(V_3\). Theorem~7 randomizes link holders and proves the
bound for adaptive cohorts.}
\label{fig:hidden-chain}

%% file: appendix.tex
\section*{Guide to the appendix}
Appendices~\ref{app:oracle}--\ref{app:anchor-free} give the model and proofs;
the subsequent sections document experiments, checks, and provenance.

\begin{center}
\small
\begin{tabular}{@{}p{0.16\linewidth}p{0.77\linewidth}@{}}
\toprule
Section & Contents\\
\midrule
Appendix~\ref{app:oracle} & Oracle inputs and charged communication ledgers.\\
Appendix~\ref{app:proofs} & Proofs for majorization, discovery, variance, and controls.\\
Appendix~\ref{app:zr-chain} & Positive-dissimilarity chain and Chebyshev bounds.\\
Appendix~\ref{app:cohort} & Charged cohort SVRS/AccSVRS and two-parameter lower bounds.\\
Appendix~\ref{app:anchor-free} & VM-SVRS, Catalyst, numerical checks, and Figure~\ref{fig:vm-ratios}.\\
\addlinespace[3pt]
Appendix~\ref{app:gpu-study} & Two-parameter GPU study and comparator ledger.\\
Appendix~\ref{app:deferred} & S-DANE accounting and controls.\\
Appendix~\ref{computational-diagnostics-and-artifact-reconciliation} & Computational checks and historical evidence.\\
Appendix~\ref{app:variance-data} & Exact and Monte Carlo variance diagnostics.\\
Appendix~\ref{app:sweep} & Charged 1,926-checkpoint sweep.\\
Appendix~\ref{app:replicates} & Thirty-seed quadratic sensitivity analysis.\\
Appendix~\ref{app:logistic} & Logistic control and tuning correction.\\
\addlinespace[3pt]
Appendix~\ref{app:historical} & Read-only audit of the original P11 study.\\
\bottomrule
\end{tabular}
\end{center}

\ifjmlrlayout\newpage\fi
\section*{Nomenclature}
Unless specified otherwise, the units of communication are strict rounds;
each contacted client receives and returns one $d$-vector.
\begin{center}
\small
\begin{tabular}{@{}p{0.23\linewidth}p{0.70\linewidth}@{}}
\toprule
Symbol & Meaning\\
\midrule
$m,s,d$ & Number of clients, maximum clients contacted per round, and parameter dimension.\\
$f_i,f$ & Client objective and average objective $f=m^{-1}\sum_i f_i$.\\
$x_0,x^\star,\Delta_0$ & Initial point, minimizer of $f$, and initial gap $f(x_0)-f(x^\star)$.\\
$\epsilon,\varepsilon$ & Relative objective-gap fraction in the lower bounds and VM results; absolute gap tolerance in the BAPG bound, respectively.\\
$\mu$ & Global strong-convexity parameter; some upper bounds require each client to be $\mu$-strongly convex.\\
$a,\delta,\delta_a$ & Designated anchor and its certified Hessian-deviation bound; $\delta_a:=\delta$ in the two-parameter results.\\
$\bar\delta$ & Certified average second-order similarity bound; for quadratics, $\bar\delta^2=\|m^{-1}\sum_i(A_i-A)^2\|_{\mathrm{op}}$.\\
$A_i,A,L_{\max}$ & Client and mean quadratic Hessians, and maximum local Hessian norm $\max_i\|A_i\|_{\mathrm{op}}$.\\
$R,C,V$ & Strict rounds, client contacts, and transmitted $d$-vectors, respectively.\\
$h,b,\xi(b)$ & Full-pass cost $h=\lceil m/s\rceil$, sampled cohort size $b\leq s$, and finite-population factor $(m-b)/[b(m-1)]$.\\
$w,x_t$ & Snapshot point and within-epoch iterate.\\
$S'_t,S_t$ & Independent estimator and proximal cohorts in VM-SVRS.\\
$g_t,u_i$ & Corrected gradient estimate and client $i$'s proximal reply in VM-SVRS.\\
$P_S,\bar P$ & Cohort-averaged shifted inverse and its expectation in the quadratic virtual-master analysis.\\
$\lambda,\kappa$ & Proximal regularization and, in the virtual-master proof, Catalyst shift; Appendix~\ref{app:zr-chain} uses $\kappa$ locally for a condition number.\\
ZR & Zero-respecting local oracle used for the lower bounds, with adaptive cohorts and random stopping allowed.\\
\bottomrule
\end{tabular}
\end{center}

\section{Oracle, notation, and complete communication ledgers}\label{app:oracle}

\subsection{Information available to the server}\label{app:information}
The optimization oracle starts with $x_0$, the integers $m$ and $s$, a valid
clientwise lower curvature bound $\mu>0$, the anchor identity $a$, and a
certified anchor-to-global Hessian bound $\delta$. In the two-parameter
setting, methods that set parameters from average second-order similarity
also receive a certified $\bar\delta$; this is a model input, not a
quantity estimated without communication. $L_{\max}$ enters the stated
rate bound, but not the VM-SVRS parameter rule. The server holds no client gradient,
linear coefficient, feature matrix, local minimizer, or aggregate gradient.
In a strict round the server chooses at most $s$ client identities, sends one
$d$-vector to each, and receives at most one $d$-vector from each. The query
may be the same point for all selected clients. The vector sent to the anchor
for a proximal solve is a proximal input, not a free evaluation of $f_a$ at an
arbitrary point. Local gradient evaluation and an exact local solve are free
\emph{between} contacts. For every contact the ledger counts one outgoing and
one incoming $d$-vector, even when a broadcast implementation could physically
reuse bytes. Neither arithmetic, bit precision, nor synchronization headers are
priced by this model.

Write $h=\lceil m/s\rceil$. A capped full-gradient pass partitions the $m$
clients into $h$ disjoint batches of size at most $s$. Each client supplies
$\nabla f_i(y)$ once, giving $m$ contacts and $2m$ transmitted vectors. The
anchor's gradient is already one of these replies; there is no extra anchor
gradient contact. After the last batch, the server knows $\nabla f(y)$ and
$\nabla f_a(y)$ and can form the residual gradient. It can only then determine
the input $y-\nabla(f-f_a)(y)/\lambda$ for the anchor prox. Thus the prox
requires a subsequent strict round and one additional contact. The formulas
for $T$ completed BAPG steps are
\[
 (R,C,V)=\bigl(T(h+1),\;T(m+1),\;2T(m+1)\bigr).
\]
The bound $s=m$ gives $h=1$, so even full participation does not collapse a
BAPG step into one strict round: the prox input depends on the returned
full gradient. No pipelining of a future step is possible before its current
iterate is known.

\begin{table*}[h!]
\centering\small
\caption{End-to-end ledger for completed operations. $h=\lceil m/s\rceil$;
$T$ counts optimizer updates, $d+1$ counts distinct quadratic probe points.
Every vector total counts a query and a reply per contact.}
\label{tab:appendix-ledger}
\begin{tabular}{p{0.25\textwidth}rrrp{0.24\textwidth}}
\toprule
Method or operation & Strict rounds & Contacts & $d$-vectors & Local work and condition\\
\midrule
Capped full-gradient step & $h$ & $m$ & $2m$ & $m$ local gradients\\
BAPG, $T$ steps & $T(h+1)$ & $T(m+1)$ & $2T(m+1)$ & $T$ anchor proximal solves\\
Cached affine, $T\geq1$ & $h+T$ & $m+T$ & $2(m+T)$ & Requires $\delta=0$\\
One-shot affine & $h+1$ & $m+1$ & $2(m+1)$ & Requires $\delta=0$, one target-dependent prox\\
Batched GD, $T$ steps & $Th$ & $Tm$ & $2Tm$ & Certified global smoothness supplied\\
Strict SAGA initialization & $h$ & $m$ & $2m$ & One gradient per table entry\\
Strict SAGA, $U$ updates after initialization & $h+U$ & $m+Us$ & $2(m+Us)$ & Certified maximum local smoothness supplied\\
Quadratic identification & $(d+1)h$ & $m(d+1)$ & $2m(d+1)$ & Quadratic class only\\
Literal S-DANE, $K$ iterations & $3K$ & $3sK$ & $6sK$ & $sK$ local solves; distinct assumptions\\
\bottomrule
\end{tabular}
\end{table*}

Batched GD uses the same full-gradient pass but updates on the server and has
no client prox reply. The implementation's SAGA table is initialized by a
capped full pass at $x_0$; thereafter a size-$s$ cohort returns fresh
individual gradients and the server updates its stored entries. The
initialization remains charged even if its objective-gap checkpoint is
identical to the starting gap. When $m$ is not divisible by $s$, the last
initialization and full-gradient batch has fewer than $s$ contacts. The
reported formula $m+Us$ uses size-$s$ SAGA updates, so it still counts exact
contacts. Quadratic identification makes $d+1$ separate full-gradient passes;
server-side matrix reconstruction and inversion do not create oracle rounds.

The S-DANE entry is a \emph{literal protocol transcription}, not a theorem
comparison or a claim that every implementation needs three stages. For each
iteration, the selected cohort first returns gradients at the current point.
The server then sends the resulting cohort mean and receives local solutions.
It finally sends the solution-dependent query vectors and receives gradients
at the local solutions. The second stage cannot begin before the first
cohort's replies, and the third depends on the second. That schedule is
feasible with three one-reply rounds and $3s$ contacts. The S-DANE theory in
\citet{jiang2024} also invokes subset internal/external dissimilarity and
bounded gradient variance, which are not consequences of one-anchor
similarity. The ledger does not translate its theorem constants into ours.

\subsection{Diagnostic evaluator and tuning information}\label{app:diagnostic-oracle}
The synthetic studies hold all generated client functions in one process.
They use this omniscient view to calculate $f(x)$ and a numerical $f^\star$
for plotting and for the saved endpoint indicator. These diagnostic values
never enter an iterate update, a client choice, or a stopping decision. The
studies therefore charge the optimization trajectory under the strict
interface, but the figures are not an executable distributed implementation
of the diagnostic evaluator. A real stopping certificate would require
additional communication or assumptions. Batched GD is supplied a certified
global smoothness $L_{\rm global}$; SAGA is supplied the maximum certified
local smoothness $L_{\rm local}$. Those comparator inputs are explicit extra
information. BAPG uses the public $\mu,\delta$ and the chosen $\lambda$.

\section{Proofs of the strict-oracle results}\label{app:proofs}

\subsection{BAPG majorization, contraction, and inexact solves}\label{app:bapg-proof}
Let $r=f-f_a$. Since $\nabla^2r=\nabla^2 f-\nabla^2 f_a$ has operator norm at
most $\delta$ everywhere, the fundamental theorem of calculus along the
segment from $y$ to $x$ gives
\begin{align*}
 \nabla r(x)-\nabla r(y)
 &=\int_0^1\nabla^2r(y+t(x-y))(x-y)\,dt,\\
 \|\nabla r(x)-\nabla r(y)\|
 &\leq\delta\|x-y\|.
\end{align*}
Integrating the directional derivative once more yields a two-sided
remainder: for every $x,y$,
\begin{equation}\label{eq:app-remainder}
 -\frac\delta2\|x-y\|^2\leq
 r(x)-r(y)-\langle\nabla r(y),x-y\rangle
 \leq\frac\delta2\|x-y\|^2.
\end{equation}
This argument never assumes $r$ is convex. For $\lambda\geq\delta$,
$\lambda>0$, adding $f_a(x)$ to the upper side of
\eqref{eq:app-remainder} gives $f(x)\leq Q_y(x)$. Conversely, the
lower side gives
\begin{equation}\label{eq:app-model-comparison}
 Q_y(z)\leq f(z)+\frac{\lambda+\delta}{2}\|z-y\|^2.
\end{equation}
The first inequality is a true majorization of $f$ at every $x$; the second
compares the model with $f$ at a candidate $z$. Both are needed: a one-sided
smoothness statement alone would not justify this particular factor.

For an exact minimizer $x^+$ of $Q_y$, choose
$z_t=(1-t)y+tx^\star$, $0\leq t\leq1$. Clientwise $\mu$-strong
convexity implies $f$ is $\mu$-strongly convex, and therefore
\begin{align*}
 f(x^+)-f(x^\star)
 &\leq Q_y(x^+)-f(x^\star)\\
 &\leq Q_y(z_t)-f(x^\star)\\
 &\leq (1-t)\{f(y)-f(x^\star)\}\\
 &\quad+\frac{(\lambda+\delta)t^2-\mu t(1-t)}2
   \|y-x^\star\|^2.
\end{align*}
The coefficient of the squared distance is
$t\{(\mu+\lambda+\delta)t-\mu\}/2$. At
$t_\star=\mu/(\mu+\lambda+\delta)$ it vanishes, and $1-t_\star$ is exactly
$q=(\lambda+\delta)/(\mu+\lambda+\delta)$. Thus
$\Delta_{t+1}\leq q\Delta_t$. Iterating gives $q^T\Delta_0$, and
$-\log q=\log(1+\mu/(\lambda+\delta))$ gives the main-text step formula.
The weaker but sometimes convenient bound
$T\leq\lceil(1+(\lambda+\delta)/\mu)\log(\Delta_0/\varepsilon)\rceil$
follows from $\log(1+u)\geq u/(1+u)$. The derivative
$\partial q/\partial\lambda=\mu/(\mu+\lambda+\delta)^2>0$ shows why
$\lambda=\delta$ is the tightest displayed choice when $\delta>0$.

If the anchor returns $\tilde x$ with
$Q_y(\tilde x)-\min Q_y\leq e_t$, the middle comparison above increases
by at most $e_t$. Consequently
$\Delta_{t+1}\leq q\Delta_t+e_t$ and expansion of the recursion gives
\begin{equation}\label{eq:app-inexact}
 \Delta_T\leq q^T\Delta_0+
  \sum_{t=0}^{T-1}q^{T-1-t}e_t.
\end{equation}
For instance, choosing $e_t\leq(1-q)\varepsilon/2$ for every completed
step and $q^T\Delta_0\leq\varepsilon/2$ makes the right-hand side at most
$\varepsilon$. Since $Q_y$ is $(\mu+\lambda)$-strongly convex, the
standard strong-convexity inequality, minimized over a displacement from
$\tilde x$, gives
\[
 Q_y(\tilde x)-\min Q_y
 \leq\frac{\|\nabla Q_y(\tilde x)\|^2}{2(\mu+\lambda)}.
\]
This is the local residual certificate used for the logistic prox. A
quadratic prox in the saved sweep is a floating-point linear solve whose
logged model-gap bound is zero by the ideal-oracle convention; those rows
are not independent finite-precision residual certificates.

\subsection{Hidden-client discovery with adaptive queries}\label{app:discovery-proof}
For Theorem 2, take $x_0=0$ when interpreting a relative reduction. The
hidden-sign family has
$f(x)=\mu x^2/2-\mu B\theta x$ and
$f(x)-f(x^\star)=\mu(x-B\theta)^2/2$; thus its initial gap is
$\mu\theta^2/2$. All local Hessians and the global Hessian equal $\mu$,
so the anchor gap is zero irrespective of which client is designated
anchor. In particular a server does not learn $B$ from its initial
metadata.

Fix all internal random coins of an arbitrary adaptive algorithm. Imagine
that every query, including a possible prox query, is answered by the
ordinary function $\mu x^2/2$. This defines one null transcript and one
sequence of contacted sets through $R$ rounds. Let $U$ be their union, so
$|U|\leq Rs$ even if the same client is visited repeatedly. Draw the hidden
identity $J$ uniformly. On $J\notin U$, induction over rounds shows that
the actual transcript coincides with the null transcript: before each
round the histories agree, hence the adaptive query and cohort agree,
and each selected client is ordinary. The algorithm's output on this
event is consequently independent of the independently drawn sign $B$.
For any fixed real output $z$,
\[
 \frac12\{(z-\theta)^2+(z+\theta)^2\}=z^2+\theta^2\geq\theta^2.
\]
The event has probability $1-|U|/m\geq(1-Rs/m)_+$. Averaging first over
$B$ and $J$, then over the algorithm's coins, gives the asserted
expected objective gap. Finally, an average over a finite distribution
of instances cannot exceed every constituent risk, so one fixed
$(J,B)$ instance has risk at least that average. This last selection is
why the theorem applies to randomized algorithms rather than only to a
random hard instance. The argument bounds information acquisition; it
does not constrain the form of the server's iterate, assume a linear
span, or yield any factor depending on a positive $\delta$.

\subsection{Finite-population identity and the prescribed-gap construction}\label{app:variance-proof}
For vectors $u_1,\ldots,u_m$ with mean $\bar u$, set $d_i=u_i-\bar u$,
so $\sum_i d_i=0$. If $S$ is uniform among all size-$s$ subsets without
replacement, indicators $I_i={\bf1}\{i\in S\}$ obey
$\mathbb E I_i=s/m$ and
$\mathbb E I_iI_j=s(s-1)/[m(m-1)]$ for $i\ne j$. Expanding the square and using
$\sum_{i\ne j}\langle d_i,d_j\rangle=-\sum_i\|d_i\|^2$ gives the exact vector identity
\begin{align}\label{eq:app-fpc}
 \mathbb E\left\|\frac1s\sum_{i\in S}u_i-\bar u\right\|^2
 &=\frac1{s^2}\left[\frac{s}{m}
 -\frac{s(s-1)}{m(m-1)}\right]\sum_i\|d_i\|^2\notag\\
 &=\frac{m-s}{s(m-1)}\frac1m\sum_i\|d_i\|^2.
\end{align}
Thus the finite-population correction is exactly
$\xi(s)=(m-s)/[s(m-1)]$, including $\xi(1)=1$ and $\xi(m)=0$.
No independence of selected client indicators is assumed.

Apply \eqref{eq:app-fpc} to
$u_i=\nabla f_i(x)-\nabla f_i(w)$. The theorem's Hessians give
$A=\mu+M$ because their deviations from $\mu+M$ are
$\delta,-M,M-\delta,0,\ldots,0$, which sum to zero. The condition
$M\geq\delta/2$ implies $A_2=\mu+2M-\delta\geq\mu$;
$A_0,A_1$, and all other $A_i$ also meet the clientwise lower bound.
The designated anchor is client zero and its gap is
$|A_0-A|=\delta$, not merely at most $\delta$. In one dimension,
$u_i=A_i(x-w)$, hence
\[
 \frac1m\sum_i|u_i-\bar u|^2
 =\frac{\delta^2+M^2+(M-\delta)^2}{m}|x-w|^2.
\]
Equation \eqref{eq:app-fpc} now yields (8). At any fixed nonzero
$x-w$, its leading term as $M\to\infty$ is $2M^2/m$ times the
strictly positive $\xi(s)$, while the anchor gap stays exactly the
prescribed $\delta$. Subtracting
$\nabla f_a(x)-\nabla f_a(w)$ from both the sample and its target
leaves their difference unchanged. If $s=m$ or $x=w$, the error is zero,
as the formula requires.

This construction blocks a variance estimate of the form
$C(\delta,\mu,m,s)\|x-w\|^2$ for a fresh sampled difference over the
stated class, because $M$ remains free. It supplies no lower bound on
optimization error after an adaptive sequence of contacts. A table
method may learn client-specific deviations, and an algorithm can
use the exact local-prox oracle rather than this estimator. In
particular, one cannot multiply the variance obstruction by a
contraction argument and claim a positive-$\delta$ minimax lower rate.

The original project's componentwise assumption,
$\sup_{i,x}\|\nabla^2f_i(x)-\nabla^2f(x)\|_{\rm op}\leq D$, is stronger.
Integrating along the $w$--$x$ segment bounds each centered
$u_i-\bar u$ by $D\|x-w\|$, and \eqref{eq:app-fpc} then gives
$\xi(s)D^2\|x-w\|^2$. This mathematical implication is valid under
that \emph{all-client} condition. The current anchor condition bounds
only the $i=a$ term, and Theorem 3 demonstrates why replacing $D$ by
$\delta$ in the same step is invalid.

\subsection{First-order offsets and sampled-anchor noise}\label{app:first-order-noise}
Theorem 3 studies a \emph{difference} of gradients at $x$ and $w$;
linear client offsets cancel from that difference. A separate question
is the noise of a fresh sampled \emph{anchor gradient} at one point.
Define $\widehat g_S(w)=s^{-1}\sum_{i\in S}\nabla f_i(w)$. Applying
equation~\eqref{eq:app-fpc} to $v_i(w)=\nabla f_i(w)$ gives, without
any Hessian assumption,
\begin{equation}\label{eq:app-anchor-noise}
 \begin{aligned}
 \mathbb E_S\|\widehat g_S(w)-\nabla f(w)\|^2
 &=\frac{\xi(s)}{m}\\
 &\quad\times\sum_i\|\nabla f_i(w)-\nabla f(w)\|^2.
 \end{aligned}
\end{equation}
The one-anchor Hessian bound places no restriction on the right-hand
side because it does not bound linear terms. In the hidden-client
family of Theorem 2, the ordinary clients' deviations from the global
gradient are $\mu B\theta$ and the hidden client's deviation is
$-(m-1)\mu B\theta$, at every $w$. Their population second moment is
$(m-1)\mu^2\theta^2$. Thus the sampled-anchor variance is exactly
\[
 \xi(s)(m-1)\mu^2\theta^2
 =\frac{m-s}{s}\mu^2\theta^2,
\]
including at the global optimum. It can grow with $\theta$ while all
Hessians are identical and $\delta=0$. By contrast, the same family's
fresh \emph{gradient-difference} estimator has zero variance because
$\nabla^2 f_i=\nabla^2 f$ for every client. In the prescribed-gap
family of Theorem 3, all linear terms are zero, so the global optimum
is the origin and sampled-anchor variance at that optimum is zero,
while a nonzero $x-w$ yields the arbitrarily large difference variance
in (8). These two examples isolate distinct information gaps.

The original sampled-anchor floor plot concerns a particular algorithm
that substitutes a sampled estimate for an exact aggregate anchor
gradient. Equation~\eqref{eq:app-anchor-noise} explains why such an
algorithm can have persistent noise when client gradients disagree at
the optimum. It does \emph{not} show that every strict-oracle algorithm
has a persistent optimization floor. A complete pass removes this
specific anchor sampling noise; a stored gradient table can also change
the estimator. Neither the floor plot nor Theorem 3 should be
promoted from an estimator statement to a universal optimization lower
bound.

\subsection{Affine caching, one-shot reduction, and quadratic identification}\label{app:controls-proof}
At $\delta=0$, the Hessian of $r=f-f_a$ vanishes everywhere; on the
connected space $\mathbb R^d$, $\nabla r(x)=c$ is constant. A capped
full-gradient pass at $x_0$ supplies
$c=\nabla f(x_0)-\nabla f_a(x_0)$. After this pass, the BAPG model at
any later $y$ has the same residual gradient. With $\lambda=\mu$,
each anchor-only prox contact is exactly the BAPG update and contracts
objective gap by at most one half. The $T$ cached contacts follow the
initial $h$ rounds, giving $(h+T,m+T,2(m+T))$ and the inexact weighted
recursion \eqref{eq:app-inexact} with $q=1/2$. This equality of iterates
uses an exact residual gradient; a perturbed cached value would
accumulate error and is outside Proposition 4.

For a fixed relative target $0<\tau<1$, choose
$\lambda=\mu\tau/(1-\tau)$. Then
$q=\lambda/(\mu+\lambda)=\tau$, so one full pass and one anchor prox
give $\Delta_1\leq\tau\Delta_0$ and cost
$(h+1,m+1,2(m+1))$. If the returned model point has error $e$,
its objective error is at most $\tau\Delta_0+e$. The parameter is
positive for every admissible target, but its prox input
$x_0-c/\lambda$ may demand high precision as $\tau\downarrow0$. The
communication result cannot be read as a uniform numerical-work bound.
For a fixed relative reduction stricter than one half, the hidden-client
family requires more than $m/(2s)$ rounds in the worst case, while this
one-shot schedule takes $h+1$; these have the same $m/s$ order up to
constants. There is no positive-$\delta$ analogue from this argument.

For quadratic identification, write
$f_i(x)=\tfrac12x^\top A_ix-b_i^\top x$ with symmetric
$A_i\succeq\mu I$. Query complete gradients at
$x_0,x_0+e_1,\ldots,x_0+e_d$. Let $g_j$ denote the average reply
at probe $j$, with $g_0=Ax_0-b$. Since
$g_j-g_0=Ae_j$, the $d$ differences are all columns of the global
$A$. The server forms $b=Ax_0-g_0$ and solves $Ax=b$ locally. The
positive eigenvalue lower bound makes $A$ invertible. No gradient at
the solved point, objective value, or extra anchor contact is needed.
All $d+1$ probes each cost $h$ rounds and $m$ contacts. This result is
class-specific: general smooth strongly convex functions are not
identified by finitely many affine gradient probes. It invalidates a
universal accuracy-dependent lower bound inferred solely from a fixed
$d$ quadratic rate plot under this oracle.

\input{positive_chain}

\input{cohort_accsvrs}

\input{anchor_free}

\input{gpu_study_results}

\input{deferred_main}

\input{diagnostic_details}

\section{Exact and Monte Carlo variance diagnostics}\label{app:variance-data}
The variance experiment instantiates Theorem 3 with $m=12$, $\mu=1$,
$x=2$, $w=0$, and uniform incomplete cohorts
$s\in\{1,2,4,8,11\}$. The first three saved files use $\delta=0$ and
$M\in\{1,3,9\}$; the fourth uses $\delta=1,M=3$. For each cell the
calculation enumerates all $\binom{12}{s}$ cohorts (at most $924$) and
averages the squared centered sample difference. It separately draws
20,000 uniform cohorts without replacement with NumPy generator seed 17.
The reported Monte Carlo standard error is the sample standard
deviation of the 20,000 \emph{squared} errors, divided by
$\sqrt{20,000}$. It is an uncertainty estimate for that simulation
mean, not a confidence interval for an optimization rate.

\begin{table*}[t]
\centering\small
\caption{All saved finite-population cells. The exact column is both
formula (8) and enumeration to floating-point precision. MC entries
are the 20,000-draw mean with its standard error. The three $\delta=0$
scales restart the same seed and therefore share paired cohort draws.}
\label{tab:appendix-variance-all}
\begin{tabular}{rrrll@{\qquad}rrrll}
\toprule
$\delta$ & $M$ & $s$ & Exact & MC (SE) & $\delta$ & $M$ & $s$ & Exact & MC (SE)\\
\midrule
0 & 1 & 1  & 0.66667 & 0.66460 (0.0105) & 0 & 3 & 1  & 6.00000 & 5.98140 (0.0948)\\
0 & 1 & 2  & 0.30303 & 0.29950 (0.00324) & 0 & 3 & 2  & 2.72727 & 2.69550 (0.0292)\\
0 & 1 & 4  & 0.12121 & 0.12093 (0.000883) & 0 & 3 & 4  & 1.09091 & 1.08833 (0.00795)\\
0 & 1 & 8  & 0.03030 & 0.03018 (0.000221) & 0 & 3 & 8  & 0.27273 & 0.27160 (0.00199)\\
0 & 1 & 11 & 0.00551 & 0.00545 (0.0000868) & 0 & 3 & 11 & 0.04959 & 0.04909 (0.000781)\\
\midrule
0 & 9 & 1  & 54.00000 & 53.83260 (0.853) & 1 & 3 & 1  & 4.66667 & 4.64680 (0.0736)\\
0 & 9 & 2  & 24.54545 & 24.25950 (0.262) & 1 & 3 & 2  & 2.12121 & 2.08525 (0.0225)\\
0 & 9 & 4  & 9.81818 & 9.79493 (0.0716) & 1 & 3 & 4  & 0.84848 & 0.84841 (0.00619)\\
0 & 9 & 8  & 2.45455 & 2.44443 (0.0179) & 1 & 3 & 8  & 0.21212 & 0.21175 (0.00155)\\
0 & 9 & 11 & 0.44628 & 0.44182 (0.00703) & 1 & 3 & 11 & 0.03857 & 0.03800 (0.000604)\\
\bottomrule
\end{tabular}
\end{table*}

At $\delta=0$, the formula reduces to
$\xi(s)(2M^2/m)|x-w|^2$. Consequently all three $M$ columns must
be exact multiples of one another by factors $1:9:81$, and the common
seed makes their Monte Carlo values the same rescaling up to rounding.
This pairing is useful for checking the implementation, but it is not
15 independent experimental replications. At positive gap the centered
Hessians are $(1,-3,2,0,\ldots,0)$, which give population second
moment $(1+9+4)/12=7/6$ and exactly the five values in the last
block of Table~\ref{tab:appendix-variance-all}. Across the twenty
cells the largest absolute standardized Monte Carlo deviation from
the exact value is about $1.60$ standard errors (the $\delta=1,s=2$
cell); the exact enumeration agrees with the formula to machine
precision. This finite calculation checks the identity and sampling,
while unboundedness at fixed $\delta$ follows only from the analytic
$M\to\infty$ argument.

\section{Charged 1,926-checkpoint sweep}\label{app:sweep}

\subsection{Instance construction and comparator steps}\label{app:sweep-design}
The full CPU sweep uses pseudorandom seed 11 and a 96-round cap. One
$m=24,d=3$ quadratic family is shared across all four perturbation
scales $0,0.4,2,6$ and all three cohort sizes $s=1,4,8$.
The common average Hessian is
$A=\operatorname{diag}(10.5,11.5,12.5)$; the client perturbation
matrices are symmetric, centered to average zero, and normalized by
the largest client operator norm before multiplication by the scale.
A common generated optimum $x^\star$ and centered client linear
offsets are held fixed. The actual anchor bound is
$\delta=\|A_0-A\|_{\rm op}$, rather than the nominal perturbation
scale. The public lower-curvature bound is $\mu=0.5$, valid but loose
for this family. Its seed-11 anchor gaps at scales $0,0.4,2,6$ are
$0$, $0.160660$, $0.803298$, and $2.409895$, respectively.
A separate control uses $m=12,n=30$ observations per client,
$d=3$ regularized-logistic instance at $s=2$. The logistic $\mu=0.5$
is supplied by its $\ell_2$ penalty. Its seed-11 anchor certificate is
$\delta=0.783869$, from a Gram-matrix bound, not an observed
supremum over $x$.

Batched GD uses $x^+=x-\nabla f(x)/L_{\rm global}$ with
$L_{\rm global}=\lambda_{\max}(A)$ for quadratics and
$\mu+\tfrac14\lambda_{\max}(\overline{X^\top X}/n)$ for logistic
regression. Strict SAGA initializes all $m$ stored gradients, then
samples a uniform size-$s$ cohort and updates using
\[
 \widehat g=\overline g_{\rm table}
 +\frac1s\sum_{i\in S}(\nabla f_i(x)-g_i^{\rm old}),
 \qquad x^+=x-\frac{\widehat g}{3L_{\rm local}}.
\]
Here $L_{\rm local}$ is the maximum certified local smoothness. The
factor $3$ is the experiment's fixed step divisor; no theorem in this
paper establishes an optimal SAGA tuning. BAPG takes
$\lambda=\delta$ at positive $\delta$ and $\lambda=\mu$ at zero.
For $0<\delta<\mu$ it also saves a \emph{previous-parameter} BAPG
trace with $\lambda=\mu$ at identical contact budgets. The $\delta=0$
quadratic scenarios additionally save cached affine and one-shot
affine controls. Quadratic identification is run only on quadratic
scenarios. All methods use the same initial point $x_0=0$.

\begin{table*}[t]
\centering\small
\caption{Full-sweep inventory and last saved objective gaps. The 13
scenario counts sum to 1,926 checkpoint rows, including initial rows.
The table shows three common methods; the saved JSON also includes
quadratic identification, affine controls where $\delta=0$, and the
previous BAPG parameter where $0<\delta<\mu$. Values shown as zero
were clipped to zero by the diagnostic evaluator at numerical precision.}
\label{tab:appendix-sweep}
\begin{tabular}{lrrrrr}
\toprule
Scenario & $s$ & Checkpoints & BAPG gap & GD gap & SAGA gap\\
\midrule
Quadratic, scale 0 & 1 & 163 & $5.53\cdot10^{-8}$ & $2.16\cdot10^{-6}$ & $5.06\cdot10^{-4}$\\
Quadratic, scale 0 & 4 & 221 & 0 & 0 & $8.05\cdot10^{-11}$\\
Quadratic, scale 0 & 8 & 254 & 0 & 0 & $8.88\cdot10^{-16}$\\
Quadratic, scale 0.4 & 1 & 92 & $4.09\cdot10^{-10}$ & $2.16\cdot10^{-6}$ & $4.99\cdot10^{-4}$\\
Quadratic, scale 0.4 & 4 & 142 & 0 & 0 & $9.58\cdot10^{-11}$\\
Quadratic, scale 0.4 & 8 & 183 & 0 & 0 & 0\\
Quadratic, scale 2 & 1 & 88 & $3.62\cdot10^{-6}$ & $2.16\cdot10^{-6}$ & $6.42\cdot10^{-4}$\\
Quadratic, scale 2 & 4 & 128 & $8.88\cdot10^{-16}$ & 0 & $1.86\cdot10^{-10}$\\
Quadratic, scale 2 & 8 & 158 & $8.88\cdot10^{-16}$ & 0 & 0\\
Quadratic, scale 6 & 1 & 88 & $8.01\cdot10^{-4}$ & $2.16\cdot10^{-6}$ & $2.04\cdot10^{-3}$\\
Quadratic, scale 6 & 4 & 128 & $1.24\cdot10^{-14}$ & 0 & $4.75\cdot10^{-10}$\\
Quadratic, scale 6 & 8 & 158 & 0 & 0 & $2.66\cdot10^{-15}$\\
Logistic & 2 & 123 & $3.87\cdot10^{-9}$ & 0 & $1.22\cdot10^{-9}$\\
\bottomrule
\end{tabular}
\end{table*}

The endpoint table can hide trajectories that briefly satisfy the
target and then move above it. Table~\ref{tab:appendix-first-crossing}
therefore reports the \emph{first saved checkpoint} with relative gap
at most $10^{-4}$ of that trace's initial gap. These are descriptive
first crossings, not endpoint successes, stopping times of the
implemented methods, or guaranteed costs. In particular, seed-11
strict SAGA at scale 6 and $s=1$ first crosses at round/contact
$36/36$, but its last relative gap is $2.78\cdot10^{-4}$, above the
$10^{-4}$ target. The zero-, 0.4-, and 2-scale $s=1$ SAGA traces
also repeatedly cross in both directions, although their final
checkpoints satisfy this seed's target. The replicated study
accordingly uses final checkpoints for its success counts.

\begin{table*}[t]
\centering\small
\caption{Seed-11 first saved relative-target crossing as strict
rounds / contacts. A dash means no saved crossing by the 96-round
cap. Later iterates may exceed the target, particularly for SAGA;
Table~\ref{tab:appendix-sweep} gives final gaps. Identification is
class-specific and absent on logistic data.}
\label{tab:appendix-first-crossing}
\begin{tabular}{lrrrrr}
\toprule
Family or scale & $s$ & BAPG & Batched GD & Strict SAGA & Quadratic identification\\
\midrule
Quadratic 0 & 1 & 50 / 50 & 72 / 72 & 35 / 35 & 96 / 96\\
Quadratic 0 & 4 & 14 / 50 & 18 / 72 & 16 / 64 & 24 / 96\\
Quadratic 0 & 8 & 8 / 50 & 9 / 72 & 14 / 112 & 12 / 96\\
Quadratic 0.4 & 1 & 50 / 50 & 72 / 72 & 74 / 74 & 96 / 96\\
Quadratic 0.4 & 4 & 14 / 50 & 18 / 72 & 16 / 64 & 24 / 96\\
Quadratic 0.4 & 8 & 8 / 50 & 9 / 72 & 20 / 160 & 12 / 96\\
Quadratic 2 & 1 & 50 / 50 & 72 / 72 & 66 / 66 & 96 / 96\\
Quadratic 2 & 4 & 14 / 50 & 18 / 72 & 28 / 112 & 24 / 96\\
Quadratic 2 & 8 & 8 / 50 & 9 / 72 & 15 / 120 & 12 / 96\\
Quadratic 6 & 1 & --- & 72 / 72 & 36 / 36 & 96 / 96\\
Quadratic 6 & 4 & 28 / 100 & 18 / 72 & 28 / 112 & 24 / 96\\
Quadratic 6 & 8 & 16 / 100 & 9 / 72 & 25 / 200 & 12 / 96\\
Logistic & 2 & 56 / 104 & 12 / 24 & 23 / 46 & ---\\
\bottomrule
\end{tabular}
\end{table*}

For $m=24,s=1$, one BAPG step costs 25 rounds, so the cap fits
three steps and its last checkpoint is at round/contact $75/75$.
GD fits four full-gradient steps at $96/96$. SAGA spends 24 rounds
initializing and 72 updates, also ending at $96/96$. Quadratic
identification requires four probes, exactly $96/96$. At $s=4$,
BAPG fits 13 steps at $91/325$, GD fits 16 at $96/384$, SAGA spends
6 initialization rounds then 90 cohort updates at $96/384$, and
identification finishes at $24/96$. At $s=8$, BAPG fits 24 steps at
$96/600$, GD 32 at $96/768$, SAGA has three initialization rounds
and 93 updates at $96/768$, and identification finishes at $12/96$.
The logistic BAPG run fits 13 seven-round steps at $91/169$;
GD and SAGA end at $96/192$. Each contact contributes two vectors.
These schedules explain why comparing only final gaps does not compare
identical contact budgets or numbers of optimizer updates.

The seed-11 zero-gap one-shot control uses
$\tau=10^{-4}$ and
$\lambda=\mu\tau/(1-\tau)=5.00050005\cdot10^{-5}$.
It goes from gap $7.34220098$ to $1.56135549\cdot10^{-10}$,
a relative gap about $2.13\cdot10^{-11}$. It uses 25 contacts
and 50 vectors at each tested cohort, in 25, 7, and 4 rounds for
$s=1,4,8$ respectively. The quadratic identification control uses
96 contacts and 192 vectors, finishing in 96, 24, and 12 rounds.
At scale $0.4,s=1$, the corrected $\lambda=\delta$ and previous
$\lambda=\mu$ BAPG traces both end at round/contact $75/75$;
their saved gaps are $4.09\cdot10^{-10}$ and
$6.87\cdot10^{-8}$. This is one fixed-seed parameter comparison,
not a general empirical superiority claim.

\begin{figure*}[t]
\centering
\includegraphics[width=0.94\textwidth,keepaspectratio]{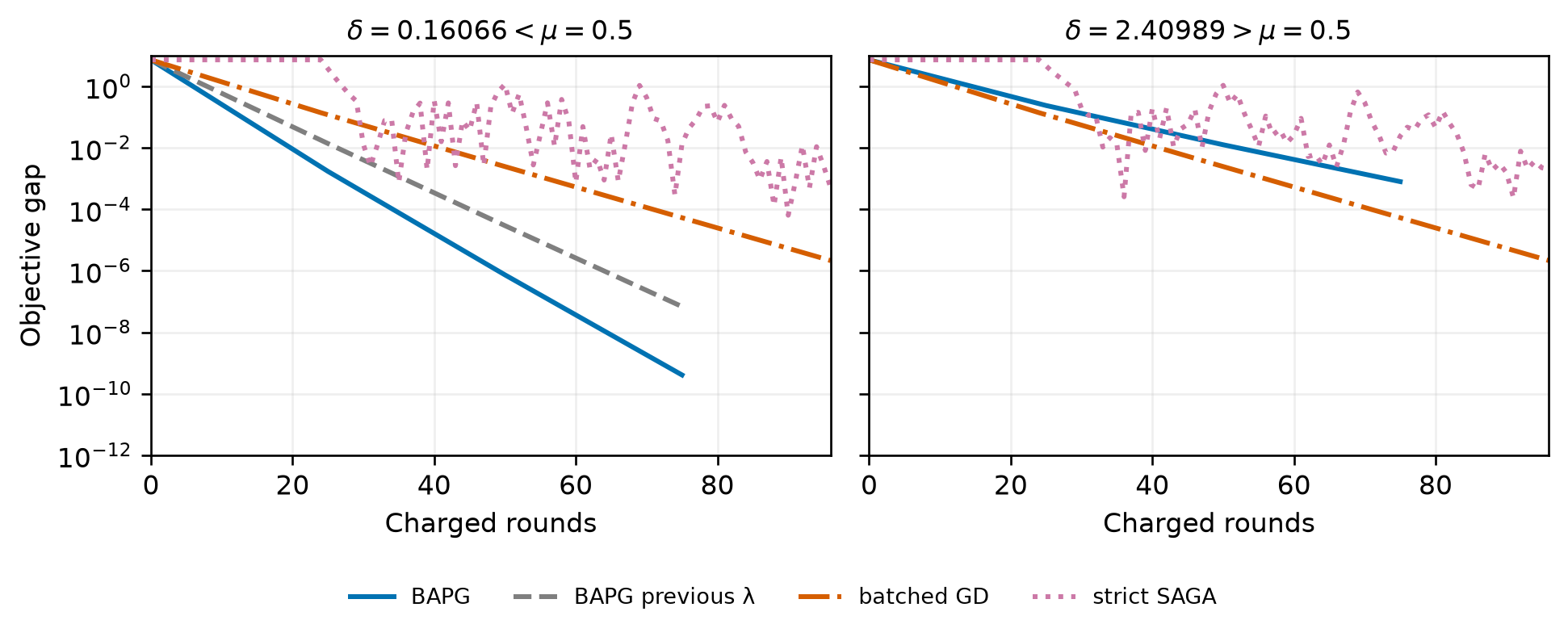}
\caption{Seed-11 charged sweep, reproduced from saved checkpoints.
The left panel uses strict rounds and the right panel client contacts;
both axes are needed because one BAPG update contains a full pass plus
an anchor prox. The diagnostic objective evaluator supplies plotted
gaps only. These curves are finite-dimensional synthetic comparisons,
not a minimax rate plot.}\label{fig:appendix-charged}
\end{figure*}

\section{Thirty-seed quadratic sensitivity and uncertainty}\label{app:replicates}

\subsection{Paired design and curvature calibration}\label{app:replicate-design}
Seeds 100--129 each generate one $m=24,d=3$ quadratic family. Within a
seed the perturbation directions, optimum, and centered linear offsets
are held fixed while the scale is changed from $0.4$ to $6$; each
resulting instance is then run with $s=1$ and $s=4$ and all four
methods. Thus the 480 rows represent
$30\times2\times2\times4$ method endpoints, not 480 independent
problem draws. A success is evaluated \emph{at the last saved
checkpoint} under the 96-round cap and requires
$\Delta_{\rm last}/\Delta_0\leq10^{-4}$. An earlier crossing is not
credited. For the budget trajectories, the latest saved iterate is
carried forward until the next method checkpoint. The objective evaluator
does not affect these iterates.

The proof's $\mu$ must lower-bound \emph{every} client Hessian. For each
seed the saved `mu\_min\_local' is the minimum eigenvalue across all
24 local Hessians, the largest valid common constant for that
quadratic instance. It is not the global minimum eigenvalue, which is
10.5 by construction at both scales. Recomputing
$q=2\delta/(\mu+2\delta)$ using this realized clientwise value
calibrates the theorem but leaves every experimental trajectory
unchanged. Table~\ref{tab:appendix-curvature-range} exposes variation
suppressed by the main-text medians. In particular the public
$\mu=0.5$ is valid but makes the reported $\delta/\mu$ and $q$
far less informative about the actual instances.

\begin{table*}[t]
\centering\small
\caption{Across-seed curvature diagnostics, shown as median [minimum,
maximum]. Each quantity is computed on the 30 paired quadratic
families; $q_{\rm local}$ inserts the minimum client eigenvalue into
the proved BAPG factor at $\lambda=\delta$.}
\label{tab:appendix-curvature-range}
\begin{adjustbox}{max width=\textwidth}
\begin{tabular}{rllll}
\toprule
Scale & Anchor $\delta$ & Min-client $\mu$ & $\delta/\mu_{\rm local}$ & $q_{\rm local}$\\
\midrule
0.4 & 0.2314 [0.1267, 0.3702] & 10.2343 [10.1549, 10.3185] & 0.02261 [0.01246, 0.03622] & 0.04327 [0.02431, 0.06755]\\
6 & 3.4710 [1.9004, 5.5527] & 5.6291 [4.8032, 6.6357] & 0.62197 [0.35066, 1.12225] & 0.55432 [0.41222, 0.69179]\\
\bottomrule
\end{tabular}
\end{adjustbox}
\end{table*}

\begin{figure*}[t]
\centering
\includegraphics[width=0.96\textwidth,keepaspectratio]{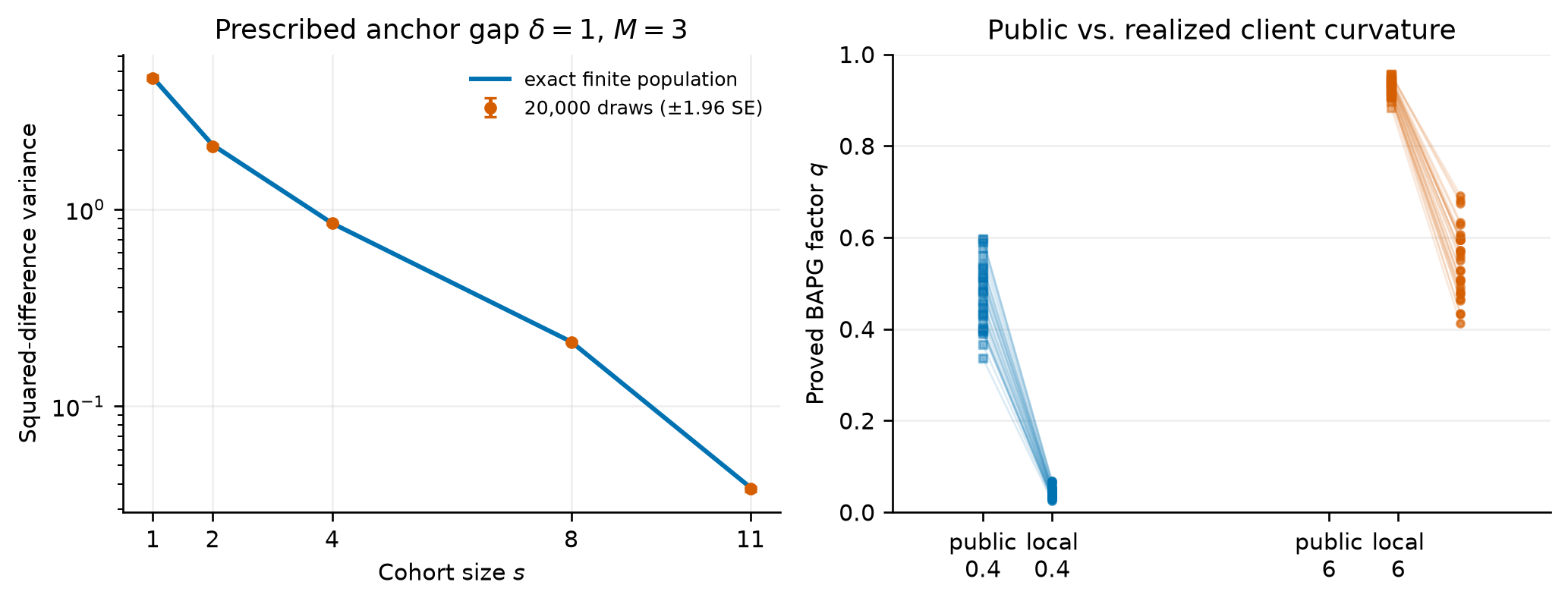}
\caption{Two independent saved-data checks. Left: the five
positive-$\delta$ variance cells compare exact enumeration with the
20,000-draw mean and unadjusted Monte Carlo intervals. Right: for
each of 30 paired quadratic seeds, the proved BAPG factor at the
public $\mu=0.5$ is connected to the same factor recalculated with
that instance's minimum client eigenvalue. Recalibration changes no
iterate. These panels check a finite-population identity and an input
sensitivity; neither is an optimization lower bound.}
\label{fig:appendix-evidence}
\end{figure*}

\begin{table*}[t]
\centering\small
\caption{Final-checkpoint quadratic outcomes over 30 seeds per cell.
Each entry is successes / 30 followed by the median final relative
gap in parentheses. The fixed target is $10^{-4}$ of the same run's
initial gap. Numerical zeros indicate the evaluator's floor and do
not rank methods below floating-point precision.}
\label{tab:appendix-replicate-endpoints}
\begin{adjustbox}{max width=\textwidth}
\begin{tabular}{rrllll}
\toprule
Scale & $s$ & BAPG & Batched GD & Strict SAGA & Quadratic identification\\
\midrule
0.4 & 1 & 30/30 ($3.18\cdot10^{-10}$) & 30/30 ($4.62\cdot10^{-8}$) & 1/30 ($2.80\cdot10^{-3}$) & 30/30 (0)\\
0.4 & 4 & 30/30 (0) & 30/30 (0) & 30/30 ($3.00\cdot10^{-11}$) & 30/30 (0)\\
6 & 1 & 5/30 ($4.11\cdot10^{-4}$) & 30/30 ($4.62\cdot10^{-8}$) & 0/30 ($2.70\cdot10^{-3}$) & 30/30 (0)\\
6 & 4 & 30/30 ($9.11\cdot10^{-14}$) & 30/30 (0) & 30/30 ($8.53\cdot10^{-12}$) & 30/30 (0)\\
\bottomrule
\end{tabular}
\end{adjustbox}
\end{table*}

For every $s=1$ quadratic row, BAPG's final cost is
$(75,75,150)$, GD's and SAGA's are $(96,96,192)$, and
identification's is $(96,96,192)$ in the order
(rounds, contacts, vectors). At $s=4$ the respective costs are
$(91,325,650)$, $(96,384,768)$, $(96,384,768)$, and
$(24,96,192)$. Identification solves the known quadratic class at its
last probe and does not yield a general smooth-function comparator.
The striking scale-6, $s=1$ BAPG/GD difference is an outcome of this
well-conditioned generator and these budgets. The BAPG contraction
bound is an upper guarantee, not a promise of empirical dominance over
GD or SAGA.

The empirical spread is also material. Table~\ref{tab:appendix-iqr}
reports Hazen quartiles of the final relative gap, rather than only
medians or success counts. At scale 6 and $s=1$, BAPG's lower quartile
is already above the $10^{-4}$ target, consistent with only five
endpoint successes. At $s=4$, several quartiles sit at the
diagnostic evaluator's numerical floor; those tiny values cannot
support a precise ordering of converged methods. The GD distribution
at a fixed $s$ is virtually identical across scales because the
quadratic generator holds the average Hessian and optimum fixed and
GD consumes only the full gradient. The $s=1$ SAGA distribution is
much wider than its single success count alone conveys.

\begin{table*}[t]
\centering\small
\caption{Hazen interquartile ranges of final relative gaps across the
30 paired quadratic seeds. These describe generated instances and
the specified round cap, not uncertainty in a minimax law.}
\label{tab:appendix-iqr}
\begin{tabular}{rrlll}
\toprule
Scale & $s$ & BAPG IQR & GD IQR & SAGA IQR\\
\midrule
0.4 & 1 & $[1.12\cdot10^{-10},1.57\cdot10^{-9}]$ & $[5.99\cdot10^{-9},2.42\cdot10^{-7}]$ & $[1.64\cdot10^{-3},1.12\cdot10^{-2}]$\\
0.4 & 4 & $[0,1.67\cdot10^{-16}]$ & $[0,1.27\cdot10^{-16}]$ & $[4.10\cdot10^{-12},1.58\cdot10^{-10}]$\\
6 & 1 & $[1.99\cdot10^{-4},1.57\cdot10^{-3}]$ & $[5.99\cdot10^{-9},2.42\cdot10^{-7}]$ & $[1.55\cdot10^{-3},8.81\cdot10^{-3}]$\\
6 & 4 & $[2.76\cdot10^{-15},2.94\cdot10^{-12}]$ & $[0,1.67\cdot10^{-16}]$ & $[2.93\cdot10^{-12},4.84\cdot10^{-11}]$\\
\bottomrule
\end{tabular}
\end{table*}

\subsection{Budget trajectories and interval interpretation}\label{app:trajectory-uncertainty}
Table~\ref{tab:appendix-budget-success} reads the saved quadratic
checkpoint arrays at four fixed round budgets. Because the criterion
uses the \emph{current} iterate, not the minimum gap seen so far, a
stochastic method's success count can decrease between budgets. This
occurs for the scale-0.4 SAGA curve: one seed succeeds at round 48,
none do at round 72, and one does at round 96. The figure in the main
text uses every integer budget from 0 to 96 with the same carry-forward
rule, and its step locations come from charged checkpoint rounds.

\begin{table}[t]
\centering\small
\caption{Successes out of 30 at strict round budgets for $s=1$.
The latest available current iterate is used.}
\label{tab:appendix-budget-success}
\begin{tabular}{llrrrr}
\toprule
Scale & Method & 24 & 48 & 72 & 96\\
\midrule
0.4 & BAPG & 0 & 2 & 30 & 30\\
0.4 & GD & 0 & 15 & 30 & 30\\
0.4 & SAGA & 0 & 1 & 0 & 1\\
0.4 & Identification & 0 & 0 & 0 & 30\\
6 & BAPG & 0 & 0 & 1 & 5\\
6 & GD & 0 & 15 & 30 & 30\\
6 & SAGA & 0 & 0 & 0 & 0\\
6 & Identification & 0 & 0 & 0 & 30\\
\bottomrule
\end{tabular}
\end{table}

The main-text 95\% Wilson intervals describe binomial proportions
under the synthetic seed generator: $5/30$ maps to
$[0.073,0.336]$, $0/30$ to $[0,0.114]$, and $30/30$ to
$[0.886,1]$. They are not uncertainty about a theorem or a minimax
constant. The scale and method outcomes are paired within seed, so
separate Wilson intervals should not be used as an independent-samples
test of their differences. The saved data support an empirical
sensitivity description, and no multiplicity-adjusted claim of method
ranking is made.

For transparency, the interval applied to a success count $k$ among
$n=30$ independent generated seeds uses $\hat p=k/n$ and $z=1.96$:
\begin{equation}\label{eq:app-wilson}
 \frac{\hat p+z^2/(2n)}{1+z^2/n}
 \;\pm\;
 \frac{z\sqrt{\hat p(1-\hat p)/n+z^2/(4n^2)}}{1+z^2/n}.
\end{equation}
It is a pointwise descriptive interval, not simultaneous coverage
across all methods, scales, and budgets. The reported Hazen quartiles
use order-statistic rank $h=np+1/2$ with interpolation when $h$ is
nonintegral; for $n=30$ and $p=0.25,0.75$, those ranks are exactly 8
and 23. This makes Table~\ref{tab:appendix-iqr} reproducible from the
saved endpoint rows without choosing an undocumented quantile
convention.

Paired outcomes can be read directly without treating two 30-seed
columns as independent. At scale 6 and $s=1$, all five BAPG successes
are also GD successes, leaving 25 GD-only successes and no BAPG-only
successes. In the same cell SAGA has zero successes, so BAPG has five
BAPG-only successes against SAGA. At scale 0.4 and $s=1$, BAPG and GD
both succeed on every seed, whereas BAPG succeeds without SAGA on 29
seeds. The table makes these particular paired counts explicit; they
remain properties of the generated distribution, round cap, and
comparator tuning. They do not imply a population-wide ranking for
other dimensions, client functions, or information budgets.

\begin{table}[t]
\centering\small
\caption{Paired endpoint success discordance at $s=1$ across the 30
quadratic seeds. ``B only'' and ``other only'' count seeds where one
method reaches the $10^{-4}$ target and the other does not.}
\label{tab:appendix-discordance}
\begin{tabular}{rrlrr}
\toprule
Scale & $s$ & Comparison & B only & Other only\\
\midrule
0.4 & 1 & BAPG vs GD & 0 & 0\\
0.4 & 1 & BAPG vs SAGA & 29 & 0\\
6 & 1 & BAPG vs GD & 0 & 25\\
6 & 1 & BAPG vs SAGA & 5 & 0\\
\bottomrule
\end{tabular}
\end{table}

\section{Logistic control and tuning correction}\label{app:logistic}

\subsection{A certified nonlinear anchor bound}\label{app:logistic-certificate}
Each logistic client has $n=30$ labeled observations $(X_{ij},y_{ij})$
and objective
\begin{align*}
 f_i(x)&=\frac1n\sum_{j=1}^n
 \log(1+\exp(-y_{ij}X_{ij}^{\top}x))\\
 &\quad+\frac\mu2\|x\|^2,\qquad\mu=0.5.
\end{align*}
Writing $G_i=X_i^\top X_i/n$, its Hessian is
$\mu I+X_i^\top W_i(x)X_i/n$ with diagonal weights between zero
and $1/4$. Hence all local objectives are $\mu$-strongly convex,
and a valid uniform certificate for designated anchor zero is
\begin{equation}\label{eq:app-logistic-delta}
 \delta_{\rm cert}
 =\frac14\{\|G_0\|_{\rm op}+\|m^{-1}\sum_iG_i\|_{\rm op}\}.
\end{equation}
The triangle inequality bounds the anchor-minus-average Hessian by
\eqref{eq:app-logistic-delta} for every $x$. It is conservative because
it ignores cancellation and the actual logistic weights. It is the
certificate passed to BAPG, not an estimated exact supremum. The
regularization lower bound $\mu=0.5$ is sharp as a uniform bound in
this model because logistic weights tend to zero along separating
directions. The 30 separate logistic datasets use seeds 5100--5129,
$m=12,d=3,s=2$, and the same 96-round cap. Their anchor-certificate
median is 0.6471, with range $[0.5173,0.8108]$.

\begin{table*}[t]
\centering\small
\caption{Regularized-logistic sensitivity across 30 datasets. All
three methods succeed at final relative gap $\leq10^{-4}$; the
quartiles are Hazen empirical quartiles of the \emph{relative} gaps.
Absolute objective gaps at numerical zero are not ordered.}
\label{tab:appendix-logistic}
\begin{adjustbox}{max width=\textwidth}
\begin{tabular}{lrrrr}
\toprule
Method & Successes & Median relative gap & Relative-gap IQR & Last rounds / contacts / vectors\\
\midrule
BAPG & 30/30 & $3.34\cdot10^{-9}$ & $[1.59\cdot10^{-9},1.74\cdot10^{-8}]$ & 91 / 169 / 338\\
Batched GD & 30/30 & 0 & $[0,0]$ & 96 / 192 / 384\\
Strict SAGA & 30/30 & $9.91\cdot10^{-11}$ & $[1.99\cdot10^{-11},5.87\cdot10^{-10}]$ & 96 / 192 / 384\\
\bottomrule
\end{tabular}
\end{adjustbox}
\end{table*}

The logistic anchor prox is computed by damped Newton iteration and
accepts a reply only after its local model-gradient residual certifies
model gap at most $10^{-10}$. This is the inexact-solve mechanism of
\eqref{eq:app-inexact}. The diagnostic global optimum is also found
numerically, so final differences near zero should not be interpreted
as exact performance ranks. This panel checks that the charged
implementation can run on a nonlinear objective; it cannot identify
an asymptotic rate from 30 small $d=3$ datasets.

\subsection{Why the saved prior-tuning runs are separate}\label{app:tuning}
An earlier saved version supplied SAGA the \emph{global} bound
$L_{\rm global}$, although the implementation's fixed step divisor
uses a bound on every local gradient. The active run instead supplies
$L_{\rm local}=\max_iL_i$. In the quadratic family the global bound
is 12.5 for all scales and seeds, while median local bounds are
12.7515 at scale 0.4 and 17.5806 at scale 6. The earlier SAGA
step $1/(3L_{\rm global})$ was therefore larger than the corrected
$1/(3L_{\rm local})$. The prior run also used a maximum-over-clients
logistic similarity certificate rather than the selected anchor's
certificate in \eqref{eq:app-logistic-delta}. Those two changes are
recorded in distinct prior-tuning JSONs rather than silently
merged into the active data.

\begin{table}[t]
\centering\small
\caption{SAGA success counts before and after correcting the supplied
smoothness bound, at the same 30 quadratic seeds and final target.
The prior rows are an audit trail, not the active comparator.}
\label{tab:appendix-tuning}
\begin{tabular}{rrcc}
\toprule
Scale & $s$ & Prior global $L$ & Active local $L$\\
\midrule
0.4 & 1 & 1/30 & 1/30\\
0.4 & 4 & 30/30 & 30/30\\
6 & 1 & 2/30 & 0/30\\
6 & 4 & 30/30 & 30/30\\
\bottomrule
\end{tabular}
\end{table}

The change in the scale-6, $s=1$ success count illustrates that even
when both curves are charged correctly, comparator tuning changes
outcomes. The corrected SAGA run should not be used to claim a
relative theorem without matching method assumptions and tuning
information. In the logistic prior file, the median all-client
certificate was 0.7656; the active anchor-only certificate median is
0.6471. This changes BAPG's parameter and trajectory, so its prior
and active logistic endpoints are also separate sensitivity runs.

\section{Read-only audit of the original P11 study}\label{app:historical}

\subsection{What carries across the change of oracle}\label{app:historical-theory}
The original project's manuscript and appendix used a different pair of
assumptions: every client Hessian, rather than one designated anchor
Hessian, was required to stay within one common similarity bound of the
global Hessian; one nominal communication iteration could collect an
exact full gradient, perform a local prox without a separately charged
contact, and in some algorithms exchange more than one returned vector
with a sampled client. Its statements about APP-SVRS, acceleration,
zero-chain lower bounds, and sampled floors were formulated against that
interface and are not theorems of the strict interface studied here.
The original paper also used a global smoothness bound $L$, whereas
our BAPG theorem does not require one for its contraction. Consequently
a fitted historical slope against its clientwise similarity parameter
cannot be relabeled as a strict-round slope against the current
single-anchor $\delta$.

One valid calculation does transfer conditionally. Under the original
\emph{all-client} bound $D$, the integral Hessian argument controls
every centered gradient displacement by $D\|x-w\|$. The exact
finite-population identity \eqref{eq:app-fpc} then yields an upper bound
$\xi(s)D^2\|x-w\|^2$ for the fresh control-variate difference. That is
a legitimate structural lemma under its stronger assumption. The
construction in Theorem 3 makes the failure of the same inference under
our anchor-only bound explicit. The original paper's idea of comparing
participation rounds and client work also survives as an accounting
question, but its numerical frontiers must be recomputed with all
full passes, local-prox contacts, and vector stages charged.

Table~\ref{tab:appendix-proof-transfer} separates reusable mathematical
facts from conclusions that require rederivation under a different
oracle. This matters especially for acceleration: a generic Catalyst
outer-loop argument can be a valid tool
\citep{lin2018catalyst}, but the claimed total complexity still depends
on an inner method satisfying its own cost and accuracy assumptions.
Likewise, Chebyshev polynomial acceleration is a valid algebraic
mechanism for a suitable self-adjoint affine map; it does not by
itself establish a strict-oracle complexity, and it cannot rule out
the finite quadratic identification method. The original manuscript's
zero-chain lower theorem also restricted algorithm iterates to a
received-gradient span. Our oracle allows a nonlinear exact local-prox
reply, so a span-restricted theorem is not automatically a lower bound
against it.

\begin{table*}[t]
\centering\small
\caption{Theory transfer audit for the original P11 manuscript and
appendix. ``Reusable'' means the stated conditional mathematical step
can be used after its assumptions are restated; it does not validate
the original end-to-end rate under the active strict oracle.}
\label{tab:appendix-proof-transfer}
\begin{tabular}{p{0.23\textwidth}p{0.34\textwidth}p{0.34\textwidth}}
\toprule
Original ingredient & Conditional fact retained & Missing step under the current oracle\\
\midrule
Hessian integral and finite-population lemma & Under a bound for every client, centered gradient differences are controlled and $\xi(s)$ is exact. & One-anchor $\delta$ does not control the other clients; Theorem 3 supplies a counterfamily.\\
Implicit sampled proximal method & Its original nominal-iteration transcript is part of the historical method record. & Charge complete anchor refreshes, the prox reply, and any multi-vector exchange; reprove its rate with matching assumptions.\\
Catalyst and Chebyshev arguments & Generic acceleration mechanisms can apply to specified inner maps or linear operators. & A valid strict-oracle inner rate and a costed implementable map are not established by the old proof.\\
Zero-chain lower-bound construction & The original span restriction identifies a limited algorithm class. & Local-prox replies and fixed-dimensional exact identification lie outside the claimed universal inference.\\
Sampled-anchor variance and floor & The finite-population anchor-gradient variance identity is exact. & A method-specific noise term is not a universal optimization floor over all strict-oracle algorithms.\\
\bottomrule
\end{tabular}
\end{table*}

The original related-work trail included DANE and ProxSkip, along with
Catalyst \citep{lin2018catalyst} and the distributed lower-bound
lineage \citep{arjevani2015communication}, alongside the similarity
papers already discussed in the main text. These references motivate
specific methods or proof styles; none supplies a strict-oracle
positive-$\delta$ upper/lower match by citation alone. The current
paper keeps the four directly comparable similarity references in its
main discussion and records the older trail here with that scope.

\subsection{Disposition of all six saved figure families}\label{app:historical-figures}
The original repository retains quick CPU, full CPU, and large GPU
JSONs and the corresponding plots. The large GPU file records
$m=40,d=120$, 16 grid values, 64 Monte Carlo replicates per primary
grid point, a 1,600 nominal-iteration budget, and a relative target
$10^{-5}$; its metadata report a Tesla T4 run. The full CPU file
uses $m=20,d=60$ and 600 nominal iterations; the quick CPU file uses
$m=12,d=40$ and 300. Table~\ref{tab:appendix-history-map}
identifies every original figure family and the inference it may
support after the oracle audit. The images reproduced below are
unmodified historical plots, so their embedded labels such as
``rounds,'' ``theory,'' and ``lower bound'' retain their \emph{original}
meaning only. The captions here, rather than their original captions,
govern their interpretation in this manuscript.

The three saved run sizes also show which descriptive patterns
persisted as the historical simulation grew. Table~\ref{tab:appendix-history-runs}
reads the saved fit fields directly; a dash denotes an absent or
nonfinite fit, not a zero slope. The acceleration-like slope stayed
near one half in the quadratic nominal-iteration plot, the
uncorrected sampled floor exponent stayed near one, and the
best-of-named zero-chain slope stayed near 0.45. Repetition of these
descriptive patterns is useful for understanding the old generator,
but none changes the missing strict-round charges or the all-client
versus one-anchor assumption. The quick and full CPU runs have a
different relative target and dimensional design from the large GPU
run; they are not a pooled statistical sample.

\begin{table*}[t]
\centering\small
\caption{Saved historical fits across run sizes. ``Accel.'' and
``unaccel.'' are fitted exponents in the original nominal iteration
metric; ``floor'' is an empirical fit against $\xi(s)$ and
``zero-chain'' is best-of-named-methods. These values are reported for
provenance, not as validation of the active theorem.}
\label{tab:appendix-history-runs}
\begin{tabular}{lrrrrrr}
\toprule
Run ($m,d$, cap) & Quad. a. & Quad. u. & Floor & Chain & Log. a. & Log. u.\\
\midrule
Quick CPU (12,40,300) & 0.456 & 0.970 & 1.095 & 0.422 & 0.377 & ---\\
Full CPU (20,60,600) & 0.474 & 0.922 & 1.007 & 0.452 & 0.451 & 0.829\\
Large GPU (40,120,1600) & 0.459 & 0.938 & 1.005 & 0.451 & 0.526 & 0.787\\
\bottomrule
\end{tabular}
\end{table*}

\begin{table*}[t]
\centering\small
\caption{Complete audit of the six historical figure families. Numeric
summaries are descriptive values saved in the original JSONs. All are
excluded from strict-oracle theorem validation.}
\label{tab:appendix-history-map}
\begin{tabular}{p{0.18\textwidth}p{0.35\textwidth}p{0.38\textwidth}}
\toprule
Figure family & Saved observation & Strict-oracle disposition\\
\midrule
Rounds versus dissimilarity & Large GPU fitted slopes: accelerated 0.459; unaccelerated 0.938. & Each nominal sliding iteration has an exact full gradient and uncharged anchor prox; the dissimilarity assumption is all-client, not anchor-only.\\
Convergence curves & Named methods have distinct gap-versus-nominal-iteration traces at fixed similarity. & Descriptive trajectories of those implementations; horizontal axis is not charged strict rounds.\\
Rounds versus participation & Full CPU control-variate count 35.0 and large GPU count 42.36 across $s$. & Code sets reference $w=y$, so sampled difference is zero and the estimator is the exact full gradient; this does not test sampled participation.\\
Sampled floor versus participation & Large GPU fitted exponent 1.005 against $\xi(s)$ for an uncorrected sampled method. & Generator-specific observed floor; it is not a universal strict-oracle floor or a bound from anchor-only $\delta$.\\
Quadratic zero-chain & Large GPU best-of-named-methods fitted slope 0.451. & Pointwise minimum over a few named methods on fixed-dimensional quadratics, not a lower bound over arbitrary adaptive algorithms with local-prox replies.\\
Logistic rounds versus dissimilarity & Large GPU accelerated and unaccelerated fitted slopes 0.526 and 0.787. & Nonlinear historical control, still using exact full gradients and uncharged prox under a different round unit.\\
\bottomrule
\end{tabular}
\end{table*}

The current redesign's quadratic identification proposition is a
concrete counterpoint to interpreting the old zero-chain figure as a
universal accuracy-dependent lower bound: at fixed $d$ this oracle
recovers the entire average quadratic from $d+1$ complete gradient
probes. The original sampled-floor figure similarly cannot establish a
universal optimization floor: changing its method, keeping a gradient
table, or paying for a complete pass changes the information available
to the optimizer. These are limits on inference, not allegations that
the original plotted values were fabricated. The original JSONs and
figures remain available for independent reanalysis.

The participation rows of the large-GPU file permit a more direct
oracle check. Its control-variate estimator uses a reference equal to
the current point, so for every cohort $S$,
\[
 g_S(y;y)=\nabla f(y)+\frac1s\sum_{i\in S}
 \{\nabla f_i(y)-\nabla f_i(y)\}=\nabla f(y).
\]
Thus the sampled correction cancels algebraically before any
simulation takes place. The resulting saved nominal iteration count
is 42.359375 at every selected $s$, with reported standard deviation
2.203457 across the GPU replicates. The uncorrected sampled method's
floor changes strongly with $s$, as Table~\ref{tab:appendix-history-participation}
shows, but that is a different estimator and a generator-specific
outcome. The last column gives the \emph{direct} strict-round cost of
one full-gradient pass followed by one anchor prox contact for $m=40$,
namely $\lceil40/s\rceil+1$. It is a ledger illustration, not a
retroactive corrected convergence rate for the old algorithms:
their original stopping logic, references, assumptions, and
multi-stage replies were not rerun in this interface.

\begin{table*}[t]
\centering\small
\caption{Selected large-GPU historical participation cells from the
saved original JSON. ``CV nominal'' is the original control-variate
iteration count; the floor is for a different, uncorrected sampled
method. The last column charges only the direct full-gradient-plus-prox
transcription in the current strict interface and must not be
identified with the original x-axis.}
\label{tab:appendix-history-participation}
\begin{tabular}{rrrrr}
\toprule
$s$ & $\xi(s)$ & CV nominal & Sampled floor & Direct strict $R$/step\\
\midrule
1 & 1.00000 & 42.359 & 172.836 & 41\\
2 & 0.48718 & 42.359 & 82.106 & 21\\
4 & 0.23077 & 42.359 & 38.288 & 11\\
8 & 0.10256 & 42.359 & 17.248 & 6\\
16 & 0.03846 & 42.359 & 6.506 & 4\\
24 & 0.01709 & 42.359 & 2.884 & 3\\
32 & 0.00641 & 42.359 & 1.055 & 3\\
40 & 0 & 42.359 & $6.38\cdot10^{-10}$ & 2\\
\bottomrule
\end{tabular}
\end{table*}

\begin{figure*}[htbp]
\centering
\begin{minipage}{0.48\textwidth}\centering
\includegraphics[width=\linewidth]{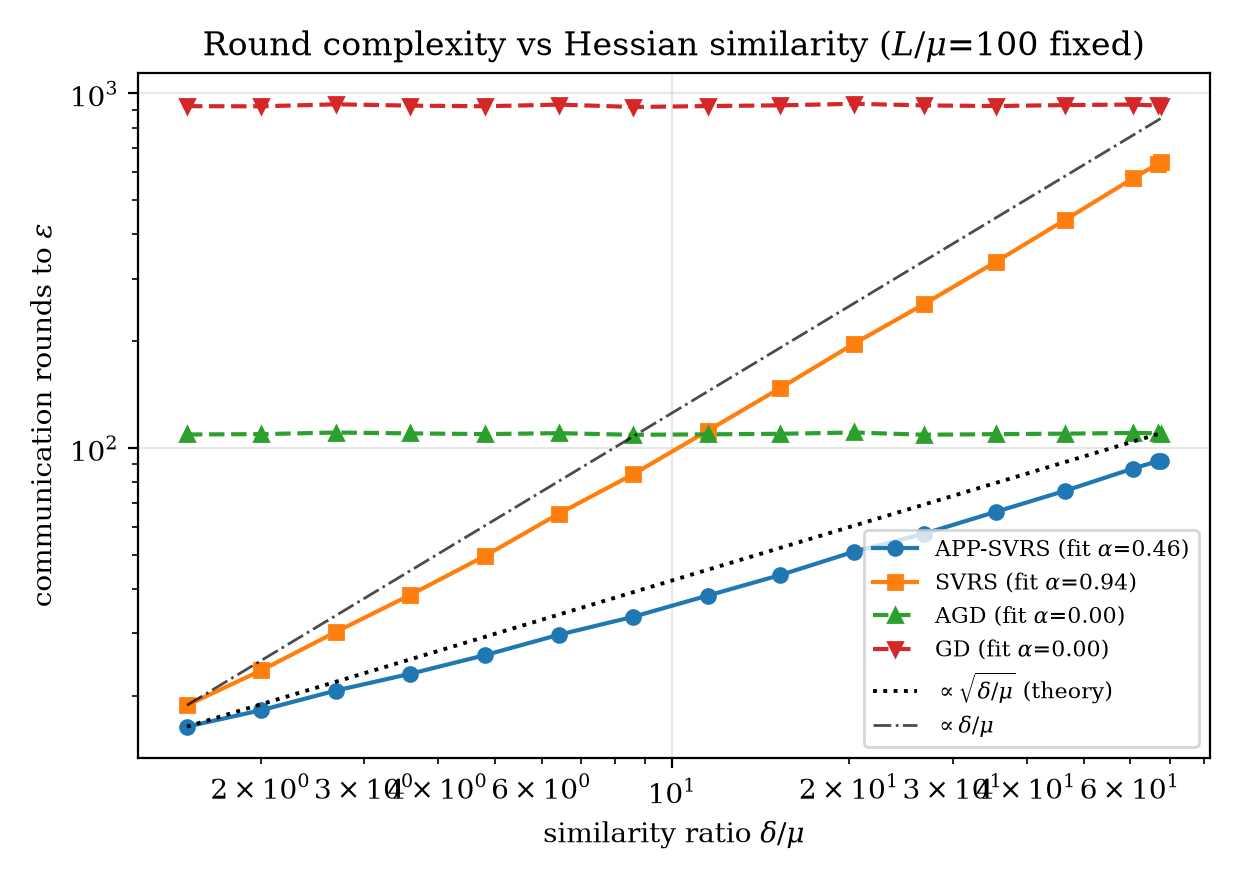}
\small (a) Rounds versus dissimilarity
\end{minipage}\hfill
\begin{minipage}{0.48\textwidth}\centering
\includegraphics[width=\linewidth]{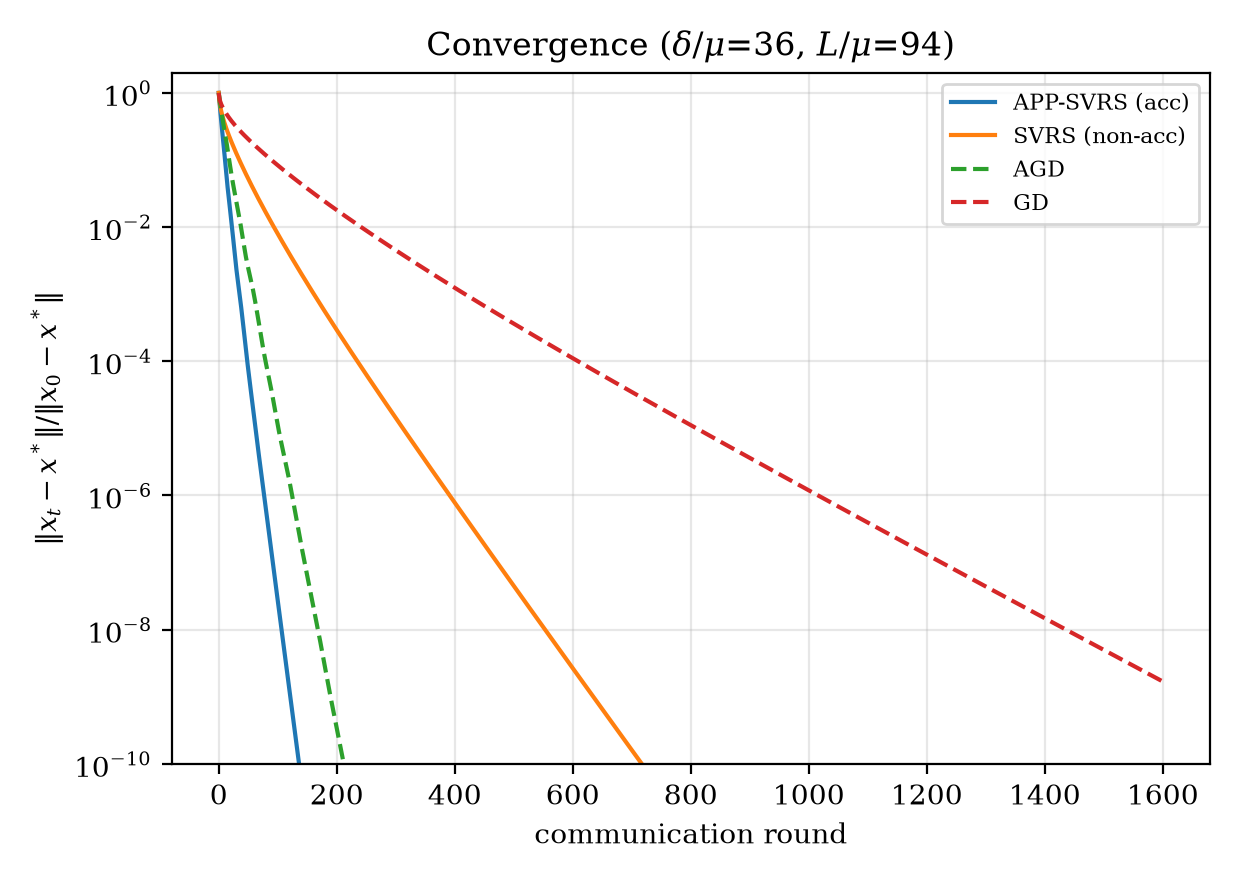}
\small (b) Convergence curves
\end{minipage}
\caption{Historical large-GPU figure families 1 and 2, reproduced
read-only. The nominal x-axis charges neither the exact full gradient
as $\lceil m/s\rceil$ strict rounds nor the subsequent anchor prox
contact. The fitted slopes and gap traces describe that original
simulation only; they do not validate the present BAPG bound or a
sampled positive-$\delta$ rate.}\label{fig:appendix-historical-a}
\end{figure*}

\begin{figure*}[htbp]
\centering
\begin{minipage}{0.48\textwidth}\centering
\includegraphics[width=\linewidth]{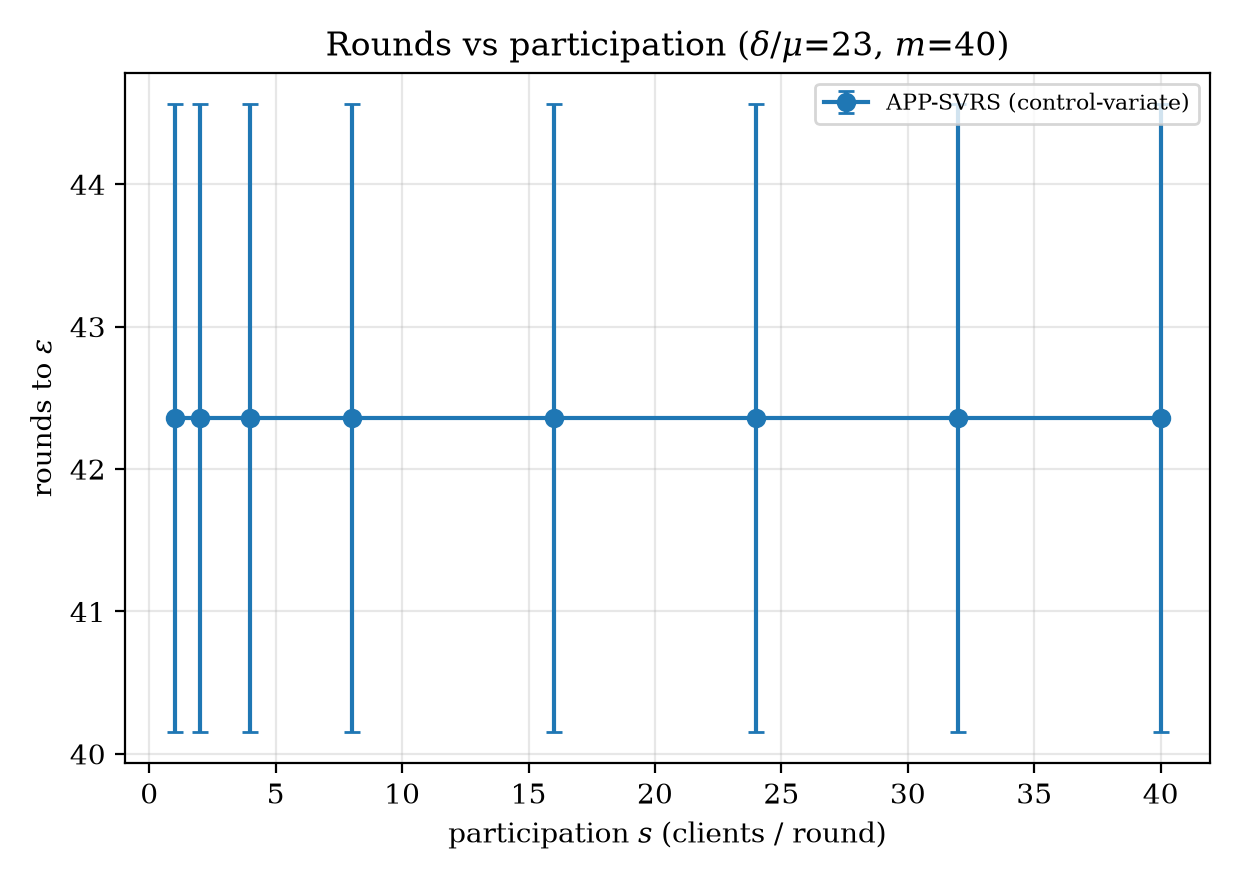}
\small (c) Rounds versus participation
\end{minipage}\hfill
\begin{minipage}{0.48\textwidth}\centering
\includegraphics[width=\linewidth]{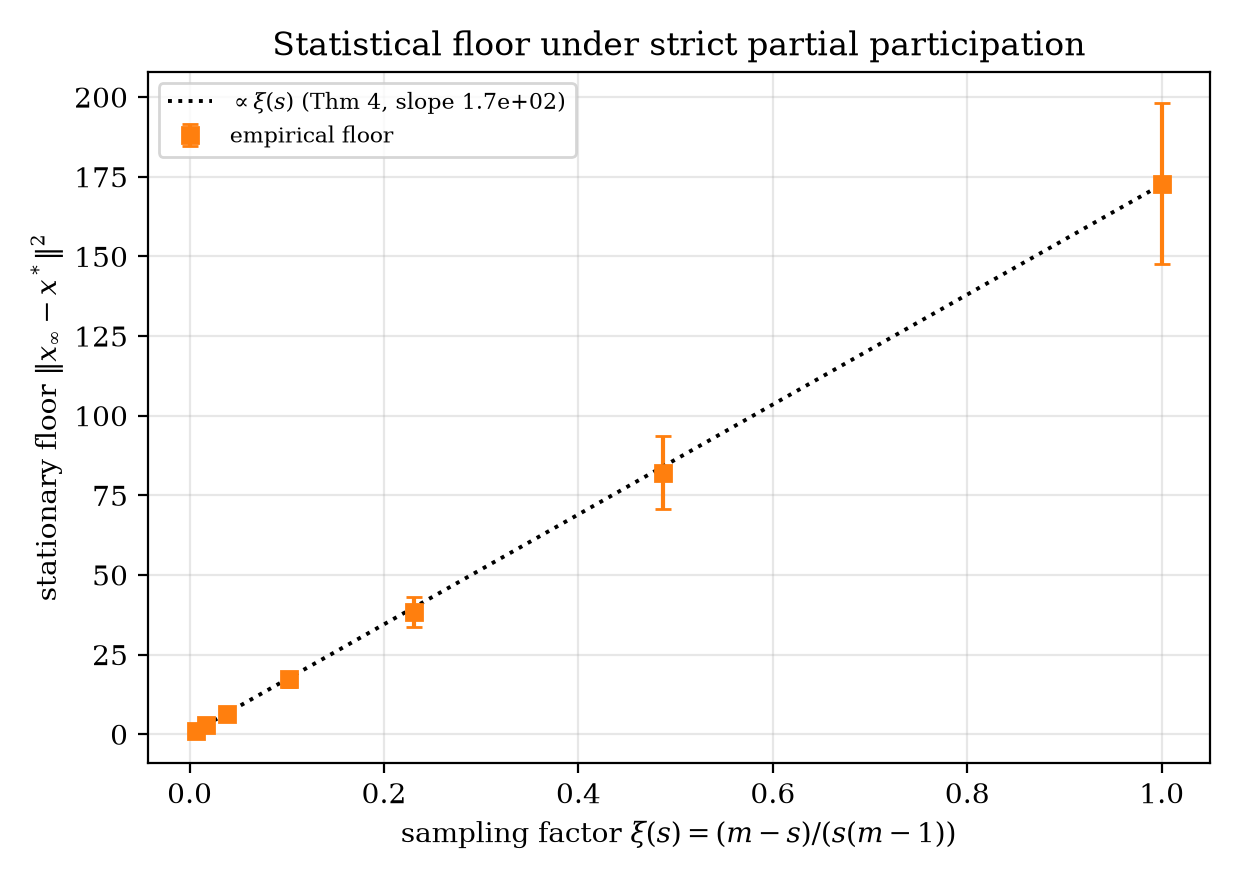}
\small (d) Sampled floor versus participation
\end{minipage}
\caption{Historical large-GPU figure families 3 and 4. In the
control-variate participation simulation, $w=y$ makes the sampled
difference identically zero, so its constant count across $s$ comes
from an exact full gradient. The uncorrected sampled method's floor
is an observation of that generator; it is not a universal strict-oracle
floor theorem.}\label{fig:appendix-historical-b}
\end{figure*}

\begin{figure*}[htbp]
\centering
\includegraphics[width=0.78\textwidth]{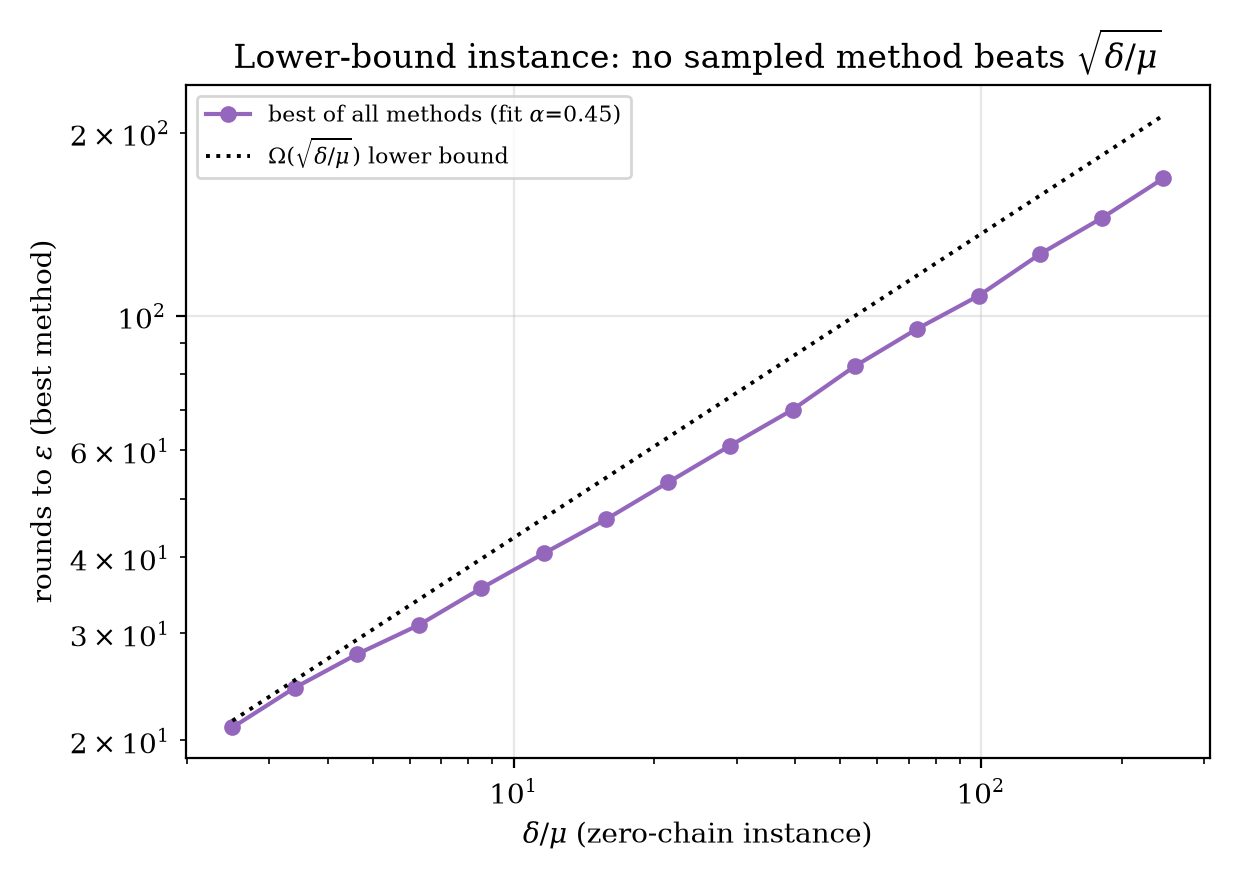}
\caption{Historical large-GPU figure family 5. The plotted
``lower bound'' is the best observed performance among named methods
on a finite-dimensional quadratic zero chain. It does not quantify
over all strict-oracle algorithms. Proposition 5 supplies a charged
exact-identification method for fixed-dimensional quadratics, so the
plot cannot certify an accuracy-dependent minimax lower bound here.}
\label{fig:appendix-historical-c}
\end{figure*}

\begin{figure*}[htbp]
\centering
\includegraphics[width=\textwidth]{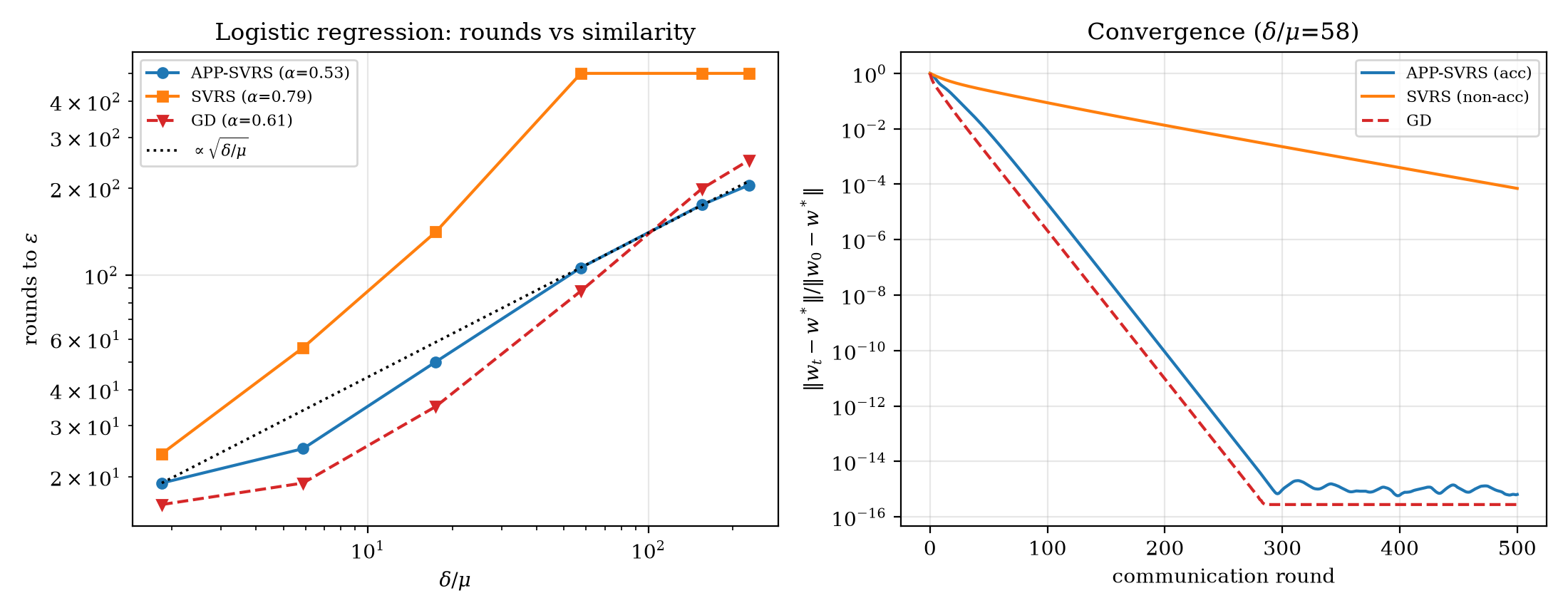}
\caption{Historical large-GPU figure family 6. The logistic fitted
slopes are descriptive of the old full-gradient and uncharged-prox
iteration convention and are distinct from the current 30-seed charged
logistic sensitivity panel. The horizontal quantity must not be read
as a strict-round cost under the anchor-only condition.}
\label{fig:appendix-historical-d}
\end{figure*}

%% file: positive_chain.tex
\section{Positive-dissimilarity quadratics under a zero-respecting oracle}
\label{app:zr-chain}

\ifjmlrlayout\else
\begin{figure}[t]
\input{hidden_chain_figure}
\end{figure}
\fi

\subsection{Oracle and hard family}
The lower bound in this section adds a restriction to the strict protocol
of Section~\ref{strict-participation-model-and-accounting}. The server
starts with linear memory $\mathcal M_0=\{0\}$. Every vector sent to a
client and the final output must lie in the span of earlier sent vectors
and oracle replies. For quadratic
$f_i(x)=\tfrac12x^\top A_ix-b_i^\top x$, each reply must be exactly one
requested gradient $A_i u-b_i$, Hessian-vector product $A_i v$,
shifted inverse $(A_i+\gamma I)^{-1}v$, or proximal point
$(A_i+\gamma I)^{-1}(b_i+\gamma v)$, with $\gamma>0$ and sent vectors
in current memory. Client identities, operations, and coefficients can
depend arbitrarily on the transcript. The restriction forbids arbitrary
encodings of local objectives. We call this the zero-respecting local
oracle (ZR); the lower bound is not for every one-vector protocol.

Fix $\mu,\delta>0$, $m\geq3$, $d\geq2$, and define
\begin{equation}\label{eq:chain-parameters}
 \kappa=1+\delta/\mu,\quad
 q=\frac{\sqrt\kappa-1}{\sqrt\kappa+1},\quad
 \alpha=\delta/4,\quad c=m\alpha.
\end{equation}
Rationalizing gives $\mu/\alpha=(1-q)^2/q$. With
$v_\ell=e_\ell-e_{\ell+1}$, set
\begin{equation}\label{eq:chain-objective}
 Q(x)=x_1^2+\sum_{\ell=1}^{d-1}(v_\ell^\top x)^2
                   +(1-q)x_d^2,\qquad
 f(x)=\tfrac\mu2\|x\|^2+\tfrac\alpha2 Q(x)-\alpha x_1.
\end{equation}
Every client has the base $\mu\|x\|^2/2$. The anchor receives the
linear term $-cx_1$, a fixed non-anchor client $c_0$ receives the
boundary terms $c\{x_1^2+(1-q)x_d^2\}/2$, and a holder $h_\ell$
receives $c(v_\ell^\top x)^2/2$. Draw $h_1$ uniformly from the
$m-1$ non-anchor clients, then draw each $h_\ell$ uniformly among
the $m-2$ non-anchor clients other than $h_{\ell-1}$. A client may
hold several links, but never adjacent links.

Averaging the local objectives gives \eqref{eq:chain-objective}; every
local Hessian is at least $\mu I$. The matrix
$L=\tfrac12\nabla^2Q$ is positive semidefinite and has norm at most
four: for any $u$, bounding each
$(u_\ell-u_{\ell+1})^2$ by $2u_\ell^2+2u_{\ell+1}^2$ leaves
coefficient at most four on each coordinate, including the boundary
terms. Since the anchor Hessian is $\mu I$,
$\|\nabla^2 f-\nabla^2 f_a\|_{\rm op}=\alpha\|L\|_{\rm op}
\leq4\alpha=\delta$.
This is an upper bound on the realized deviation.

The first, interior, and last stationarity rows of
\eqref{eq:chain-objective} at $x_j=q^j$ are respectively
\[
 \mu q+\alpha(2q-q^2)-\alpha=0,\quad
 \mu q^j+\alpha(2q^j-q^{j-1}-q^{j+1})=0,\quad
 \mu q^d+\alpha((2-q)q^d-q^{d-1})=0.
\]
Each equality follows from $\mu/\alpha=(1-q)^2/q$.
Strong convexity makes this stationary point unique. The quadratic
optimum identity therefore gives
\begin{equation}\label{eq:chain-optimum}
 x_j^\star=q^j,\qquad
 \Delta_0=f(0)-f(x^\star)=\alpha q/2.
\end{equation}
For $V_J=\operatorname{span}(e_1,\ldots,e_J)$, $J<d$, every
$x\in V_J$ misses coordinate $J+1$ of $x^\star$. Hence
\begin{equation}\label{eq:chain-gap}
 \frac{f(x)-f(x^\star)}{\Delta_0}
 \geq\frac{\mu q^{2J+2}/2}{\alpha q/2}
 =(1-q)^2q^{2J}.
\end{equation}
This simple bound need not equal the best gap over $V_J$.

\subsection{Support closure and adaptive discovery}
For a fixed client, its link Hessian terms are disjoint two-coordinate
blocks. Boundary diagonal terms do not connect them. Across the cut
$V_J\oplus V_J^\perp$, only link $J$ can cross. If client $i$ does
not hold that link, its Hessian and shifted inverse preserve $V_J$.
If it does, it holds neither adjacent link, so the crossing block ends
at $J+1$ and both operators map $V_J$ into $V_{J+1}$. The linear
vector is nonzero only at the anchor and lies in $V_1$, so gradients
and local proximal replies have the same closure. Starting from
$V_0$, only an anchor gradient or prox can reveal $e_1$. Since all
queries in a round precede its replies, the frontier advances at most
one coordinate per round.

At the moment frontier $J\geq1$ is first reached, every earlier query
was supported on $V_{J-1}$ or less. Link $J$ and later links form
blocks disconnected from this span, so their owners have not affected
any earlier reply. Conditional on the transcript at that moment,
$h_J$ is uniform among $m-1$ non-anchor clients if $J=1$, or among
$m-2$ non-anchor clients excluding the revealed $h_{J-1}$ if $J>1$.
The same statement holds if the eventual holder was previously
queried about an earlier, nonadjacent link: the unseen block did not
change that reply.

Fix this conditional history and follow the continuation in which no
queried client holds link $J$. It tests at most $js$ identities during
the next $j$ rounds. The actual transcript agrees until $h_J$ is
tested, so
\begin{equation}\label{eq:chain-hit}
 \Pr\{\text{advance from $J$ within $j$ rounds}\mid
          \text{history at first reaching $J$}\}
 \leq\min\{1,js/(m-2)\}.
\end{equation}
This null-continuation argument permits adaptive selection and repeated
contacts. It uses the holder distribution only at the start of the
waiting period; after unsuccessful contacts its posterior need not
remain uniform over all $m-2$ identities.

Fix $0<\epsilon<(1-q)^2$ and let
\begin{equation}\label{eq:chain-K}
 K=\left\lceil\frac{\log((1-q)^2/\epsilon)}
                         {2\log(1/q)}\right\rceil,\qquad d\geq K+1.
\end{equation}
After $R<K$ rounds the frontier is at most $R$, so
\eqref{eq:chain-gap} is strictly above $\epsilon$.

If $K\geq2$ and $s\leq(m-2)/2$, set
$j_0=\lfloor(m-2)/(2s)\rfloor\geq1$ and $N=K-1\geq1$.
Let $W_J$ count rounds to advance from frontier $J$ to $J+1$;
infinite waits are allowed. Equation~\eqref{eq:chain-hit} gives
$\Pr(W_J\leq j_0\mid\text{previous discovery history})\leq1/2$.
Couple these fast-wait indicators to independent fair bits using
independent uniforms. Hoeffding's inequality gives
\begin{equation}\label{eq:chain-fast}
 \Pr\left\{\sum_{J=1}^{N}{\bf1}\{W_J\leq j_0\}
                         >3N/4\right\}\leq e^{-N/8}.
\end{equation}
If frontier $K$ is reached by a round
$R<1+N(j_0+1)/4$, fewer than $N/4$ waits can last $j_0+1$
rounds or more. Thus more than $3N/4$ waits are fast, an event
bounded by \eqref{eq:chain-fast}. By \eqref{eq:chain-gap}, on its
complement the relative gap exceeds $\epsilon$. Averaging over the
finite holder-assignment distribution and the algorithm's coins
produces a fixed assignment with algorithmic failure probability at
least $1-e^{-N/8}$. Also,
$\Pr(W_J\geq j_0+1\mid\text{history})\geq1/2$, so the expected
rounds to reveal $K$ coordinates under the hard distribution are at
least $1+N(j_0+1)/2$.

The ZR restriction is essential. Dense probes can activate several
links at once, and arbitrary vector replies can encode local
structure. Indeed, the global objective in \eqref{eq:chain-objective}
does not depend on the random holder assignment; an unrestricted
server that knows this hard-family formula could output its optimum
without querying anyone. Proposition~5 separately gives exact
identification of any fixed-dimensional quadratic after
$(d+1)\lceil m/s\rceil$ gradient rounds under the broader oracle.
Neither observation contradicts this ZR lower bound, and no
unrestricted-message lower rate follows from it.

\subsection{Preconditioned Chebyshev upper bound and cost}
For any quadratic family in the anchor model, let $A=\nabla^2 f$
and $A_a=\nabla^2 f_a$, and put $P=A_a+\delta I$ for $\delta>0$.
The anchor condition and $A\succeq\mu I$ give
\begin{equation}\label{eq:chain-sandwich}
 A\preceq P\preceq A+2\delta I
   \preceq(1+2\delta/\mu)A,
 \qquad \ell P\preceq A\preceq P,
 \quad\ell=\frac\mu{\mu+2\delta}.
\end{equation}
Thus $H=P^{-1/2}AP^{-1/2}$ has spectrum in $[\ell,1]$.
With $T_k$ the Chebyshev polynomial of the first kind, set
\[
 p_k(t)=\frac{T_k((1+\ell-2t)/(1-\ell))}
               {T_k((1+\ell)/(1-\ell))},\qquad p_k(0)=1.
\]
Its three-term recurrence needs one $P^{-1}\nabla f(x)$ per step
and gives error $e_k=p_k(P^{-1}A)e_0$. On $[\ell,1]$ the numerator
has magnitude at most one. The Chebyshev closed form yields
\[
 T_k((1+\ell)/(1-\ell))\geq\frac1{2\rho^k},\qquad
 \rho=\frac{\sqrt{1+2\delta/\mu}-1}
            {\sqrt{1+2\delta/\mu}+1}.
\]
The spectral theorem in the $A$-norm now proves
\begin{equation}\label{eq:chain-cheb-gap}
 \Delta_k=\tfrac12\|e_k\|_A^2
           \leq4\rho^{2k}\Delta_0.
\end{equation}
Hence $k=\lceil\log(4/\epsilon)/(2\log(1/\rho))\rceil$
steps suffice for relative error $\epsilon$.

To apply $P^{-1}\nabla f(x)$, collect the full gradient at $x$ in
$\lceil m/s\rceil$ capped batches; the anchor's batch also gives
$\nabla f_a(x)$. Then send
$u=x-\{\nabla f(x)-\nabla f_a(x)\}/\delta$ to the anchor.
Its local proximal reply with parameter $\delta$ is
\[
 z=(A_a+\delta I)^{-1}(b_a+\delta u)
   =x-P^{-1}\nabla f(x).
\]
The recurrence uses linear combinations of prior iterates and this
reply, so it is ZR from $x_0=0$ (or when $x_0$ belongs to initial
memory). The prox input is available only after all
gradient batches, giving exactly $\lceil m/s\rceil+1$ rounds,
$m+1$ contacts, and $2(m+1)$ vector transmissions per step.
Local solve work and precision remain unpriced. At $\delta=0$,
Proposition~6 provides the separate one-pass result.

The lower and upper formulas share the $m/s$ and high-accuracy,
high-$\delta/\mu$ quadratic ZR dependence. The lower depth contains
$\log((1-q)^2/\epsilon)/\log(1/q)$; the upper depth contains
$\log(4/\epsilon)/\log(1/\rho)$. Their logarithms differ near
the initial-error threshold. The probabilistic lower statement also
needs $s\leq(m-2)/2$; the deterministic depth statement does not.
For a quantitative comparison, put
$h(t)=\log\{(\sqrt{1+t}+1)/(\sqrt{1+t}-1)\}$.
It is decreasing, while $\sqrt t\,h(t)$ is increasing: with
$v=\sqrt{1+t}$, its derivative has the sign of
$v\operatorname{artanh}(1/v)-1>0$.
Consequently
$1\leq h(t)/h(2t)\leq\sqrt2$, which is exactly
$1\leq\log(1/q)/\log(1/\rho)\leq\sqrt2$ for
$t=\delta/\mu$. If $\epsilon\leq(1-q)^4$, then
$\log((1-q)^2/\epsilon)\geq\tfrac12\log(1/\epsilon)$.
Thus the lower and upper depth expressions differ by constants in
that high-accuracy regime; as $\delta/\mu\to\infty$ each depth is
proportional to $\sqrt{\delta/\mu}\log(1/\epsilon)$.

%% file: cohort_accsvrs.tex
\section{Cohort SVRS and AccSVRS with a charged anchor}\label{app:cohort}

This appendix proves Theorems~9--10, Corollaries~11--12, and
Proposition~13 of Section~\ref{two-parameter-main}. The algorithms and the proof skeleton
are those of \citet{lin2023}: their Algorithms~1--2, Lemma~D.1,
Lemmas~3.1 and~3.5, and Theorems~3.3 and~3.6. We change three
ingredients and recheck every constant.
(i) The source samples one client per step; here a uniform cohort of
$b$ clients is drawn without replacement, which contributes the exact
finite-population factor $\xi(b)$ of \eqref{eq:app-fpc}.
(ii) The source's free master inherits the deviation
$\sqrt n\,\delta$ from average similarity. Here the anchor bound
$\delta$ is a separate parameter, and the anchor is an ordinary client
whose contacts are charged.
(iii) Every cost is converted to strict rounds and contacts.
Numerical checks of the constants and expected-potential bounds use
nonconvex clients and inexact anchor solves. They support but do not
replace the proof.

\subsection{Setting, parameters, and the epoch}\label{app:cohort-setting}
Let $f=\frac1m\sum_if_i$ be differentiable and $\mu$-strongly convex.
Individual $f_i$ need not be convex. With $x^\star=\arg\min f$ and
$\Delta_0=f(x_0)-f(x^\star)$, assume
\begin{align}
 &\|\nabla(f_a-f)(x)-\nabla(f_a-f)(y)\|\le\delta\|x-y\|,
 \tag{A1}\\
 &\frac1m\sum_{i=1}^m\|\nabla(f_i-f)(x)-\nabla(f_i-f)(y)\|^2
 \le\bar\delta^2\|x-y\|^2. \tag{A2}
\end{align}
For quadratics, (A1) is condition~(1) and
$\bar\delta^2=\|\frac1m\sum_i(A_i-A)^2\|_{\rm op}$.
Let $\mathcal C_d(\mu,\delta,\bar\delta,m)$ be the class of
$d$-dimensional quadratic families with every $A_i\succeq\mu I$ that
satisfy (A1)--(A2). The upper bounds below hold on the larger,
possibly nonquadratic, class defined by (A1)--(A2).

\textbf{Lemma C.1 (an average bound implies an anchor bound).} Under
(A2), every client satisfies (A1) with $\delta=\sqrt{m-1}\,\bar\delta$.
The factor is attained.

\textit{Proof.} Put $v_i=\nabla(f_i-f)(x)-\nabla(f_i-f)(y)$, so that
$\sum_iv_i=0$. Cauchy--Schwarz gives
$\|v_a\|^2=\|\sum_{i\ne a}v_i\|^2\le(m-1)\sum_{i\ne a}\|v_i\|^2$.
Hence $\sum_i\|v_i\|^2\ge\frac{m}{m-1}\|v_a\|^2$, and (A2) gives
$\|v_a\|^2\le(m-1)\bar\delta^2\|x-y\|^2$. Equality holds for
$v_a=(m-1)u$ and $v_i=-u$ for $i\ne a$. $\square$

Thus, without loss of generality, $\delta\le\sqrt{m-1}\,\bar\delta$.
Fix a cohort size $b\in\{1,\ldots,m\}$ and set
\begin{equation}\label{eq:cohort-parameters}
 \begin{gathered}
 E=\lceil m/b\rceil,\quad p=\frac1E,\quad
 \xi=\xi(b)=\frac{m-b}{b(m-1)},\\
 \lambda=\frac1\theta=4\max\{\mu,\ \delta,\ \bar\delta\sqrt{\xi E}\},\quad
 \tilde\kappa=\frac{\mu E}{\lambda}=\frac{\mu\theta}{p}.
 \end{gathered}
\end{equation}
Then $\delta\theta\le\frac14$, $\mu\theta\le\frac14$, and
$\theta^2\xi\bar\delta^2\le p/16$. Since $\lceil m/b\rceil\le(m+b-1)/b$
and $(m-b)(m+b-1)=m(m-1)-b(b-1)$, we also have $\xi E\le m/b^2$, so
$\lambda\le4\max\{\mu,\delta,\sqrt m\,\bar\delta/b\}$.

Let $h(x)=f_a(x)+\|x\|^2/(2\theta)-f(x)$, and let
$D_h(x,y)=h(x)-h(y)-\langle\nabla h(y),x-y\rangle$ be its Bregman
divergence. By (A1), $\nabla h-x/\theta$ is $\delta$-Lipschitz. Hence
\begin{equation}\label{eq:cohort-Dh}
 \frac{1-\delta\theta}{2\theta}\|x-y\|^2\le D_h(x,y)
 \le\frac{1+\delta\theta}{2\theta}\|x-y\|^2 .
\end{equation}
So $h$ is convex and $L_h$-smooth with $L_h=(1+\delta\theta)/\theta$, and
$\|\nabla h(x)-\nabla h(y)\|^2\le2L_hD_h(x,y)$. We also use the
three-point identity
\begin{equation}\label{eq:cohort-three-point}
 D_h(x,w)-D_h(x,w')
 =\langle x-w,\nabla h(w')-\nabla h(w)\rangle-D_h(w,w').
\end{equation}

\textbf{Epoch $\mathrm{CSVRS}(w_0)$.}
\begin{enumerate}\tightlist
\item \emph{Snapshot.} Collect all $m$ gradients $\nabla f_i(w_0)$,
including the anchor's. This takes $\lceil m/s\rceil$ rounds and
$m$ contacts. Then $g_0=\nabla f(w_0)$.
\item Draw $T\sim\mathrm{Geom}(p)$ on $\{1,2,\ldots\}$, independently
of everything else. Set $x_0=w_0$.
\item For $t=0,\ldots,T-1$:
 \begin{itemize}\tightlist
 \item Gradient estimate. Put $\hat g_0=g_0$. For $t\ge1$, draw a uniform
 $b$-subset $S_t$ and set
 $\hat g_t=g_0+\frac1b\sum_{i\in S_t}[\nabla f_i(x_t)-\nabla f_i(w_0)]$.
 This takes $\lceil b/s\rceil$ rounds and $b$ contacts; the
 $\nabla f_i(w_0)$ are stored from the snapshot.
 \item Anchor step. Contact the anchor for
 \[x_{t+1}\approx\hat x_{t+1}=\arg\min_x A^t(x),\qquad
 A^t(x)=\langle\hat g_t-\nabla f_a(x_t),x\rangle
       +\frac1{2\theta}\|x-x_t\|^2+f_a(x).\]
 This takes one round and one contact.
 \end{itemize}
\item Output $w^+=x_T$.
\end{enumerate}
For quadratic $f_a$, the anchor step is
$\hat x_{t+1}=x_t-(A_a+\lambda I)^{-1}\hat g_t$. This is one shifted-inverse
reply to the single vector $\hat g_t$. For nonquadratic $f_a$, the anchor
also needs $x_t$. Either it stores its previous reply ($x_0=w_0$ was
its snapshot query), or one more anchor round per step is charged; the
latter changes only constants.

An epoch with $T$ steps costs
$\lceil m/s\rceil+1+(T-1)(\lceil b/s\rceil+1)$ rounds and
$m+1+(T-1)(b+1)$ contacts. Its expectation is
$\lceil m/s\rceil+1+(E-1)(\lceil b/s\rceil+1)$ rounds. For $b=s$ this is
$3\lceil m/s\rceil-1$ rounds.

\subsection{One step and one epoch}
\textbf{Lemma C.2 (one step).} Let $\mathcal F_t$ contain all randomness
before $S_t$. Let $x$ be $\mathcal F_t$-measurable. Suppose the anchor
solve satisfies
\begin{equation}\label{eq:cohort-inexact}
 \|\nabla A^t(x_{t+1})\|^2\le\frac{\mu\lambda}{20}\|x_t-\hat x_{t+1}\|^2,
\end{equation}
which exact solves satisfy trivially. Then
\[
 \mathbb E_t[f(x_{t+1})-f(x)]\le D_h(x,x_t)
 -\Bigl(1+\frac{2\mu\theta}5\Bigr)\mathbb E_tD_h(x,x_{t+1})
 +\frac{2p}9D_h(w_0,x_t).
\]

\textit{Proof.} Let $\bar f_S=\frac1b\sum_{i\in S}f_i$ and
$e_t=\nabla(\bar f_{S_t}-f)(x_t)-\nabla(\bar f_{S_t}-f)(w_0)$, with
$e_0=0$. Direct expansion gives $\hat g_t=\nabla f(x_t)+e_t$ and
\begin{equation}\label{eq:cohort-gradA}
 \nabla A^t(u)=\nabla f(u)+\nabla h(u)-\nabla h(x_t)+e_t .
\end{equation}
Apply \eqref{eq:app-fpc} with cohort size $b$ to the mean-zero vectors
$\nabla(f_i-f)(x_t)-\nabla(f_i-f)(w_0)$, then use (A2). This gives
$\mathbb E_te_t=0$ and
$\mathbb E_t\|e_t\|^2\le\xi\bar\delta^2\|x_t-w_0\|^2$.

Start from strong convexity,
$f(x_{t+1})-f(x)\le\langle\nabla f(x_{t+1}),x_{t+1}-x\rangle
-\frac\mu2\|x_{t+1}-x\|^2$. Substitute \eqref{eq:cohort-gradA} at
$u=x_{t+1}$, and use the identity
$\langle\nabla h(x_t)-\nabla h(x_{t+1}),x_{t+1}-x\rangle
=D_h(x,x_t)-D_h(x,x_{t+1})-D_h(x_{t+1},x_t)$:
\begin{align*}
 f(x_{t+1})-f(x)\le{}&D_h(x,x_t)-D_h(x,x_{t+1})-D_h(x_{t+1},x_t)
 +\langle e_t,x-x_t\rangle+\langle e_t,x_t-x_{t+1}\rangle\\
 &+\langle\nabla A^t(x_{t+1}),x_{t+1}-x\rangle
 -\frac\mu2\|x_{t+1}-x\|^2 .
\end{align*}
The term $\langle e_t,x-x_t\rangle$ has conditional mean zero. Young's
inequality bounds the next term by
$\frac{1-\delta\theta}{4\theta}\|x_{t+1}-x_t\|^2
+\frac{\theta}{1-\delta\theta}\|e_t\|^2$. It also bounds the gradient
term by $\frac\mu4\|x_{t+1}-x\|^2+\frac1\mu\|\nabla A^t(x_{t+1})\|^2$.
By \eqref{eq:cohort-Dh},
$D_h(x_{t+1},x_t)\ge\frac{1-\delta\theta}{2\theta}\|x_{t+1}-x_t\|^2$.
Using $\delta\theta\le\frac14$, we get
$\frac\mu4\|x_{t+1}-x\|^2\ge\frac{\mu\theta}{2(1+\delta\theta)}D_h(x,x_{t+1})
\ge\frac{2\mu\theta}5D_h(x,x_{t+1})$, and
\[
 \frac{\theta}{1-\delta\theta}\xi\bar\delta^2\|x_t-w_0\|^2
 \le\frac{2\theta^2\xi\bar\delta^2}{(1-\delta\theta)^2}D_h(w_0,x_t)
 \le\frac{2p/16}{9/16}D_h(w_0,x_t)=\frac{2p}9D_h(w_0,x_t).
\]
It remains to show that
$B=\frac1\mu\|\nabla A^t(x_{t+1})\|^2-\frac{1-\delta\theta}{4\theta}
\|x_{t+1}-x_t\|^2\le0$.
By (A1), $f_a$ is $(\mu-\delta)$-strongly convex (weakly convex if
$\delta>\mu$). So $A^t$ is $(\lambda+\mu-\delta)\ge\lambda/2$-strongly
convex, and $\|x_{t+1}-\hat x_{t+1}\|\le2\theta\|\nabla A^t(x_{t+1})\|$.
With $\frac{1-\delta\theta}{4\theta}\ge\frac1{8\theta}$ and
$\|x_{t+1}-x_t\|^2\ge\frac12\|\hat x_{t+1}-x_t\|^2
-\|x_{t+1}-\hat x_{t+1}\|^2$,
\[
 B\le\Bigl(\frac1\mu+\frac\theta2\Bigr)\|\nabla A^t(x_{t+1})\|^2
 -\frac1{16\theta}\|\hat x_{t+1}-x_t\|^2
 \le\Bigl(\frac1{20\theta}+\frac{\mu}{40}-\frac1{16\theta}\Bigr)
 \|\hat x_{t+1}-x_t\|^2\le0,
\]
using \eqref{eq:cohort-inexact} and $\mu\theta\le\frac14$. Taking
$\mathbb E_t$ completes the proof. $\square$

\textbf{Lemma C.3 (one epoch).} Let $w^+=\mathrm{CSVRS}(w_0)$. For every
$x$ independent of the epoch's cohorts and of $T$,
\begin{align}
 \mathbb E[f(w^+)-f(x)]
 &\le pD_h(x,w_0)-\Bigl(p+\frac{2\mu\theta}5\Bigr)\mathbb ED_h(x,w^+)
 +\frac{2p}9\mathbb ED_h(w_0,w^+)\label{eq:cohort-epoch}\\
 &=p\langle x-w_0,\mathbb E\nabla h(w^+)-\nabla h(w_0)\rangle
 -\frac{7p}9\mathbb ED_h(w_0,w^+)
 -\frac{2\mu\theta}5\mathbb ED_h(x,w^+).\label{eq:cohort-epoch-ip}
\end{align}

\textit{Proof.} Generate the infinite sequence $x_0,x_1,\ldots$ with
independent cohorts, independently of $T$, and set $w^+=x_T$. For any
nonnegative sequence $a_t$ determined by the cohorts,
$\Pr(T=k)=p(1-p)^{k-1}$ gives
$\mathbb E a_{T-1}=pa_0+(1-p)\mathbb Ea_T$. Apply Lemma~C.2 at $t=k-1$,
take total expectations, weight by $\Pr(T=k)$, and sum:
\begin{align*}
\mathbb E[f(x_T)-f(x)]
 &\le\mathbb ED_h(x,x_{T-1})-\Bigl(1+\frac{2\mu\theta}5\Bigr)
 \mathbb ED_h(x,x_T)+\frac{2p}9\mathbb ED_h(w_0,x_{T-1})\\
 &=pD_h(x,w_0)-\Bigl(p+\frac{2\mu\theta}5\Bigr)\mathbb ED_h(x,x_T)
 +\frac{2p(1-p)}9\mathbb ED_h(w_0,x_T),
\end{align*}
since $D_h(w_0,x_0)=0$. This implies \eqref{eq:cohort-epoch}.
Identity \eqref{eq:cohort-three-point} turns
$pD_h(x,w_0)-pD_h(x,w^+)$ into the inner product minus
$pD_h(w_0,w^+)$, which gives \eqref{eq:cohort-epoch-ip}. $\square$

\subsection{Theorem 9: cohort SVRS}\label{app:cohort-svrs}
\textbf{Theorem 9 (restated).} Let $w_{k+1}=\mathrm{CSVRS}(w_k)$ with
\eqref{eq:cohort-parameters}, and define
$\Phi_k=\mathbb E[f(w_k)-f(x^\star)]
+\frac12(p+\frac{2\mu\theta}5)\mathbb ED_h(x^\star,w_k)$. Then
\[
 \Phi_{k+1}\le\max\Bigl\{\frac12,\frac1{1+0.4\tilde\kappa}\Bigr\}\Phi_k
 \le\Bigl(1-\min\Bigl\{\frac12,\frac{\tilde\kappa}5\Bigr\}\Bigr)\Phi_k,
 \qquad
 \Phi_0\le\Bigl(\frac54+\frac5{8\tilde\kappa}\Bigr)\Delta_0 .
\]
Hence $\mathbb E[f(w_K)-f(x^\star)]\le\varepsilon$ for
$K=\lceil\max\{2,5/\tilde\kappa\}\log(\Phi_0/\varepsilon)\rceil$.
The expected cost is $K$ times the epoch cost of
Section~\ref{app:cohort-setting}. For $b=s$, this is
$O\bigl((m/s+\lambda/\mu)\log(\Phi_0/\varepsilon)\bigr)$ rounds.

\textit{Proof.} Apply \eqref{eq:cohort-epoch} with $x=x^\star$ and with
$x=w_k$; both are independent of epoch $k$. Adding the two gives
\[
 \begin{aligned}
 \mathbb E[2f(w_{k+1})-f(w_k)-f(x^\star)]
 &\le pD_h(x^\star,w_k)-\Bigl(p+\frac{2\mu\theta}5\Bigr)
 \mathbb ED_h(x^\star,w_{k+1})\\
 &\quad-\Bigl(\frac{5p}9+\frac{2\mu\theta}5\Bigr)\mathbb ED_h(w_k,w_{k+1}).
 \end{aligned}
\]
Drop the last term and halve. This gives
$\Phi_{k+1}\le\frac12\mathbb E[f(w_k)-f^\star]+\frac p2\mathbb E
D_h(x^\star,w_k)$, which implies the first rate because
$\frac{2\mu\theta}{5p}=0.4\tilde\kappa$. Next,
$\frac1{1+x}\le1-\frac x2$ for $x\in[0,1]$, which gives the second
rate. For $\Phi_0$, use \eqref{eq:cohort-Dh},
$\frac{1+\delta\theta}{2\theta}\le\frac5{8\theta}$, and
$\frac\mu2\|w_0-x^\star\|^2\le\Delta_0$:
$\frac12(p+\frac{2\mu\theta}5)\frac5{8\theta}\cdot\frac2\mu\Delta_0
=(\frac{5}{8\tilde\kappa}+\frac14)\Delta_0$.
The epochs are independent, so expected costs add. $\square$

\subsection{Theorem 10: cohort AccSVRS}\label{app:cohort-accsvrs}
\textbf{Algorithm.} Put $y_0=z_0=x_0$, $\tau=\frac14\min\{1,
\sqrt{\tilde\kappa}\}$, and $\alpha=\theta/(2p\tau)=E/(2\lambda\tau)$.
For $k\ge0$:
\begin{enumerate}\tightlist
\item $x_{k+1}=\tau z_k+(1-\tau)y_k$.
\item $y_{k+1}=\mathrm{CSVRS}(x_{k+1})$, with a fresh snapshot at $x_{k+1}$.
\item Draw an independent uniform $b$-subset $S'_k$ and form
\[
 \mathcal G_{k+1}
 =p\bigl[\nabla(f_a-\bar f_{S'_k})(x_{k+1})
 -\nabla(f_a-\bar f_{S'_k})(y_{k+1})+\lambda(x_{k+1}-y_{k+1})\bigr].
\]
The $\nabla f_j(x_{k+1})$ come from the snapshot; the
$\nabla f_j(y_{k+1})$ for $j\in S'_k$ and $\nabla f_a(y_{k+1})$ cost
$\lceil b/s\rceil+1$ rounds and $b+1$ contacts.
\item $z_{k+1}=\arg\min_u\{\langle\mathcal G_{k+1},u\rangle
+\frac1{2\alpha}\|u-z_k\|^2+\frac{3\mu}{20}\|u-y_{k+1}\|^2\}
=\frac{z_k+0.3\mu\alpha\,y_{k+1}-\alpha\mathcal G_{k+1}}{1+0.3\mu\alpha}$.
\end{enumerate}
Since $\nabla h+\nabla(f-\bar f_S)=\nabla(f_a-\bar f_S)+(\cdot)/\theta$,
we have $\mathcal G_{k+1}=p[\nabla h(x_{k+1})-\nabla h(y_{k+1})+n_{k+1}]$,
where $n_{k+1}=\nabla(f-\bar f_{S'_k})(x_{k+1})-\nabla(f-\bar f_{S'_k})(y_{k+1})$.

\textbf{Lemma C.4 (mapping).} $\mathbb E_{S'}\mathcal G_{k+1}
=p[\nabla h(x_{k+1})-\nabla h(y_{k+1})]$ and
$\mathbb E_{S'}\|\mathcal G_{k+1}\|^2\le\frac{8p^2}{3\theta}
D_h(x_{k+1},y_{k+1})$.

\textit{Proof.} $S'_k$ is independent of $(x_{k+1},y_{k+1})$, so
\eqref{eq:app-fpc} and (A2) give $\mathbb E n_{k+1}=0$ and
$\mathbb E\|n_{k+1}\|^2\le\xi\bar\delta^2\|x_{k+1}-y_{k+1}\|^2$. By
smoothness of $h$ and \eqref{eq:cohort-Dh},
\[
 \frac{\mathbb E\|\mathcal G_{k+1}\|^2}{p^2}
 \le\Bigl[\frac{2(1+\delta\theta)}\theta
 +\frac{2\theta\xi\bar\delta^2}{1-\delta\theta}\Bigr]D_h
 \le\Bigl[\frac5{2\theta}+\frac83\cdot\frac{p}{16\theta}\Bigr]D_h
 \le\frac8{3\theta}D_h.\qquad\square
\]

\textbf{Lemma C.5 (one outer step).} For $x$ independent of outer step $k$,
\[
 \frac\alpha\tau\mathbb E_k[f(y_{k+1})-f(x)]
 +\frac{1+0.3\mu\alpha}2\mathbb E_k\|x-z_{k+1}\|^2
 \le(1-\tau)\frac\alpha\tau[f(y_k)-f(x)]+\frac12\|x-z_k\|^2 .
\]

\textit{Proof.} Write $D=D_h(x_{k+1},y_{k+1})$ and $u_k=\nabla h(y_{k+1})
-\nabla h(x_{k+1})$. Apply \eqref{eq:cohort-epoch-ip} with $w_0=x_{k+1}$
and split $x-x_{k+1}=(z_k-x_{k+1})+(x-z_k)$, where
$z_k-x_{k+1}=\frac{1-\tau}\tau(x_{k+1}-y_k)$. Applying
\eqref{eq:cohort-epoch-ip} with $x=y_k$ and dropping its last term gives
$p\mathbb E_k\langle x_{k+1}-y_k,u_k\rangle\le f(y_k)
-\mathbb E_kf(y_{k+1})-\frac{7p}9\mathbb E_kD$. Hence
\[
 \mathbb E_k[f(y_{k+1})-f(x)]\le\frac{1-\tau}\tau
 \mathbb E_k[f(y_k)-f(y_{k+1})]-\frac{7p}{9\tau}\mathbb E_kD
 +\mathbb E_k\Bigl[\langle z_k-x,\mathcal G_{k+1}\rangle
 -\frac{3\mu}{20}\|x-y_{k+1}\|^2\Bigr].
\]
Here we used Lemma~C.4 for the mean of $\mathcal G_{k+1}$ and
$\frac{2\mu\theta}5D_h(x,y_{k+1})\ge\frac{\mu(1-\delta\theta)}5
\|x-y_{k+1}\|^2\ge\frac{3\mu}{20}\|x-y_{k+1}\|^2$. The $z$-objective is
$(1/\alpha+0.3\mu)$-strongly convex, so the bracket is at most
$\frac\alpha2\|\mathcal G_{k+1}\|^2+\frac1{2\alpha}\|x-z_k\|^2
-\frac{1+0.3\mu\alpha}{2\alpha}\|x-z_{k+1}\|^2$. By Lemma~C.4,
$\frac\alpha2\mathbb E\|\mathcal G_{k+1}\|^2-\frac{7p}{9\tau}D
\le(\frac{4\alpha p^2}{3\theta}-\frac{7p}{9\tau})D\le0$ whenever
$\alpha\le7\theta/(12p\tau)$, which our $\alpha=\theta/(2p\tau)$
satisfies. Moving the $f(y_{k+1})$ terms left and multiplying by
$\alpha$ gives the claim. $\square$

\textbf{Theorem 10 (restated).} Let
$\Phi_k=\mathbb E[f(y_k)-f(x^\star)]+\frac{\tau(1+0.3\mu\alpha)}{2\alpha}
\mathbb E\|x^\star-z_k\|^2$. Then
\[
 \Phi_{k+1}\le\Bigl(1-\frac{\min\{1,\sqrt{\tilde\kappa}\}}4\Bigr)\Phi_k,
 \qquad\Phi_0\le1.2\,\Delta_0 ,
\]
so $\mathbb E[f(y_K)-f(x^\star)]\le\varepsilon$ for
$K=\lceil\max\{4,4/\sqrt{\tilde\kappa}\}\log(1.2\Delta_0/\varepsilon)\rceil$.
One outer step costs
$\lceil m/s\rceil+E(\lceil b/s\rceil+1)+1$ rounds and
$m+E(b+1)+1$ contacts in expectation.
\begin{itemize}\tightlist
\item For $b=s$, an outer step costs $3\lceil m/s\rceil+1$ rounds, and the
expected total satisfies
\begin{equation}\label{eq:cohort-total}
\begin{aligned}
 \bar R&\le16\Bigl(\Bigl\lceil\frac ms\Bigr\rceil+\sqrt{\Bigl\lceil\frac ms
 \Bigr\rceil\frac\lambda\mu}\Bigr)\log\frac{1.2\Delta_0}\varepsilon
 +4\Bigl\lceil\frac ms\Bigr\rceil\\
 &=O\Bigl(\Bigl[\frac ms+\sqrt{\frac ms}\sqrt{\frac\delta\mu}
 +\frac{m^{3/4}}s\sqrt{\frac{\bar\delta}\mu}\Bigr]
 \log\frac{\Delta_0}\varepsilon\Bigr).
\end{aligned}
\end{equation}
\item For $b=m$, the steps are deterministic ($T\equiv1$ and
$\mathcal G$ is noise-free), an outer step costs $2\lceil m/s\rceil+2$
rounds, and $\lambda=4\max\{\mu,\delta\}$. This gives
$O((m/s)(1+\sqrt{\delta/\mu})\log(\Delta_0/\varepsilon))$, a
nonquadratic analogue of Theorem~8.
\end{itemize}
At $s=b=1$ and $\delta\le\sqrt m\,\bar\delta$,
\eqref{eq:cohort-total} is the source rate
$O((m+m^{3/4}\sqrt{\bar\delta/\mu})\log)$.

\textit{Proof.} Take $x=x^\star$ in Lemma~C.5 and multiply by
$\tau/\alpha$. This gives $\Phi_{k+1}\le\max\{1-\tau,(1+0.3\mu\alpha)^{-1}\}
\Phi_k$, where $0.3\mu\alpha=0.15\tilde\kappa/\tau$. There are two cases.
\begin{itemize}\tightlist
\item If $\tilde\kappa\ge1$, then $\tau=\frac14$ and
$(1+0.6\tilde\kappa)^{-1}\le0.625<1-\tau$.
\item If $\tilde\kappa<1$, then $\tau=\sqrt{\tilde\kappa}/4$ and
$t=0.6\sqrt{\tilde\kappa}\le\frac35$. Since $(1+t)(1-\frac{5t}8)\ge1$
on $[0,\frac35]$, we get $(1+t)^{-1}\le1-0.375\sqrt{\tilde\kappa}<1-\tau$.
\end{itemize}
For the initial potential, $z_0=y_0$ and
$\frac\mu2\|y_0-x^\star\|^2\le\Delta_0$ give
$\Phi_0\le(1+\frac\tau{\alpha\mu}+0.3\tau)\Delta_0$. Here
$\frac\tau{\alpha\mu}=\frac{2\tau^2}{\tilde\kappa}
=\frac{\min\{1,\tilde\kappa\}}{8\tilde\kappa}\le\frac18$ and
$0.3\tau\le0.075$, so $\Phi_0\le1.2\Delta_0$.

For the costs, the epoch's expected cost is that of
Section~\ref{app:cohort-setting}, and the mapping adds
$\lceil b/s\rceil+1$ rounds. For $b=s$, $E=\lceil m/s\rceil$, and
$\tilde\kappa=\mu\lceil m/s\rceil/\lambda$. Then
$(3E+1)\max\{4,4/\sqrt{\tilde\kappa}\}\le16E+16\sqrt{E\lambda/\mu}$,
and the ceiling of $K$ adds at most $4E$. The order bound follows from
$\lambda\le4\max\{\mu,\delta,\sqrt m\,\bar\delta/s\}$ and
$\lceil m/s\rceil\le2m/s$. $\square$

\textbf{Zero-respecting implementation.} Take quadratic clients, exact
anchor solves, and $x_0=0$. Every query point is a linear combination of
earlier replies: the snapshot point, cohort points, $x_{k+1}$, and
$y_{k+1}$. Every reply is a gradient or the shifted inverse
$(A_a+\lambda I)^{-1}\hat g_t$ with $\hat g_t$ in memory, and the output
$y_K$ is in the span. So cohort AccSVRS is an algorithm of the ZR local
oracle of Appendix~\ref{app:zr-chain}, and Corollary~11 applies to it.

\subsection{Corollary 11: the two-parameter ZR lower bound}\label{app:cohort-lower}
\textbf{Lemma C.6 (average dissimilarity of the chain).} Every member of
the hard family of Appendix~\ref{app:zr-chain} with parameter $\delta$
satisfies (A2) with $\bar\delta=\frac{\sqrt3}2\delta\sqrt m$.

\textit{Proof.} Write $A_i=\mu I+m\alpha M_i$. Here $M_i\succeq0$
collects client $i$'s links $v_\ell v_\ell^\top$. For $c_0$ it also
contains the boundary terms $e_1e_1^\top+(1-q)e_de_d^\top$, and the
anchor has $M_a=0$. Then $L=\sum_iM_i$ and $A=\mu I+\alpha L$. Since
$\sum_i(A_i-A)=0$,
\[
 \frac1m\sum_i(A_i-A)^2=\alpha^2\Bigl(m\sum_iM_i^2-L^2\Bigr),
 \qquad 0\preceq m\sum_iM_i^2-L^2\preceq m\sum_iM_i^2 .
\]
A client never holds adjacent links. So each row of $M_i$ has diagonal at
most $2$ (one link plus at most one boundary term, $1-q<1$) and
off-diagonal mass at most $1$, and Gershgorin gives $\|M_i\|\le3$. Hence
$M_i^2\preceq3M_i$, and $m\sum_iM_i^2\preceq3mL$. The paragraph after
\eqref{eq:chain-objective} gives $\|L\|\le4$. Therefore
$\bar\delta^2\le12m\alpha^2=\frac34m\delta^2$, using $\alpha=\delta/4$.
$\square$

Numerical checks confirm the bound
for $m$ from $4$ to $128$ and $d$ from $16$ to $256$. The worst observed
ratio $\bar\delta/(\delta\sqrt m)$ is $0.70$, and
$\bar\delta\to\delta\sqrt{m/2-1}$ for large $m,d$.

\textbf{Corollary 11 (restated).} Fix $\mu,\delta_a,\bar\delta>0$,
$m\ge3$, and $s\le(m-2)/2$. Let
$\delta=\min\{\delta_a,2\bar\delta/\sqrt{3m}\}$ and suppose
$\delta\ge3\mu$. Let $q$ be given by \eqref{eq:chain-parameters}. Let
$\varepsilon>0$ satisfy $4\varepsilon\le(1-q)^4$, and suppose $K$ from
\eqref{eq:chain-K} at accuracy $4\varepsilon$ satisfies $K\ge7$. Let
$d\ge K+1$. Consider any ZR algorithm, with adaptive cohorts and a
possibly random number of rounds, whose output satisfies
$\mathbb E[f(\hat x)-f(x^\star)]\le\varepsilon\Delta_0$ on every member
of $\mathcal C_d(\mu,\delta_a,\bar\delta,m)$. Its expected number of
rounds $\bar R$ on some member obeys
\[
 \bar R\ \ge\ \max\Bigl\{\frac m{16s},\
 \frac{m}{1792\,s}\sqrt{\frac{\min\{\delta_a,\bar\delta/\sqrt m\}}\mu}
 \log\frac1{4\varepsilon}\Bigr\}.
\]
Without the ZR restriction, Theorem~2 still gives the participation
term: for any algorithm with a fixed budget of $R$ rounds and accuracy
$\varepsilon<1$, $R\ge(1-\varepsilon)m/s$.

\textit{Proof.} \emph{Embedding.} Theorem~7's family with parameter
$\delta$ has anchor deviation at most $\delta\le\delta_a$. By Lemma~C.6
its average dissimilarity is at most
$\frac{\sqrt3}2\delta\sqrt m\le\bar\delta$. Its clients are
$\mu$-strongly convex. So the family lies in
$\mathcal C_d(\mu,\delta_a,\bar\delta,m)$. Also
$\delta\ge\min\{\delta_a,\bar\delta/\sqrt m\}$.

\emph{From expectations to Theorem~7's event.} Let
$\Theta=1+(K-1)(j_0+1)/4$, with $K$ and $j_0$ as in Theorem~7 at
accuracy $4\varepsilon$. Suppose $\bar R<\Theta/4$ on every member.
By Markov's inequality, $\Pr(R\ge\Theta)<\frac14$ and
$\Pr(f(\hat x)-f^\star>4\varepsilon\Delta_0)\le\frac14$, so the event
$\mathcal E=\{R<\Theta,\ f(\hat x)-f^\star\le4\varepsilon\Delta_0\}$
has probability above $\frac12$ on every member. Define $\mathcal A'$
to run the algorithm for at most $\lceil\Theta\rceil-1$ rounds. It
returns $\hat x$ if the algorithm has stopped, and $0$ otherwise.
$\mathcal A'$ is ZR with fewer than $\Theta$ rounds, and on
$\mathcal E$ its output equals $\hat x$. Theorem~7 at accuracy
$4\varepsilon$ yields a member on which $\mathcal A'$ has gap at most
$4\varepsilon\Delta_0$ with probability at most
$e^{-(K-1)/8}\le e^{-3/4}<\frac12$. This contradicts
$\Pr(\mathcal E)>\frac12$. Hence $\bar R\ge\Theta/4$ on some member.

\emph{Constants.} With $\varepsilon'=4\varepsilon\le(1-q)^4$, we have
$\log((1-q)^2/\varepsilon')\ge\frac12\log(1/\varepsilon')$. Also
$\log(1/q)=\log(1+\frac2{\sqrt\kappa-1})\le\frac2{\sqrt\kappa-1}$, and
$\sqrt\kappa-1\ge\frac12\sqrt{\delta/\mu}$ for $\kappa=1+\delta/\mu\ge4$.
Hence $K\ge\frac1{16}\sqrt{\delta/\mu}\log(1/\varepsilon')$. Next,
$K\ge7$ gives $K-1\ge\frac67K$, and $m\ge3$ gives
$j_0+1>(m-2)/(2s)\ge m/(6s)$. So
$\Theta/4>\frac{Km}{112s}$. This is at least $\frac m{16s}$ and at least
$\frac m{1792s}\sqrt{\delta/\mu}\log\frac1{4\varepsilon}$. The last
sentence of the corollary is Theorem~2: with $R<(1-\varepsilon)m/s$,
(7) gives expected gap above $\varepsilon\Delta_0$ on a member with
$\delta=\bar\delta=0$. $\square$

\subsection{Large cohorts: a two-group chain and batched Acc-S-DANE}
\label{app:large-cohorts}
Corollary~11 needs $s\le(m-2)/2$. For larger cohorts Theorem~7's
spreading argument stops working, and its chain has
$\bar\delta=\Theta(\delta\sqrt m)$. The construction below keeps
$\bar\delta=O(\delta)$ for every $m$. It slows every algorithm
regardless of $s$, including $s=m$. The one-coordinate-per-round
argument is the classical one of \citet{arjevani2015communication}; the
new part is that it fits the two-parameter class with a charged anchor.

\textbf{Lemma C.7 (two-group chain).} Fix $\mu,\delta>0$, $m\ge3$, and
$d\ge2$, and keep $q,\alpha,c,Q,f$ of \eqref{eq:chain-parameters} and
\eqref{eq:chain-objective}. Split the $m-1$ non-anchor clients into a
group $O$ of size $n_O=\lceil(m-1)/2\rceil$ and a group $E$ of size
$n_E=\lfloor(m-1)/2\rfloor\ge1$. Every client has the base
$\mu\|x\|^2/2$. The anchor also receives $-cx_1$. Each member of $O$
receives $\frac c{n_O}\{x_1^2+(1-q)x_d^2+\sum_{\ell\ \rm odd}
(v_\ell^\top x)^2\}/2$, and each member of $E$ receives
$\frac c{n_E}\sum_{\ell\ \rm even}(v_\ell^\top x)^2/2$. Then the
local objectives average to $f$, every client is $\mu$-strongly convex,
$\|\nabla^2f-\nabla^2f_a\|_{\rm op}\le\delta$, and
$\bar\delta\le\sqrt3\,\delta$. Every reply of the ZR oracle to vectors
in $V_J$ lies in $V_{J+1}$, for every client and every $J<d$.

\textit{Proof.} The group weights sum to $c$ per term, so averaging
over $m$ clients gives $\alpha$ per term; this is $f$. The anchor
Hessian is $\mu I$, so the deviation is $\alpha\|L\|\le4\alpha=\delta$
as in Appendix~\ref{app:zr-chain}. Write $P_O,P_E\succeq0$ for the
matrices of the bracketed quadratic forms, so $L=P_O+P_E$. With
$B_i=\nabla^2f_i-\mu I$ and $\frac1m\sum_iB_i=\alpha L$,
\[
 \frac1m\sum_i(\nabla^2f_i-\nabla^2f)^2
 =\frac1m\sum_iB_i^2-\alpha^2L^2
 =\alpha^2\Bigl(\frac m{n_O}P_O^2+\frac m{n_E}P_E^2-L^2\Bigr).
\]
The links in $P_O$ are disjoint $2\times2$ blocks of norm $2$, and the
boundary terms add at most $1$ on the diagonal. Gershgorin gives
$\|P_O\|,\|P_E\|\le3$, so $P_j^2\preceq3P_j$. For $m\ge3$,
$m/n_j\le m/\lfloor(m-1)/2\rfloor\le4$. Hence the matrix above is at
most $12\alpha^2L\preceq48\alpha^2I=3\delta^2I$, which gives
$\bar\delta\le\sqrt3\,\delta$. Each client's Hessian, and hence its
shifted inverse, is block diagonal with blocks of size at most two that
never straddle two links. Only the block containing link $J$ can cross
the cut $V_J\oplus V_J^\perp$, and it ends at coordinate $J+1$. The
only linear term is the anchor's $-ce_1\in V_1$. So every gradient,
Hessian-vector, shifted-inverse, and proximal reply maps $V_J$ into
$V_{J+1}$. $\square$

\textbf{Corollary 12 (lower bound for every cohort size).} Fix
$\mu,\delta_a,\bar\delta>0$, $m\ge3$, and any $s\le m$. Let
$\delta=\min\{\delta_a,\bar\delta/\sqrt3\}\ge3\mu$, let $q$ be given by
\eqref{eq:chain-parameters}, let $4\varepsilon\le(1-q)^4$, let
$K'=\min\{R:(1-q)^2q^{2R}\le4\varepsilon\}$, and let $d\ge K'+1$.
Consider any ZR algorithm, with adaptive cohorts and a possibly random
number of rounds $R$, whose output satisfies
$\mathbb E[f(\hat x)-f^\star]\le\varepsilon\Delta_0$ on the instance of
Lemma~C.7. Then
\[
 \mathbb E[R]\ \ge\ \frac3{64}\sqrt{\frac{\min\{\delta_a,\bar\delta/\sqrt3\}}\mu}
 \log\frac1{4\varepsilon}.
\]
Combined with Corollary~11, for every $s\le m$ the expected rounds on
some member of the class are at least a constant times
\begin{equation}\label{eq:all-s-lower}
 \max\Bigl\{\frac ms,\
 \frac ms\sqrt{\frac{\min\{\delta_a,\bar\delta/\sqrt m\}}\mu}L_\varepsilon,\
 \sqrt{\frac{\min\{\delta_a,\bar\delta/\sqrt3\}}\mu}L_\varepsilon\Bigr\},
 \qquad L_\varepsilon=\log\frac1{4\varepsilon}.
\end{equation}

\textit{Proof.} By Lemma~C.7 the instance has anchor deviation at most
$\delta\le\delta_a$, average dissimilarity at most
$\sqrt3\,\delta\le\bar\delta$, and $\mu$-strongly convex clients. The
server memory starts at $\{0\}\subset V_0$. By Lemma~C.7, each round
adds replies in $V_{J+1}$ when the memory lies in $V_J$, whatever the
cohort and its size. After $R$ rounds the output lies in $V_R$. This
holds on every sample path. If $R\le K'-1\le d-2$, then
\eqref{eq:chain-gap} and the minimality of $K'$ give
$f(\hat x)-f^\star\ge(1-q)^2q^{2(K'-1)}\Delta_0>4\varepsilon\Delta_0$.
By Markov's inequality, this event has probability at most $\frac14$.
So $\Pr(R\ge K')\ge\frac34$ and $\mathbb E[R]\ge\frac34K'$. As in
Corollary~11, $\log((1-q)^2/(4\varepsilon))\ge\frac12\log(1/(4\varepsilon))$
and $\log(1/q)\le4\sqrt{\mu/\delta}$, so
$K'\ge\frac1{16}\sqrt{\delta/\mu}\log(1/(4\varepsilon))$.

For \eqref{eq:all-s-lower}: if $s\le(m-2)/2$, the first two terms are
Corollary~11. If $s>(m-2)/2$, then $m/s<6$ for $m\ge3$. The first term
is at most $6$, and a valid output needs at least one round with
probability at least $1-\varepsilon$, because $x_0=0$ has gap
$\Delta_0$. The second term is at most $6$ times the third, since
$\bar\delta/\sqrt m\le\bar\delta/\sqrt3$. $\square$

\textbf{Proposition 13 (batched Acc-S-DANE).} Assume every $f_i$ is
$\mu$-strongly convex, as in Section~\ref{strict-participation-model-and-accounting},
and that (A2) holds with $\bar\delta\ge\mu/8$. Run Acc-S-DANE
\citep[Algorithm~2]{jiang2024} with full participation, $\lambda=2\bar\delta$,
and exact local steps. Each iteration contacts all $m$ clients
twice, in capped batches. The first $\lceil m/s\rceil$ rounds collect
$\nabla f_i(y_r)$. The next $\lceil m/s\rceil$ rounds return the local
proximal points
$x_{i,r+1}=\arg\min_x f_i(x)+\langle\nabla f(y_r)-\nabla f_i(y_r),x\rangle
+\frac\lambda2\|x-y_r\|^2$. Local optimality gives
$\nabla f_i(x_{i,r+1})=\nabla f_i(y_r)-\nabla f(y_r)-\lambda(x_{i,r+1}-y_r)$,
so no third reply is needed. After
\[
 R^\ast=\Bigl\lceil2\sqrt{8\bar\delta/\mu}\,
 \log\bigl(1+2/\sqrt\varepsilon\bigr)\Bigr\rceil
\]
iterations, $f(x_{R^\ast})-f^\star\le\varepsilon\Delta_0$
deterministically. This uses $2\lceil m/s\rceil R^\ast$ strict rounds,
of order $O((m/s)\sqrt{\bar\delta/\mu}\log(1/\varepsilon))$. For
quadratic clients and $x_0=0$, the method is ZR.

\textit{Proof.} For $s=n$, condition~(5) of \citet{jiang2024}
($\delta$-SOD) is (A2) with $\delta=\bar\delta$. Exact local steps make
their accuracy condition hold with zero left side. Their Theorem~6, for
$\mu\le8\bar\delta$ and $t=\sqrt{\mu/(8\bar\delta)}\le1$, gives
$f(x_R)-f^\star\le2\mu D^2/[(1+t)^R-(1-t)^R]^2$ with
$D=\|x_0-x^\star\|$. Strong convexity gives $\mu D^2\le2\Delta_0$, and
$(1+t)^R-(1-t)^R\ge(1+t)^R-1$. So the gap is at most $\varepsilon\Delta_0$
once $(1+t)^R\ge1+2/\sqrt\varepsilon$. Since $\log(1+t)\ge t/2$ for
$t\le1$, $R^\ast$ suffices. For ZR, the local point is the proximal
reply $(A_i+\lambda I)^{-1}(b_i+\lambda v)$ with
$v=y_r-(\nabla f(y_r)-\nabla f_i(y_r))/\lambda$ in memory. The updates of
$y_r$, $v_r$, and $x_{r+1}$ are linear combinations of memory. $\square$

Numerical checks cover Lemma~C.7 for $3\le m\le101$ (worst observed
$\bar\delta/\delta=0.92$), its $s=m$ frontier, and the bound of
\citet[Theorem~6]{jiang2024} on random strongly convex federations
(worst observed gap-to-bound ratio $0.11$).

\textbf{Where the bounds match.} Take $\delta_a\le\sqrt{m-1}\,\bar\delta$
(Lemma~C.1), $\min\{\delta_a,\bar\delta/\sqrt m\}\ge3\mu$, $\mu$-strongly
convex clients, and the high-accuracy regime of Corollaries~11--12.
Lower bounds are \eqref{eq:all-s-lower}. Upper bounds are the minimum of
Theorems~8 and~10 and Proposition~13. We omit $L_\varepsilon$ and
constants.
\begin{itemize}\tightlist
\item If $\delta_a\le\bar\delta/\sqrt m$, Theorem~8 gives
$(m/s)\sqrt{\delta_a/\mu}$. This matches the second term for
$s\le(m-2)/2$ and the third term within a factor $6$ otherwise.
\item If $\bar\delta/\sqrt m\le\delta_a\le\sqrt m\,\bar\delta/s$, the
$\bar\delta$ term dominates \eqref{eq:cohort-total}. It equals
$(m/s)\sqrt{(\bar\delta/\sqrt m)/\mu}$, matching the second term; the $m/s$
term is smaller because $\bar\delta/\sqrt m\ge3\mu$.
\item If $\delta_a>\sqrt m\,\bar\delta/s$ (poor anchor), Theorem~10
costs $\sqrt{m/s}\sqrt{\delta_a/\mu}$ and Proposition~13 costs
$(m/s)\sqrt{\bar\delta/\mu}$. Their minimum is within
\[
 O\bigl(\min\{\sqrt s,\ m^{1/4},\ m/s\}\bigr)
\]
of \eqref{eq:all-s-lower}. The factor $\sqrt s$ is Theorem~10 against
the second term:
$\sqrt{s\delta_a/(\sqrt m\,\bar\delta)}\le\sqrt s$. The factor
$m^{1/4}$ is Proposition~13 against the second term. The factor $m/s$
is Proposition~13 against the third term when
$\delta_a\ge\bar\delta/\sqrt3$, and Theorem~10 against the third term
(ratio $\sqrt{m/s}$) otherwise.
\end{itemize}
So the complexity is settled up to constants when $s=O(1)$ or
$s=\Omega(m)$. This includes full participation, where it is
$\Theta(\sqrt{\min\{\delta_a,\bar\delta\}/\mu}\,L_\varepsilon)$. Among
Theorems~8 and~10 and Proposition~13, the largest remaining gap is
$O(m^{1/4})$, attained for $\sqrt m\leq s\leq m^{3/4}$; before
Corollary~12 and Proposition~13 it was $\sqrt s$, up to $\sqrt{m/2}$.
The anchor-free method of Appendix~\ref{app:anchor-free} (Corollary~15)
reduces the polynomial gap to
$O(1+\min\{\sqrt s/m^{1/4},\sqrt{m/s}\})\leq O(m^{1/8})$
(Appendix~\ref{app:vm-gap}). Corollary~15 carries the logarithmic
overhead stated there. If clients may
be nonconvex, as Theorems~9--10 allow, Proposition~13 and Corollary~15
are unavailable. The lower bounds still apply, and the gap is
$O(\min\{\sqrt s,m^{3/4}/\sqrt s\})\leq O(m^{3/8})$.

In the poor-anchor case, other cohort sizes $b$ do not improve
Theorem~10: \eqref{eq:cohort-parameters} gives an anchor term of order
$\sqrt{mb}/s\cdot\sqrt{\delta_a/\mu}$ for $b\ge s$, and of order
$\sqrt{m/b}\sqrt{\delta_a/\mu}$ for $b<s$. Both are minimized at $b=s$.
Removing the anchor term requires sampling cohorts without an anchor,
which Appendix~\ref{app:anchor-free} does. Any rate outside the ZR
oracle remains open; Proposition~5 shows that the unrestricted oracle
has no accuracy-dependent lower bound on fixed-dimensional quadratics.

%% file: anchor_free.tex
\section{Anchor-free cohort SVRS and its Catalyst acceleration}\label{app:anchor-free}

This appendix proves Theorem~14 and Corollary~15 of
Section~\ref{two-parameter-main}. Theorems~9--10 route every step
through the anchor, so their rates carry $\sqrt{\delta_a/\mu}$, and
a poor anchor costs up to $\sqrt{m/s}\sqrt{\delta_a/\mu}$. The method
below uses no designated master client. In each step the anchor's
proximal reply is replaced by the \emph{average} of the proximal replies
of a fresh uniform cohort. On quadratics, the mean of these replies is
a proximal step on a ``virtual master'' whose Hessian is within
$\bar\delta^2/(\mu+\lambda)$ of the global one (Lemma~C.8). An individual
client need not be this close: its deviation can be $\sqrt{m-1}\,\bar\delta$
(Lemma~C.1). The price is that the virtual master is random. Its
fluctuation is controlled by sampling without replacement (Lemma~C.8(ii)).
Throughout this appendix the clients are quadratic,
$f_i(x)=\frac12x^\top A_ix-b_i^\top x$, with every $A_i\succeq\mu I$.
We write $A=\frac1m\sum_iA_i$, $D_i=A_i-A$,
$\bar\delta^2=\|\frac1m\sum_iD_i^2\|_{\rm op}$,
$\delta_{\max}=\max_i\|D_i\|\leq\sqrt{m-1}\,\bar\delta$,
$L=\|A\|$, $L_{\max}=\max_i\|A_i\|$, and $\xi=\xi(s)=\frac{m-s}{s(m-1)}$.
Cohorts have exactly $s$ clients, so each cohort contact is one strict
round.

\subsection{The method}\label{app:vm-method}
Let $h=f+\frac\kappa2\|\cdot-y\|^2$ for a center $y$ and a shift
$\kappa\geq0$; $\kappa=0$ gives $f$. Client $i$'s part is
$h_i=f_i+\frac\kappa2\|\cdot-y\|^2$ with Hessian $A_i+\kappa I$.
Fix $\lambda>0$ and an epoch length $T\geq1$.

\textbf{VM-SVRS epoch} from a snapshot $w$:
\begin{enumerate}\tightlist
\item Collect $\nabla h(w)=\nabla f(w)+\kappa(w-y)$ in $\lceil m/s\rceil$
rounds ($m$ contacts). Put $x_0=w$ and $g_0=\nabla h(w)$.
\item For $t=0,\ldots,T-1$:
 \begin{enumerate}\tightlist
 \item If $t\geq1$, draw a uniform $s$-cohort $S'_t$, send it
 $x_t-w$, and receive the Hessian-vector products
 $A_i(x_t-w)$. Form
 $g_t=\nabla h(w)+\frac1s\sum_{i\in S'_t}(A_i+\kappa I)(x_t-w)$.
 \item Draw a uniform $s$-cohort $S_t$, independent of $S'_t$ and of
 the past. Send it $g_t$. Each $i\in S_t$ replies
 $u_i=x_t-(A_i+(\kappa+\lambda)I)^{-1}g_t$, which is the local
 proximal point
 $\arg\min_u h_i(u)+\langle g_t-\nabla h_i(x_t),u\rangle
 +\frac\lambda2\|u-x_t\|^2$.
 Set $x_{t+1}=\frac1s\sum_{i\in S_t}u_i$.
 \end{enumerate}
\item Return $x_T$.
\end{enumerate}
The epoch costs $\lceil m/s\rceil+2T-1$ strict rounds and
$m+(2T-1)s$ contacts, and each contact carries one $d$-vector each way.
Steps (a) and (b) cannot share a round: $g_t$ depends on $x_t$, which
is known only after the previous master replies, and the analysis
needs $S_t$ independent of $g_t$. A Hessian-vector product, a gradient,
and a shifted-inverse reply are all allowed ZR replies. The server
forms only linear combinations of memory. With $x_0=0$ the method is
therefore ZR. The local proximal form also defines the method for
nonquadratic clients, but the analysis below uses the linearity of the
replies.

\textbf{Notation for one objective $h$.} Let $\mu_h=\mu+\kappa$,
$X_i=A_i+(\kappa+\lambda)I$, $\bar X=A+(\kappa+\lambda)I$,
$H_i=X_i^{-1}$, $\bar P=\frac1m\sum_iH_i$, and $N=\bar P^{-1}$. For a
cohort $S$, let $P_S=\frac1s\sum_{i\in S}H_i$. Then
$x_{t+1}=x_t-P_{S_t}g_t$, the same form as an SVRS step with master
matrix $P_{S_t}^{-1}$, and $\mathbb EP_S=\bar P$. With $x^\star_h$ the
minimizer of $h$ and $e=x-x^\star_h$, the potential is
$\Phi(x)=e^\top Ne$, and $\beta=\bar\delta^2/(\mu_h+\lambda)$.

\subsection{Lemmas}
\textbf{Lemma C.8 (virtual master).} (i) $\bar X^{-1}\preceq\bar P
\preceq\bar X^{-1}+\beta\bar X^{-2}$, hence
$\bar X-\beta I\preceq N\preceq\bar X$.
(ii) With $V=\mathbb E_S(P_S-\bar P)N(P_S-\bar P)$,
\[
 V\preceq\xi\,\kappa_c\,\bar\delta^2\,\bar X^{-2},\qquad
 \kappa_c=\frac1{\mu_h+\lambda}\Bigl(1+\frac{\delta_{\max}}{\mu_h+\lambda}\Bigr).
\]

\textit{Proof.} (i) Since $X_i=\bar X+D_i$, the resolvent identity
gives $H_i=\bar X^{-1}-\bar X^{-1}D_iH_i=\bar X^{-1}-H_iD_i\bar X^{-1}$.
Substituting the second form into the first,
\[
 H_i=\bar X^{-1}-\bar X^{-1}D_i\bar X^{-1}
 +\bar X^{-1}D_iH_iD_i\bar X^{-1}.
\]
Average over $i$. Since $\sum_iD_i=0$,
$\bar P=\bar X^{-1}+\bar X^{-1}\bigl(\frac1m\sum_iD_iH_iD_i\bigr)\bar X^{-1}$.
Each $D_iH_iD_i\succeq0$, which gives the lower bound. Since
$H_i\preceq(\mu_h+\lambda)^{-1}I$,
$\frac1m\sum_iD_iH_iD_i\preceq\frac1{\mu_h+\lambda}\frac1m\sum_iD_i^2
\preceq\beta I$, which gives the upper bound. Inversion reverses the
order. $\bar X$ commutes with $(\bar X+\beta I)\bar X^{-2}$, so
$N\succeq\bar X^2(\bar X+\beta I)^{-1}\succeq\bar X-\beta I$, because
$x^2/(x+\beta)\geq x-\beta$. Also $N\preceq\bar X$.

(ii) For symmetric $Y_1,\ldots,Y_m$ with $\sum_iY_i=0$ and any
$M\succeq0$, the indicator computation behind \eqref{eq:app-fpc} gives
the matrix identity
$\mathbb E\bigl(\frac1s\sum_{i\in S}Y_i\bigr)M\bigl(\frac1s\sum_{i\in S}Y_i\bigr)
=\xi\cdot\frac1m\sum_iY_iMY_i$. The cross terms carry the same
indicator moments, and $\sum_{i\ne j}Y_iMY_j=-\sum_iY_iMY_i$.
Take $Y_i=H_i-\bar P$ and $M=N$. For any symmetric $C$,
$\frac1m\sum_i(H_i-C)N(H_i-C)=\frac1m\sum_iY_iNY_i+(\bar P-C)N(\bar P-C)$,
so $C=\bar X^{-1}$ gives an upper bound. Now
$H_i-\bar X^{-1}=-\bar X^{-1}D_iH_i=-H_iD_i\bar X^{-1}$, so
\[
 (H_i-\bar X^{-1})N(H_i-\bar X^{-1})
 =\bar X^{-1}D_i\,(H_iNH_i)\,D_i\bar X^{-1}.
\]
By (i), $H_iNH_i\preceq H_i\bar XH_i=H_i(X_i-D_i)H_i=H_i-H_iD_iH_i
\preceq\kappa_cI$. Hence
$V\preceq\xi\kappa_c\bar X^{-1}\bigl(\frac1m\sum_iD_i^2\bigr)\bar X^{-1}
\preceq\xi\kappa_c\bar\delta^2\bar X^{-2}$. $\square$

\textbf{Lemma C.9 (one step).} Let $x$ be the current point of an epoch
with snapshot $w$, let $\zeta=\frac1s\sum_{i\in S'}D_i(x-w)$ for an
independent uniform $s$-cohort $S'$ (with $\zeta=0$ at $t=0$), and let
$x^+=x-P_S(\nabla h(x)+\zeta)$. Then $g=\nabla h(x)+\zeta$ is the
$g_t$ of the method, and, with $A_h=A+\kappa I$,
\[
 \mathbb E\,\Phi(x^+)=\Phi(x)-2e^\top A_he
 +e^\top A_h(\bar P+V)A_he+\mathbb E\,\zeta^\top(\bar P+V)\zeta,\qquad
 \mathbb E\|\zeta\|^2\leq\xi\bar\delta^2\|x-w\|^2.
\]

\textit{Proof.} $\nabla h(w)+\frac1s\sum_{S'}(A_i+\kappa I)(x-w)
=\nabla h(x)+\zeta$, because $\nabla h$ is affine with Hessian $A_h$.
Write $e^+=e-P_Sg$. Then
$\Phi(x^+)=\Phi(x)-2g^\top P_SNe+g^\top P_SNP_Sg$. Taking the expectation
over $S$, which is independent of $g$, and using
$\mathbb EP_SNP_S=\bar PN\bar P+V=\bar P+V$ and $\bar PN=I$, gives
$\Phi(x)-2g^\top e+g^\top(\bar P+V)g$. Now take the expectation over
$S'$, with $g=A_he+\zeta$ and $\mathbb E\zeta=0$. The bound on $\zeta$ is
\eqref{eq:app-fpc} with $u_i=D_i(x-w)$ and $\frac1m\sum_i\|D_iv\|^2\leq
\bar\delta^2\|v\|^2$. $\square$

\subsection{Theorem 14: epoch contraction}\label{app:vm-theorem}
\textbf{Theorem 14 (VM-SVRS).} Let every $A_i\succeq\mu I$, $\kappa\geq0$,
and $\mu_h=\mu+\kappa$. Suppose
\begin{equation}\label{eq:vm-conditions}
 \lambda\geq\max\{\mu_h,\ 4\bar\delta,\ 4\bar\delta\sqrt{\xi T}\},
 \qquad T\geq\max\{\lceil m/s\rceil,\ 2(\mu_h+\lambda)/\mu_h\}.
\end{equation}
Then one epoch from any $w$ satisfies
$\mathbb E\,\Phi(x_T)\leq0.28\,\Phi(w)$. Moreover
$h(x)-h^\star\leq\frac8{15}\Phi(x)$ and
$\Phi(x)\leq\frac{2(\mu_h+\lambda)}{\mu_h}(h(x)-h^\star)$. After $K$
epochs from $x_0$, therefore,
$\mathbb E[h(x)-h^\star]\leq\frac{16(\mu_h+\lambda)}{15\mu_h}0.28^K
(h(x_0)-h^\star)$. The method uses neither $\delta_a$ nor an anchor.

\textit{Proof.} \emph{Constants.} Put $\rho=\mu_h/(\mu_h+\lambda)$.
Since $\lambda\geq4\bar\delta$, $\beta\leq\lambda^2/(16(\mu_h+\lambda))
\leq(\mu_h+\lambda)/16$. By Lemma~C.1, $\delta_{\max}\leq\sqrt m\,\bar\delta$,
and $\lambda\geq4\bar\delta\sqrt{\xi m/s}$ because $T\geq m/s$. Hence
$\xi\delta_{\max}/(\mu_h+\lambda)\leq\sqrt{\xi s}/4\leq\frac14$, using
$\xi s=\frac{m-s}{m-1}\leq1$. So $\xi\kappa_c\bar\delta^2\leq
\xi\beta+\frac14\beta\leq\frac54\beta$, and
\[
 \gamma:=\beta+\xi\kappa_c\bar\delta^2\leq\tfrac94\beta
 \leq\tfrac9{64}(\mu_h+\lambda),\qquad
 \bar P+V\preceq\bar X^{-1}+\gamma\bar X^{-2}.
\]
By Lemma~C.8(i) and $\bar X\succeq(\mu_h+\lambda)I$,
$N\succeq\bar X-\beta I\succeq\frac{15}{16}\bar X$.

\emph{Descent.} $A_h$ and $\bar X=A_h+\lambda I$ commute. For an
eigenvalue $a\geq\mu_h$ of $A_h$,
\[
 2a-\frac{a^2}{a+\lambda}-\frac{\gamma a^2}{(a+\lambda)^2}
 \geq a\Bigl(1-\frac{\gamma}{4\lambda}\Bigr)
 \geq\frac{119}{128}\,a,
\]
because $a/(a+\lambda)\leq1$, $a/(a+\lambda)^2\leq1/(4\lambda)$, and
$\gamma/(4\lambda)\leq\frac9{64}\cdot\frac{2\lambda}{4\lambda}=\frac9{128}$
when $\mu_h\leq\lambda$. So
$2e^\top A_he-e^\top A_h(\bar P+V)A_he\geq0.9297\,e^\top A_he$. Also
$\Phi\leq e^\top\bar Xe\leq(1+\lambda/\mu_h)e^\top A_he$, so
$e^\top A_he\geq\rho\Phi$.

\emph{Noise.} $\|\bar P+V\|\leq(1+\frac9{64})/(\mu_h+\lambda)$ and
$\|x-w\|^2\leq2\|e\|^2+2\|e_w\|^2$ with
$\|e\|^2\leq\frac{16}{15}\Phi/(\mu_h+\lambda)$. The conditions give
$\xi\bar\delta^2\leq\lambda^2/(16T)\leq(\mu_h+\lambda)^2/(16T)$. Hence
\[
 \mathbb E\,\zeta^\top(\bar P+V)\zeta
 \leq\frac{73}{64}\cdot\frac{16}{15}\cdot\frac2{16T}\,
 (\Phi(x)+\Phi(w))\leq\frac{0.1521}T(\Phi(x)+\Phi(w))
 \leq0.0761\,\rho\,(\Phi(x)+\Phi(w)),
\]
using $T\geq2/\rho$.

\emph{Recursion.} Let $a_t=\mathbb E\,\Phi(x_t)$ and $\eta=0.0761\rho$.
Lemma~C.9, the tower property, and the fresh cohorts of each step give
$a_1\leq(1-0.9297\rho)a_0$ (no noise at $t=0$) and, for $t\geq1$,
$a_{t+1}\leq(1-0.8536\rho)a_t+\eta a_0$. Unrolling,
\[
 a_T\leq e^{-0.8536\rho T}a_0+\frac{\eta}{0.8536\rho}a_0
 \leq(e^{-1.707}+0.0892)\,a_0\leq0.271\,a_0,
\]
since $\rho T\geq2$. For the conversions,
$h-h^\star=\frac12e^\top A_he\leq\frac12e^\top\bar Xe\leq\frac8{15}\Phi$,
and $\Phi\leq(1+\lambda/\mu_h)e^\top A_he$. $\square$

By itself, Theorem~14 gives a non-accelerated rate. The condition
$T\geq2(\mu_h+\lambda)/\mu_h$ together with $\lambda\geq4\bar\delta\sqrt{\xi T}$
forces $\lambda\gtrsim\xi\bar\delta^2/\mu$, which is a
$\bar\delta^2/\mu^2$ term. Catalyst removes it. We use Theorem~14 only
as an inner solver, with the shift $\kappa$ chosen so that
\eqref{eq:vm-conditions} holds with $T=O(m/s)$.

\subsection{Corollary 15: Catalyst acceleration}\label{app:vm-catalyst}
\textbf{Parameters.} Set
\[
 T=\max\{\lceil m/s\rceil,4\},\quad
 \lambda=4\max\{\mu,\ \bar\delta,\ \bar\delta\sqrt{\xi T}\},\quad
 \mu'=\max\Bigl\{\mu,\ \frac{2\lambda}{T-2}\Bigr\},\quad
 \kappa=\mu'-\mu,
\]
$q=\mu/\mu'$, $\rho_c=0.9\sqrt q$, $\beta_c=(1-\sqrt q)/(1+\sqrt q)$,
$U=\|\nabla f(x_0)\|^2/(2\mu)\geq f(x_0)-f^\star$, and
$\varepsilon_k=\frac29U(1-\rho_c)^k$.

\textbf{CatVM.} Put $y_0=x_0$. For $k=1,2,\ldots$, let
$h_k=f+\frac\kappa2\|\cdot-y_{k-1}\|^2$. Run VM-SVRS epochs on $h_k$
from the snapshot $w_0=x_{k-1}$. At each snapshot $w_j$, the gradient
$\nabla h_k(w_j)$ has already been collected for the epoch. Stop at the
first $j$ with $\|\nabla h_k(w_j)\|^2\leq2\mu'\varepsilon_k$, and set
$x_k=w_j$ and $y_k=x_k+\beta_c(x_k-x_{k-1})$. The stopping gradient is
reused: $\nabla h_{k+1}(x_k)=\nabla f(x_k)+\kappa(x_k-y_k)$.

\textbf{Corollary 15.} Let every $A_i\succeq\mu I$, put
$\ell=\log(mL_{\max}/\mu)$, and let $\epsilon\in(0,1)$.
\begin{enumerate}\tightlist
\item[(i)] Every $h_k$ satisfies \eqref{eq:vm-conditions} with
$\mu_h=\mu'$.
\item[(ii)] Surely, $f(x_k)-f^\star\leq\frac{800}q(1-\rho_c)^{k+1}U$ for
every $k$. Hence $f(x_K)-f^\star\leq\epsilon(f(x_0)-f^\star)$ once
$K\geq\frac1{0.9\sqrt q}\log\frac{800L}{q\mu\epsilon}$.
\item[(iii)] Conditional on the past, outer step $k$ runs
$J_k$ epochs with $\mathbb EJ_k\leq2+\log_{1/0.28}C_2$, where
$C_2=\frac{8L(\mu'+\lambda)}{15\mu\mu'}\cdot144000\bigl(\frac L\mu
+\frac4{q^2}\bigr)$, so $\mathbb EJ_k=O(\ell)$.
\item[(iv)] The expected number of strict rounds to reach
$f(x_K)-f^\star\leq\epsilon(f(x_0)-f^\star)$ is
\begin{equation}\label{eq:vm-rate}
 \bar R=O\Bigl(\Bigl[\Bigl\lceil\frac ms\Bigr\rceil
 +\Bigl(\sqrt{\Bigl\lceil\frac ms\Bigr\rceil}+\frac{m^{3/4}}s\Bigr)
 \sqrt{\frac{\bar\delta}\mu}\Bigr]\,
 \ell\,\Bigl(\ell+\log\frac1\epsilon\Bigr)\Bigr).
\end{equation}
\item[(v)] With $x_0=0$ the method is ZR, and it makes no separate
anchor contact. The anchor can appear in a sampled cohort. The rate
does not depend on $\delta_a$.
\end{enumerate}

\textit{Proof.} (i) $h_k$ has client Hessians $A_i+\kappa I\succeq\mu'I$ and
the same $D_i$, hence the same $\bar\delta$, $\delta_{\max}$, and $\xi$.
$\lambda\geq4\max\{\bar\delta,\bar\delta\sqrt{\xi T}\}$ by definition.
$\lambda\geq\mu'$, because $\lambda\geq4\mu$ and $2\lambda/(T-2)\leq\lambda$
for $T\geq4$. $T\geq\lceil m/s\rceil$, and
$2(\mu'+\lambda)/\mu'=2+2\lambda/\mu'\leq2+(T-2)=T$ because
$\mu'\geq2\lambda/(T-2)$.

(ii) If $\|\nabla h_k(x_k)\|^2\leq2\mu'\varepsilon_k$, then
$h_k(x_k)-h_k^\star\leq\varepsilon_k$ by $\mu'$-strong convexity. This
is criterion~(C1) of \citet{lin2018catalyst}. Their Algorithm~2 with
$\alpha_0=\sqrt q$ keeps $\alpha_k=\sqrt q$, so its momentum is
$\beta_c$. Their Theorem~3 bounds $f(x_k)-f^\star$ by
$A_{k-1}\bigl(\sqrt{S_0}+3\sum_{j\leq k}\sqrt{\varepsilon_j/A_{j-1}}\bigr)^2$,
where $A_j\leq(1-\sqrt q)^j$ and
$S_0=(1-\sqrt q)(f(x_0)-f^\star)+\frac{\gamma_0}2\|x_0-x^\star\|^2$
with $\gamma_0=\mu(1-\sqrt q)$. Hence $S_0\leq2U$. Their
Proposition~5 evaluates this bound for $\varepsilon_k=\frac29(f(x_0)-f^\star)(1-\rho_c)^k$,
$\rho_c<\sqrt q$, using only $S_0\leq2(f(x_0)-f^\star)$. The bound is
homogeneous of degree one in $(S_0,\varepsilon_1,\varepsilon_2,\ldots)$,
so the same computation with $U$ gives
$f(x_k)-f^\star\leq\frac8{(\sqrt q-\rho_c)^2}(1-\rho_c)^{k+1}U
=\frac{800}q(1-\rho_c)^{k+1}U$. The criterion is checked exactly, so
this holds surely. Finally, $U\leq(L/\mu)(f(x_0)-f^\star)$ and
$(1-\rho_c)^{K+1}\leq e^{-0.9\sqrt qK}$.

(iii) Write $F_j=f(x_j)-f^\star$. By (ii),
$F_j\leq\frac{800}qU(1-\rho_c)^{j+1}=\frac{3600}q(1-\rho_c)^{j+1-k}\varepsilon_k$.
So $F_{k-1}\leq\frac{3600}q\varepsilon_k$ and, since $1-\rho_c\geq0.1$,
$F_{k-2}\leq\frac{36000}q\varepsilon_k$. For $k\geq2$,
$x_{k-1}-y_{k-1}=-\beta_c(x_{k-1}-x_{k-2})$ with $\beta_c\leq1$, so
\[
 \|\nabla h_k(x_{k-1})\|^2\leq2\|\nabla f(x_{k-1})\|^2
 +\frac{4\kappa^2}\mu\cdot2(F_{k-1}+F_{k-2})
 \leq\Bigl(4L+\frac{16\kappa^2}\mu\Bigr)\frac{36000}q\varepsilon_k
 \leq C_1\mu'\varepsilon_k,
\]
with $C_1=144000(L/\mu+4/q^2)$, using $\kappa\leq\mu'=\mu/q$. For
$k=1$, $\|\nabla h_1(x_0)\|^2=2\mu U=9\mu\varepsilon_1/(1-\rho_c)
\leq C_1\mu'\varepsilon_1$. Let $e$ be the error with respect to
$x^\star_{h_k}$ and $A'=A+\kappa I$. Then
$\Phi_k(w_0)\leq e^\top(A'+\lambda I)e\leq\frac{\mu'+\lambda}{\mu'^2}
\|\nabla h_k(w_0)\|^2$, and
$\|\nabla h_k(w)\|^2\leq\|A'\|e^\top A'e\leq\frac{16}{15}(L+\kappa)\Phi_k(w)$.
Continue the epochs past the stopping time, and let $w_j$ be the $j$th
snapshot. Theorem~14 gives $\mathbb E\Phi_k(w_j)\leq0.28^j\Phi_k(w_0)$.
By Markov's inequality, and with $(L+\kappa)/\mu'\leq L/\mu$,
\[
 \Pr(J_k>j)\leq\Pr\bigl(\|\nabla h_k(w_j)\|^2>2\mu'\varepsilon_k\bigr)
 \leq\frac{8(L+\kappa)(\mu'+\lambda)}{15\mu'^2}C_1\,0.28^j\leq C_2\,0.28^j.
\]
Summing $\min\{1,C_2\,0.28^j\}$ over $j\geq0$ gives
$\mathbb EJ_k\leq\lceil\log_{1/0.28}C_2\rceil+1/0.72$. For the size of
$C_2$: $(\mu'+\lambda)/\mu'\leq T/2\leq m+2$ by (i);
$1/q=\mu'/\mu\leq\max\{1,\lambda/\mu\}$; $\bar\delta\leq\max_i\|D_i\|\leq
L_{\max}$; and $\xi T\leq m+4$. Hence $\lambda/\mu\leq4(L_{\max}/\mu)\sqrt{m+4}$,
and $\log C_2=O(\ell)$.

(iv) The first outer step adds one gradient pass. Each outer step then
costs at most $J_k(\lceil m/s\rceil+2T-1)\leq3TJ_k$ rounds. By the tower
property, $\bar R\leq\lceil m/s\rceil+3TK\max_k\mathbb E[J_k\mid\text{past}]$
with $K$ from (ii), and $\log\frac{800L}{q\mu\epsilon}=O(\ell+\log\frac1\epsilon)$.
It remains to bound $T/\sqrt q$:
\[
 \frac T{\sqrt q}\leq T+T\sqrt{\frac{2\lambda}{\mu(T-2)}}
 \leq T+2\sqrt{\frac{T\lambda}\mu}
 \leq T+4\Bigl(\sqrt T+\sqrt{\frac{T\bar\delta}\mu}
 +(\xi T)^{1/4}\sqrt{\frac{T\bar\delta}\mu}\Bigr),
\]
using $T/(T-2)\leq2$ and $\sqrt{\max\{a,b,c\}}\leq\sqrt a+\sqrt b+\sqrt c$.
Now $T\leq5\lceil m/s\rceil$ and $\sqrt T\leq T$. If $s\leq m/4$, then
$T=\lceil m/s\rceil\leq2m/s$ and $\xi\leq1/s$, so
$T^{3/4}\xi^{1/4}\leq2m^{3/4}/s$. If $s>m/4$, then $T=4$ and
$\xi<4/m$, so $T^{3/4}\xi^{1/4}\leq4m^{-1/4}\leq4m^{3/4}/s$. This gives
\eqref{eq:vm-rate}.

(v) The replies are a gradient, Hessian-vector products, and shifted
inverses $(A_i+(\kappa+\lambda)I)^{-1}g$. Catalyst's $y_k$, the shifts
$\kappa(x-y_{k-1})$, and the stopping test are linear combinations of
memory, and the stopping rule is adaptive, as the lower bounds allow.
$\square$

\subsection{Where the bounds now match}\label{app:vm-gap}
Take the setting of the paragraph ``Where the bounds match'' in
Appendix~\ref{app:large-cohorts}: quadratic clients with $A_i\succeq\mu I$,
$\min\{\delta_a,\bar\delta/\sqrt m\}\geq3\mu$, and high accuracy.
The comparison below concerns polynomial dependence only: it suppresses
constants and logarithmic factors. Corollary~15 carries
$\ell(\ell+\log(1/\epsilon))$, while the lower bound carries
$\log(1/\epsilon)$. Only the poor-anchor case
$\delta_a>\sqrt m\,\bar\delta/s$
changes. Theorem~10 costs
$\frac ms+\sqrt{\frac ms}\sqrt{\frac{\delta_a}\mu}
+\frac{m^{3/4}}s\sqrt{\frac{\bar\delta}\mu}$, and Corollary~15 costs
the same expression with $\delta_a$ replaced by $\bar\delta$. Their
minimum is therefore
\[
 \frac ms+\frac{m^{3/4}}s\sqrt{\frac{\bar\delta}\mu}
 +\sqrt{\frac ms}\sqrt{\frac{\min\{\delta_a,\bar\delta\}}\mu}.
\]
The first two terms are at most a constant times the lower bound
\eqref{eq:all-s-lower}. Here $\delta_a>\bar\delta/\sqrt m$, so the
second term of \eqref{eq:all-s-lower} is
$\frac ms\sqrt{\bar\delta/(\sqrt m\,\mu)}=\frac{m^{3/4}}s\sqrt{\bar\delta/\mu}$.
The last term is at most $\sqrt s/m^{1/4}$ times that second term, and
at most $\sqrt{m/s}$ times the third term
$\sqrt{\min\{\delta_a,\bar\delta\}/\mu}$ (up to the constant
$\sqrt3$). Hence the ratio of the polynomial parts of the best upper and lower
bounds is
\[
 O\Bigl(1+\min\Bigl\{\frac{\sqrt s}{m^{1/4}},\ \sqrt{\frac ms}\Bigr\}\Bigr)
 \leq O(m^{1/8}),
\]
with equality of the two terms inside the minimum at $s=m^{3/4}$.
The polynomial parts match up to constants when $s\leq\sqrt m$ or
$s=\Omega(m)$. The full logarithmic ratio for the Catalyst upper bound
is $\ell(\ell+\log(1/\epsilon))/\log(1/\epsilon)$.
When $\log(1/\epsilon)\geq\ell$, this overhead is $O(\ell)$.
Before Corollary~15 the gap was
$O(\min\{\sqrt s,m^{1/4},m/s\})$. Every factor
improves: $\sqrt s/m^{1/4}\leq\sqrt s$, $\sqrt s/m^{1/4}\leq m^{1/4}$,
and $\sqrt{m/s}\leq m/s$. Corollary~15 also dominates Proposition~13 in
polynomial dependence, since $\sqrt{m/s}\leq m/s$ and $m^{3/4}/s\leq m/s$. The
hard instances of Theorem~7 and Lemma~C.7 are quadratic with
$\mu$-strongly convex clients, so the lower bounds apply to the same
class. If clients may be nonconvex, Corollary~15 is unavailable and the
gap $O(m^{3/8})$ of Appendix~\ref{app:large-cohorts} stands.

\textbf{Limitations.} The analysis needs quadratic clients, because the
virtual master is the average of linear proximal maps, and it needs
$\mu$-strongly convex clients. Catalyst adds the logarithmic overhead
stated in Corollary~15, and
the mean-square constants are conservative: Section~\ref{app:vm-numerics}
finds the best fixed $\lambda$ at $1/4$ of the base scale or below
it ($1/8$ to $1/4$ on the dip family; on the random families the
rounds level off as $\lambda$ decreases). The base scale is
$\max\{\mu,\bar\delta,\bar\delta\sqrt{\xi\lceil m/s\rceil}\}$,
while Corollary~15 uses $4$ times it.

\subsection{Numerical checks}\label{app:vm-numerics}
Exact numerical checks of Lemmas~C.8--C.9 use random federations with
$\mu$-strongly convex clients. The worst violation of Lemma~C.8(i)
is $9.7\times10^{-16}$;
for Lemma~C.8(ii) the identity error is $5\times10^{-18}$ with no bound
violation; and the relative error of Lemma~C.9 is $1.6\times10^{-13}$.
At the parameters of \eqref{eq:vm-conditions}, it then computes the
exact worst-case expected epoch contraction over snapshots by
enumerating all cohorts. The worst value found is $0.007$, against the
bound $0.28$. Further checks establish two facts. At $s=m$, VM-SVRS is
SVRS with the deterministic master $\bar P^{-1}$. On a ``dip'' family,
where the anchor's Hessian
departs from the mean in a subspace of the other clients' Hessians, the
anchor master diverges and the virtual master does not.

A CPU pilot ($m=16$, $d=32$, $\bar\delta=10$, $\mu=1$, four seeds, relative
gap $10^{-6}$) uses the strict ledger for every method. Each tuned
method takes the best of a common ten-point multiplier grid
($1/256$ to $2$ times its own base scale). On the dip family, tuned
VM-SVRS is fastest or tied at every $s\in\{2,4,16\}$: $63$, $34$ and
$14$ rounds, against $74$, $38$ and $14$ for tuned cohort SVRP,
$299$, $156$ and $153$ for tuned APSVRG, and $368$, $290$ and $15$ for
tuned S-DANE. With the anchor as master, the same method does not reach
the target at $s\in\{4,16\}$. On the random poor-, mid- and good-anchor
families, tuned SVRP is fastest or tied. Tuned VM-SVRS is within a
factor $1.3$--$1.6$ at $s=2$ and ties at $s=16$. There, every
prox-type method is best at the small-$\lambda$ end of the grid, where
the rounds level off. VM-SVRS then pays for its two rounds per inner
step, against one for SVRP. The value of the new method is
therefore its anchor-free guarantee and its behaviour when every client
departs from the mean. It is not a uniform speed-up over tuned SVRP.
The theorem-parameter Catalyst arm is conservative, as expected from
its constants.

\paragraph{Full GPU study and paired intervals.}
The fresh fp64 run uses $m\in\{64,256\}$, $d=128$,
$s\in\{4,16,64,m\}$, $\bar\delta\in\{10,100\}$, and 12 seeds for
each of four families. Its seven distinct $(m,d,s)$ units cover 56
family--parameter cells, with a 20,000-round cap and a relative-gap
target of $10^{-6}$. The saved summary records 11,385.3 seconds of wall
time on two RTX~4090 GPUs and 7.69 GB maximum allocated device memory.
The 51,552 per-seed first-hit rows reproduce every hit count and
complete-seed median in the 4,296 aggregate rows. All eight evaluation
seeds contribute to both comparisons in every cell below, so no paired
seed is dropped.

\input{vm_ratio_plot}

\begin{table*}[t]
\centering\small
\caption{Fresh 12-seed VM GPU study at relative gap $10^{-6}$, with
four tuning seeds and eight held-out evaluation seeds.
Entries are medians across cells of paired geometric-mean round ratios
on evaluation seeds 4--11, with 95\% two-level bootstrap intervals
(2,000 resamples). Multipliers were selected on seeds 0--3.
Ratios below one favor VM or CatVM, respectively. For
$s\in\{4,16,64\}$ each family has four cells; for $s=256$ it has
two. Intervals describe the selected synthetic grid and are not
adjusted for comparisons across cells.}
\label{tab:vm-gpu-ci}
\begin{tabular}{lrrll}
\toprule
Family & $s$ & Cells & Tuned VM / tuned SVRP & CatVM-thm / CAccSVRS-thm\\
\midrule
poor & 4 & 4 & 1.32 [1.31,1.33] & 0.70 [0.65,0.82]\\
poor & 16 & 4 & 1.27 [1.23,1.31] & 0.40 [0.30,0.54]\\
poor & 64 & 4 & 1.11 [1.00,1.25] & 0.45 [0.33,0.66]\\
poor & 256 & 2 & 1.00 [1.00,1.00] & 0.39 [0.37,0.42]\\
mid & 4 & 4 & 1.24 [1.12,1.48] & 1.73 [1.57,1.95]\\
mid & 16 & 4 & 1.25 [1.14,1.34] & 1.73 [1.59,1.86]\\
mid & 64 & 4 & 1.00 [0.85,1.23] & 1.81 [1.60,2.02]\\
mid & 256 & 2 & 1.00 [1.00,1.00] & 1.94 [1.80,2.10]\\
good & 4 & 4 & 1.40 [1.24,1.57] & 1.73 [1.55,1.86]\\
good & 16 & 4 & 1.25 [1.20,1.31] & 2.31 [1.72,3.15]\\
good & 64 & 4 & 1.00 [0.85,1.23] & 5.79 [4.73,7.37]\\
good & 256 & 2 & 1.00 [1.00,1.00] & 9.14 [7.33,11.41]\\
dip & 4 & 4 & 0.91 [0.75,1.06] & 0.92 [0.66,1.37]\\
dip & 16 & 4 & 0.74 [0.62,0.84] & 0.52 [0.49,0.56]\\
dip & 64 & 4 & 0.84 [0.66,1.00] & 0.60 [0.57,0.65]\\
dip & 256 & 2 & 1.00 [1.00,1.00] & 0.71 [0.66,0.77]\\
\bottomrule
\end{tabular}
\end{table*}

With incomplete participation, the tuned VM/SVRP paired
geometric-mean round ratio is below one in nine of ten dip cells,
no poor-anchor cells, and one cell each in the mid- and good-anchor
families. The 16 full-participation cells are ties. The fixed theorem-parameter
CatVM/CAccSVRS ratio is below one in all 14 poor-anchor cells and 12
dip cells; it is above one in every mid- and good-anchor cell. The
pooled intervals in Table~\ref{tab:vm-gpu-ci} are descriptive for this
synthetic grid and do not correct for multiple cell comparisons. Among
the 40 incomplete-participation cells, the selected VM and SVRP
multipliers are at the smallest tested value in 25 and 23 cells,
respectively; ``tuned'' means best on the tested grid. Other baselines
appear in the full study report, so these two ratios do not identify
the globally fastest method. Per-seed first-hit records and paired
intervals are archived; full gap trajectories are retained separately.

%% file: vm_ratio_plot.tex
\begin{figure*}[t]
\centering
\begin{tikzpicture}[x=1cm,y=1cm,font=\footnotesize]
\definecolor{vmblue}{HTML}{0072B2}
\definecolor{vmorange}{HTML}{D55E00}
\begin{scope}[shift={(0.000,0.000)}]
\node[font=\small\bfseries,anchor=west] at (0.7,3.75) {Dip family};
\draw[black!12,line width=0.35pt] (0.760,0.746) -- (6.500,0.746);
\node[anchor=east] at (0.680,0.746) {0.5};
\draw[densely dashed,black!55,line width=0.35pt] (0.760,1.725) -- (6.500,1.725);
\node[anchor=east] at (0.680,1.725) {1};
\draw[black!12,line width=0.35pt] (0.760,2.705) -- (6.500,2.705);
\node[anchor=east] at (0.680,2.705) {1.5};
\draw[black!75,line width=0.55pt] (0.760,0.550) -- (0.760,3.630);
\draw[black!75,line width=0.55pt] (0.76,0.55) -- (6.50,0.55);
\node[anchor=north] at (1.380,0.490) {4};
\node[anchor=north] at (3.200,0.490) {16};
\node[anchor=north] at (5.020,0.490) {64};
\node[anchor=north] at (3.63,0.09) {Cohort size $s$};
\draw[vmblue,line width=0.65pt] (1.150,1.544) -- (1.150,2.181);
\draw[vmblue,line width=0.65pt] (1.105,1.544) -- (1.195,1.544);
\draw[vmblue,line width=0.65pt] (1.105,2.181) -- (1.195,2.181);
\filldraw[draw=vmblue,fill=vmblue] (1.150,1.824) circle (1.45pt);
\draw[vmblue,line width=0.65pt] (2.970,1.096) -- (2.970,1.461);
\draw[vmblue,line width=0.65pt] (2.925,1.096) -- (3.015,1.096);
\draw[vmblue,line width=0.65pt] (2.925,1.461) -- (3.015,1.461);
\filldraw[draw=vmblue,fill=vmblue] (2.970,1.297) circle (1.45pt);
\draw[vmorange,line width=0.65pt] (1.455,1.497) -- (1.455,1.964);
\draw[vmorange,line width=0.65pt] (1.410,1.497) -- (1.500,1.497);
\draw[vmorange,line width=0.65pt] (1.410,1.964) -- (1.500,1.964);
\filldraw[draw=vmorange,fill=vmorange] (1.455,1.708) circle (1.45pt);
\draw[vmorange,line width=0.65pt] (3.275,0.924) -- (3.275,1.076);
\draw[vmorange,line width=0.65pt] (3.230,0.924) -- (3.320,0.924);
\draw[vmorange,line width=0.65pt] (3.230,1.076) -- (3.320,1.076);
\filldraw[draw=vmorange,fill=vmorange] (3.275,0.975) circle (1.45pt);
\draw[vmorange,line width=0.65pt] (5.095,0.924) -- (5.095,1.166);
\draw[vmorange,line width=0.65pt] (5.050,0.924) -- (5.140,0.924);
\draw[vmorange,line width=0.65pt] (5.050,1.166) -- (5.140,1.166);
\filldraw[draw=vmorange,fill=vmorange] (5.095,1.048) circle (1.45pt);
\draw[vmblue,line width=0.65pt] (1.305,1.189) -- (1.305,1.671);
\draw[vmblue,line width=0.65pt] (1.260,1.189) -- (1.350,1.189);
\draw[vmblue,line width=0.65pt] (1.260,1.671) -- (1.350,1.671);
\draw[vmblue,line width=0.85pt,fill=white] (1.250,1.337) rectangle (1.360,1.447);
\draw[vmblue,line width=0.65pt] (3.125,1.097) -- (3.125,1.657);
\draw[vmblue,line width=0.65pt] (3.080,1.097) -- (3.170,1.097);
\draw[vmblue,line width=0.65pt] (3.080,1.657) -- (3.170,1.657);
\draw[vmblue,line width=0.85pt,fill=white] (3.070,1.307) rectangle (3.180,1.417);
\draw[vmorange,line width=0.65pt] (1.610,1.159) -- (1.610,1.281);
\draw[vmorange,line width=0.65pt] (1.565,1.159) -- (1.655,1.159);
\draw[vmorange,line width=0.65pt] (1.565,1.281) -- (1.655,1.281);
\draw[vmorange,line width=0.85pt,fill=white] (1.555,1.173) rectangle (1.665,1.283);
\draw[vmorange,line width=0.65pt] (3.430,0.978) -- (3.430,1.350);
\draw[vmorange,line width=0.65pt] (3.385,0.978) -- (3.475,0.978);
\draw[vmorange,line width=0.65pt] (3.385,1.350) -- (3.475,1.350);
\draw[vmorange,line width=0.85pt,fill=white] (3.375,1.082) rectangle (3.485,1.192);
\draw[vmorange,line width=0.65pt] (5.250,1.132) -- (5.250,1.184);
\draw[vmorange,line width=0.65pt] (5.205,1.132) -- (5.295,1.132);
\draw[vmorange,line width=0.65pt] (5.205,1.184) -- (5.295,1.184);
\draw[vmorange,line width=0.85pt,fill=white] (5.195,1.102) rectangle (5.305,1.212);
\end{scope}
\begin{scope}[shift={(7.150,0.000)}]
\node[font=\small\bfseries,anchor=west] at (0.7,3.75) {Poor anchor};
\draw[black!12,line width=0.35pt] (0.760,0.746) -- (6.500,0.746);
\node[anchor=east] at (0.680,0.746) {0.5};
\draw[densely dashed,black!55,line width=0.35pt] (0.760,1.725) -- (6.500,1.725);
\node[anchor=east] at (0.680,1.725) {1};
\draw[black!12,line width=0.35pt] (0.760,2.705) -- (6.500,2.705);
\node[anchor=east] at (0.680,2.705) {1.5};
\draw[black!75,line width=0.55pt] (0.760,0.550) -- (0.760,3.630);
\draw[black!75,line width=0.55pt] (0.76,0.55) -- (6.50,0.55);
\node[anchor=north] at (1.380,0.490) {4};
\node[anchor=north] at (3.200,0.490) {16};
\node[anchor=north] at (5.020,0.490) {64};
\node[anchor=north] at (3.63,0.09) {Cohort size $s$};
\draw[vmblue,line width=0.65pt] (1.150,2.325) -- (1.150,2.325);
\draw[vmblue,line width=0.65pt] (1.105,2.325) -- (1.195,2.325);
\draw[vmblue,line width=0.65pt] (1.105,2.325) -- (1.195,2.325);
\filldraw[draw=vmblue,fill=vmblue] (1.150,2.325) circle (1.45pt);
\draw[vmblue,line width=0.65pt] (2.970,2.178) -- (2.970,2.178);
\draw[vmblue,line width=0.65pt] (2.925,2.178) -- (3.015,2.178);
\draw[vmblue,line width=0.65pt] (2.925,2.178) -- (3.015,2.178);
\filldraw[draw=vmblue,fill=vmblue] (2.970,2.178) circle (1.45pt);
\draw[vmorange,line width=0.65pt] (1.455,2.365) -- (1.455,2.365);
\draw[vmorange,line width=0.65pt] (1.410,2.365) -- (1.500,2.365);
\draw[vmorange,line width=0.65pt] (1.410,2.365) -- (1.500,2.365);
\filldraw[draw=vmorange,fill=vmorange] (1.455,2.365) circle (1.45pt);
\draw[vmorange,line width=0.65pt] (3.275,2.325) -- (3.275,2.325);
\draw[vmorange,line width=0.65pt] (3.230,2.325) -- (3.320,2.325);
\draw[vmorange,line width=0.65pt] (3.230,2.325) -- (3.320,2.325);
\filldraw[draw=vmorange,fill=vmorange] (3.275,2.325) circle (1.45pt);
\draw[vmorange,line width=0.65pt] (5.095,2.033) -- (5.095,2.338);
\draw[vmorange,line width=0.65pt] (5.050,2.033) -- (5.140,2.033);
\draw[vmorange,line width=0.65pt] (5.050,2.338) -- (5.140,2.338);
\filldraw[draw=vmorange,fill=vmorange] (5.095,2.193) circle (1.45pt);
\draw[vmblue,line width=0.65pt] (1.305,2.306) -- (1.305,2.355);
\draw[vmblue,line width=0.65pt] (1.260,2.306) -- (1.350,2.306);
\draw[vmblue,line width=0.65pt] (1.260,2.355) -- (1.350,2.355);
\draw[vmblue,line width=0.85pt,fill=white] (1.250,2.274) rectangle (1.360,2.384);
\draw[vmblue,line width=0.65pt] (3.125,2.178) -- (3.125,2.178);
\draw[vmblue,line width=0.65pt] (3.080,2.178) -- (3.170,2.178);
\draw[vmblue,line width=0.65pt] (3.080,2.178) -- (3.170,2.178);
\draw[vmblue,line width=0.85pt,fill=white] (3.070,2.123) rectangle (3.180,2.233);
\draw[vmorange,line width=0.65pt] (1.610,2.365) -- (1.610,2.365);
\draw[vmorange,line width=0.65pt] (1.565,2.365) -- (1.655,2.365);
\draw[vmorange,line width=0.65pt] (1.565,2.365) -- (1.655,2.365);
\draw[vmorange,line width=0.85pt,fill=white] (1.555,2.310) rectangle (1.665,2.420);
\draw[vmorange,line width=0.65pt] (3.430,2.325) -- (3.430,2.325);
\draw[vmorange,line width=0.65pt] (3.385,2.325) -- (3.475,2.325);
\draw[vmorange,line width=0.65pt] (3.385,2.325) -- (3.475,2.325);
\draw[vmorange,line width=0.85pt,fill=white] (3.375,2.270) rectangle (3.485,2.380);
\draw[vmorange,line width=0.65pt] (5.250,2.178) -- (5.250,2.178);
\draw[vmorange,line width=0.65pt] (5.205,2.178) -- (5.295,2.178);
\draw[vmorange,line width=0.65pt] (5.205,2.178) -- (5.295,2.178);
\draw[vmorange,line width=0.85pt,fill=white] (5.195,2.123) rectangle (5.305,2.233);
\end{scope}
\begin{scope}[shift={(0.000,-4.800)}]
\node[font=\small\bfseries,anchor=west] at (0.7,3.75) {Mid anchor};
\draw[black!12,line width=0.35pt] (0.760,0.746) -- (6.500,0.746);
\node[anchor=east] at (0.680,0.746) {0.5};
\draw[densely dashed,black!55,line width=0.35pt] (0.760,1.725) -- (6.500,1.725);
\node[anchor=east] at (0.680,1.725) {1};
\draw[black!12,line width=0.35pt] (0.760,2.705) -- (6.500,2.705);
\node[anchor=east] at (0.680,2.705) {1.5};
\draw[black!75,line width=0.55pt] (0.760,0.550) -- (0.760,3.630);
\draw[black!75,line width=0.55pt] (0.76,0.55) -- (6.50,0.55);
\node[anchor=north] at (1.380,0.490) {4};
\node[anchor=north] at (3.200,0.490) {16};
\node[anchor=north] at (5.020,0.490) {64};
\node[anchor=north] at (3.63,0.09) {Cohort size $s$};
\draw[vmblue,line width=0.65pt] (1.150,1.570) -- (1.150,2.703);
\draw[vmblue,line width=0.65pt] (1.105,1.570) -- (1.195,1.570);
\draw[vmblue,line width=0.65pt] (1.105,2.703) -- (1.195,2.703);
\filldraw[draw=vmblue,fill=vmblue] (1.150,2.039) circle (1.45pt);
\draw[vmblue,line width=0.65pt] (2.970,1.762) -- (2.970,2.178);
\draw[vmblue,line width=0.65pt] (2.925,1.762) -- (3.015,1.762);
\draw[vmblue,line width=0.65pt] (2.925,2.178) -- (3.015,2.178);
\filldraw[draw=vmblue,fill=vmblue] (2.970,2.016) circle (1.45pt);
\draw[vmorange,line width=0.65pt] (1.455,2.138) -- (1.455,3.278);
\draw[vmorange,line width=0.65pt] (1.410,2.138) -- (1.500,2.138);
\draw[vmorange,line width=0.65pt] (1.410,3.278) -- (1.500,3.278);
\filldraw[draw=vmorange,fill=vmorange] (1.455,2.609) circle (1.45pt);
\draw[vmorange,line width=0.65pt] (3.275,2.325) -- (3.275,2.325);
\draw[vmorange,line width=0.65pt] (3.230,2.325) -- (3.320,2.325);
\draw[vmorange,line width=0.65pt] (3.230,2.325) -- (3.320,2.325);
\filldraw[draw=vmorange,fill=vmorange] (3.275,2.325) circle (1.45pt);
\draw[vmorange,line width=0.65pt] (5.095,2.178) -- (5.095,2.178);
\draw[vmorange,line width=0.65pt] (5.050,2.178) -- (5.140,2.178);
\draw[vmorange,line width=0.65pt] (5.050,2.178) -- (5.140,2.178);
\filldraw[draw=vmorange,fill=vmorange] (5.095,2.178) circle (1.45pt);
\draw[vmblue,line width=0.65pt] (1.305,2.032) -- (1.305,2.644);
\draw[vmblue,line width=0.65pt] (1.260,2.032) -- (1.350,2.032);
\draw[vmblue,line width=0.65pt] (1.260,2.644) -- (1.350,2.644);
\draw[vmblue,line width=0.85pt,fill=white] (1.250,2.281) rectangle (1.360,2.391);
\draw[vmblue,line width=0.65pt] (3.125,1.934) -- (3.125,2.285);
\draw[vmblue,line width=0.65pt] (3.080,1.934) -- (3.170,1.934);
\draw[vmblue,line width=0.65pt] (3.080,2.285) -- (3.170,2.285);
\draw[vmblue,line width=0.85pt,fill=white] (3.070,2.062) rectangle (3.180,2.172);
\draw[vmorange,line width=0.65pt] (1.610,1.855) -- (1.610,2.340);
\draw[vmorange,line width=0.65pt] (1.565,1.855) -- (1.655,1.855);
\draw[vmorange,line width=0.65pt] (1.565,2.340) -- (1.655,2.340);
\draw[vmorange,line width=0.85pt,fill=white] (1.555,2.023) rectangle (1.665,2.133);
\draw[vmorange,line width=0.65pt] (3.430,2.136) -- (3.430,2.451);
\draw[vmorange,line width=0.65pt] (3.385,2.136) -- (3.475,2.136);
\draw[vmorange,line width=0.65pt] (3.385,2.451) -- (3.475,2.451);
\draw[vmorange,line width=0.85pt,fill=white] (3.375,2.287) rectangle (3.485,2.397);
\draw[vmorange,line width=0.65pt] (5.250,1.373) -- (5.250,1.446);
\draw[vmorange,line width=0.65pt] (5.205,1.373) -- (5.295,1.373);
\draw[vmorange,line width=0.65pt] (5.205,1.446) -- (5.295,1.446);
\draw[vmorange,line width=0.85pt,fill=white] (5.195,1.366) rectangle (5.305,1.476);
\end{scope}
\begin{scope}[shift={(7.150,-4.800)}]
\node[font=\small\bfseries,anchor=west] at (0.7,3.75) {Good anchor};
\draw[black!12,line width=0.35pt] (0.760,0.746) -- (6.500,0.746);
\node[anchor=east] at (0.680,0.746) {0.5};
\draw[densely dashed,black!55,line width=0.35pt] (0.760,1.725) -- (6.500,1.725);
\node[anchor=east] at (0.680,1.725) {1};
\draw[black!12,line width=0.35pt] (0.760,2.705) -- (6.500,2.705);
\node[anchor=east] at (0.680,2.705) {1.5};
\draw[black!75,line width=0.55pt] (0.760,0.550) -- (0.760,3.630);
\draw[black!75,line width=0.55pt] (0.76,0.55) -- (6.50,0.55);
\node[anchor=north] at (1.380,0.490) {4};
\node[anchor=north] at (3.200,0.490) {16};
\node[anchor=north] at (5.020,0.490) {64};
\node[anchor=north] at (3.63,0.09) {Cohort size $s$};
\draw[vmblue,line width=0.65pt] (1.150,2.021) -- (1.150,3.532);
\draw[vmblue,line width=0.65pt] (1.105,2.021) -- (1.195,2.021);
\draw[vmblue,line width=0.65pt] (1.105,3.532) -- (1.195,3.532);
\filldraw[draw=vmblue,fill=vmblue] (1.150,2.729) circle (1.45pt);
\draw[vmblue,line width=0.65pt] (2.970,2.178) -- (2.970,2.178);
\draw[vmblue,line width=0.65pt] (2.925,2.178) -- (3.015,2.178);
\draw[vmblue,line width=0.65pt] (2.925,2.178) -- (3.015,2.178);
\filldraw[draw=vmblue,fill=vmblue] (2.970,2.178) circle (1.45pt);
\draw[vmorange,line width=0.65pt] (1.455,2.107) -- (1.455,3.113);
\draw[vmorange,line width=0.65pt] (1.410,2.107) -- (1.500,2.107);
\draw[vmorange,line width=0.65pt] (1.410,3.113) -- (1.500,3.113);
\filldraw[draw=vmorange,fill=vmorange] (1.455,2.537) circle (1.45pt);
\draw[vmorange,line width=0.65pt] (3.275,2.178) -- (3.275,2.319);
\draw[vmorange,line width=0.65pt] (3.230,2.178) -- (3.320,2.178);
\draw[vmorange,line width=0.65pt] (3.230,2.319) -- (3.320,2.319);
\filldraw[draw=vmorange,fill=vmorange] (3.275,2.257) circle (1.45pt);
\draw[vmorange,line width=0.65pt] (5.095,2.178) -- (5.095,2.178);
\draw[vmorange,line width=0.65pt] (5.050,2.178) -- (5.140,2.178);
\draw[vmorange,line width=0.65pt] (5.050,2.178) -- (5.140,2.178);
\filldraw[draw=vmorange,fill=vmorange] (5.095,2.178) circle (1.45pt);
\draw[vmblue,line width=0.65pt] (1.305,1.881) -- (1.305,2.644);
\draw[vmblue,line width=0.65pt] (1.260,1.881) -- (1.350,1.881);
\draw[vmblue,line width=0.65pt] (1.260,2.644) -- (1.350,2.644);
\draw[vmblue,line width=0.85pt,fill=white] (1.250,2.155) rectangle (1.360,2.265);
\draw[vmblue,line width=0.65pt] (3.125,1.930) -- (3.125,2.285);
\draw[vmblue,line width=0.65pt] (3.080,1.930) -- (3.170,1.930);
\draw[vmblue,line width=0.65pt] (3.080,2.285) -- (3.170,2.285);
\draw[vmblue,line width=0.85pt,fill=white] (3.070,2.106) rectangle (3.180,2.216);
\draw[vmorange,line width=0.65pt] (1.610,2.193) -- (1.610,2.725);
\draw[vmorange,line width=0.65pt] (1.565,2.193) -- (1.655,2.193);
\draw[vmorange,line width=0.65pt] (1.565,2.725) -- (1.655,2.725);
\draw[vmorange,line width=0.85pt,fill=white] (1.555,2.412) rectangle (1.665,2.522);
\draw[vmorange,line width=0.65pt] (3.430,2.049) -- (3.430,2.451);
\draw[vmorange,line width=0.65pt] (3.385,2.049) -- (3.475,2.049);
\draw[vmorange,line width=0.65pt] (3.385,2.451) -- (3.475,2.451);
\draw[vmorange,line width=0.85pt,fill=white] (3.375,2.255) rectangle (3.485,2.365);
\draw[vmorange,line width=0.65pt] (5.250,1.373) -- (5.250,1.446);
\draw[vmorange,line width=0.65pt] (5.205,1.373) -- (5.295,1.373);
\draw[vmorange,line width=0.65pt] (5.205,1.446) -- (5.295,1.446);
\draw[vmorange,line width=0.85pt,fill=white] (5.195,1.366) rectangle (5.305,1.476);
\end{scope}
\node[rotate=90] at (-0.40,-1.12) {VM-SVRS / SVRP rounds};
\node[anchor=west] at (0.75,4.28) {\textcolor{vmblue}{$\bullet$} $m=64,\bar\delta=10$};
\node[anchor=west] at (4.05,4.28) {\textcolor{vmblue}{$\square$} $m=64,\bar\delta=100$};
\node[anchor=west] at (7.20,4.28) {\textcolor{vmorange}{$\bullet$} $m=256,\bar\delta=10$};
\node[anchor=west] at (10.55,4.28) {\textcolor{vmorange}{$\square$} $m=256,\bar\delta=100$};
\end{tikzpicture}
\caption{Fresh-seed 4-seed tuning and 8-seed evaluation split of the VM study. Tuned VM-SVRS uses fewer strict rounds than tuned SVRP in 9 of 10 dip-family settings with incomplete participation and in 2 of 30 other settings. Points are paired geometric-mean ratios on evaluation seeds 4--11; bars are paired bootstrap 95\% intervals. Values below one favor VM-SVRS. Full-participation ties are omitted. Multipliers were selected on disjoint tuning seeds from the tested grid.}
\label{fig:vm-ratios}
\end{figure*}

%% file: gpu_study_results.tex
\section{Completed two-parameter GPU study}\label{app:gpu-study}

\paragraph{Prespecified instance grid and generator.}
The full fp64 study uses
\(m\in\{32,128\}\), \(d\in\{64,256\}\),
\(s\in\{1,4,16,m\}\), \(\delta_a,\bar\delta\in\{1,10,100\}\),
16 seeds per cell, \(\mu=1\), a nominal spectral scale
\(L/\mu=1000\), and at most 20,000 strict rounds per arm.
For each feasible \((m,d,\delta_a,\bar\delta,\text{seed})\), the
generator draws an orthogonal \(Q\), puts a structured perturbation in
a \(k=\max\{2,\operatorname{round}(d/8)\}\)-dimensional top eigenspace
\(U\), and sets
\[
 A_i=\bar A+US_iU^\top,\qquad
 \sum_i S_i=0,\qquad
 \|S_0\|_{\rm op}=\delta_a,\qquad
 \left\|\frac1m\sum_i S_i^2\right\|_{\rm op}^{1/2}=\bar\delta.
\]
The construction checks local \(\mu\)-strong convexity and sets
\(b_i=A_i x^\star-o_i\), where \(\sum_i o_i=0\), so
\(\nabla f(x^\star)=0\). The same instance seed is reused across
cohort sizes and arms. Because zero-sum perturbations require
\(\bar\delta\geq\delta_a/\sqrt{m-1}\), three of nine
\((\delta_a,\bar\delta)\) pairs are infeasible for \(m=32\), and
one is infeasible for \(m=128\). This leaves 28 feasible
\((m,d,\delta_a,\bar\delta)\) cells and 112 cell--cohort settings.

\paragraph{Metrics and comparators.}
Each method starts at \(x_0=0\). The diagnostic evaluator computes
the exact closed-form quadratic gap without changing any iterate.
We record the first charged round at which the gap divided by its
initial value is at most \(10^{-4}\), the number of client contacts,
and the final relative gap. The table stores a median first-hit round
only if all 16 seeds reach the target; otherwise it stores the hit
count and leaves that median undefined. There are 23 reported outputs,
including weighted S-DANE diagnostics, but weighted diagnostics are
excluded from the fastest-output comparison. Comparator multipliers
\(0.5\) and \(2\) are exploratory tuning; their wins do not establish
source-theorem guarantees. Exact local linear solves and matrix
factorizations are unpriced in the strict communication-round metric.
Direct-transcription checks verify the theorem arms, their
round/contact ledgers, and the principal baselines.

\begin{table}[t]
\centering\footnotesize
\caption{Cell-level ratio of median first-hit rounds for source-parameter
AccSVRS divided by theorem-parameter cohort AccSVRS. The table reports
the median of this ratio across 28 feasible cells, and the number of
cells in which the cohort arm uses fewer median rounds. \(s=m\) means
\(s=32\) or \(128\) as appropriate. All displayed arms hit in all
16 seeds of every cell.}
\label{tab:gpu-source-ratio}
\setlength{\tabcolsep}{5pt}
\begin{tabular}{lrr}
\toprule
Cohort size & Median ratio & Cohort faster\\
\midrule
\(s=1\)  & 0.845 & 4/28\\
\(s=4\)  & 2.818 & 25/28\\
\(s=16\) & 6.440 & 27/28\\
\(s=m\)  & 10.959 & 28/28\\
\bottomrule
\end{tabular}
\end{table}

\paragraph{Full grid and interpretation.}
Table~\ref{tab:gpu-full-grid} reports every feasible cell for the
theorem-parameter cohort AccSVRS arm. ``Spread'' is the largest divided
by the smallest value over available \(s\) of
\(\text{median rounds}/[E+\sqrt{E\lambda_s/\mu}]\), where
\(E=\lceil m/s\rceil\) and \(\lambda_s\) is~(14) with \(b=s\).
The ratio absorbs the common logarithmic accuracy factor within each
cell. Its median across the 28 cells is 2.25, with 21 cells at most 3
and a maximum of 11.48. Cohort SVRS has median spread 1.54, with 26
of 28 cells at most 2. All seven cells with cohort AccSVRS spread
above 3 have \(m=128\), and their largest observed-to-shape ratio
occurs at \(s=m\). These are descriptive finite-grid observations,
not estimates of theorem constants.

\begin{table*}[t]
\centering\scriptsize
\caption{Median strict rounds to relative gap \(10^{-4}\) for
theorem-parameter cohort AccSVRS over the full feasible grid.
``--'' denotes a cohort size absent for that \(m\); all listed
medians use 16 of 16 seeds. The final column is the observed-to-shape
spread defined in the text; all entries come from the saved summary.}
\label{tab:gpu-full-grid}
\setlength{\tabcolsep}{4pt}
\input{tables/scaling_1e-4.tex}
\end{table*}

The theorem-parameter cohort AccSVRS arm hits the target in all
\(112\times16\) trials. Cohort SVRS has 15 of 16 hits in each of two
settings at \(s=1\). Across all 112 settings, the smallest median
among complete-seed, non-weighted outputs belongs to BAPG 36 times,
APSVRG \(\times0.5\) 27 times, SVRP \(\times2\) 21 times, S-DANE
\(\times0.5\) 19 times, AGD eight times, and SVRP \(\times1\) once.
Neither theorem-parameter cohort arm has the smallest median. This
comparison is about strict rounds only; local computation and device
wall time have different meanings. Many grid cells also lie outside
the high-accuracy parameter restrictions of Corollary~11, so the
experiment does not validate or refute that lower bound.

\paragraph{Evidence available for uncertainty analysis.}
The saved aggregate results cover 16 completed units, 2,548 method
rows, and 16 seeds per row. The run took 3,993.9 seconds on two
RTX~4090 GPUs and allocated at most 17.3 GB of device memory. Per-seed
gap trajectories are retained separately, so the aggregate results
alone cannot support paired uncertainty intervals or a
trajectory-level audit.

%% file: tables/scaling_1e-4.tex
\begin{tabular}{rrrrrrrrrr}
\toprule
$m$ & $d$ & $\delta_a$ & $\bar\delta$ & $s=1$ & $s=4$ & $s=16$ & $s=32$ & $s=128$ & spread\\
\midrule
32 & 64 & 1 & 1 & 307 & 71 & 33 & 36 & -- & 2.46\\
32 & 64 & 1 & 10 & 1150 & 175 & 44.5 & 22 & -- & 1.61\\
32 & 64 & 1 & 100 & 2759 & 685 & 163.5 & 16 & -- & 1.89\\
32 & 64 & 10 & 10 & 1240.5 & 216 & 150 & 122 & -- & 2.22\\
32 & 64 & 10 & 100 & 2792.5 & 685 & 172.5 & 76 & -- & 1.14\\
32 & 64 & 100 & 100 & 3092 & 685 & 398 & 348 & -- & 1.78\\
32 & 256 & 1 & 1 & 214 & 60 & 31 & 32 & -- & 2.94\\
32 & 256 & 1 & 10 & 974 & 189 & 46 & 24 & -- & 1.26\\
32 & 256 & 1 & 100 & 2713 & 680 & 213 & 16 & -- & 2.46\\
32 & 256 & 10 & 10 & 974 & 189 & 108 & 110 & -- & 2.28\\
32 & 256 & 10 & 100 & 2713 & 651 & 206 & 80 & -- & 1.43\\
32 & 256 & 100 & 100 & 2560 & 632 & 435 & 350 & -- & 1.96\\
128 & 64 & 1 & 1 & 470 & 113 & 81 & -- & 28 & 4.21\\
128 & 64 & 1 & 10 & 1660 & 507 & 177 & -- & 20 & 1.74\\
128 & 64 & 1 & 100 & 5621.5 & 1389 & 461 & -- & 16 & 1.60\\
128 & 64 & 10 & 1 & 301 & 267 & 246.5 & -- & 124 & 11.48\\
128 & 64 & 10 & 10 & 1803 & 507 & 219 & -- & 110 & 3.07\\
128 & 64 & 10 & 100 & 6132.5 & 1465 & 461 & -- & 72 & 1.48\\
128 & 64 & 100 & 10 & 1660 & 785.5 & 730.5 & -- & 354 & 3.74\\
128 & 64 & 100 & 100 & 6806 & 1482 & 589.5 & -- & 354 & 2.52\\
128 & 256 & 1 & 1 & 486 & 93 & 33 & -- & 32 & 5.84\\
128 & 256 & 1 & 10 & 2862 & 496 & 159 & -- & 24 & 1.48\\
128 & 256 & 1 & 100 & 7713 & 1449 & 411 & -- & 16 & 1.63\\
128 & 256 & 10 & 1 & 486 & 206 & 159 & -- & 122 & 7.00\\
128 & 256 & 10 & 10 & 2862 & 496 & 159 & -- & 112 & 2.83\\
128 & 256 & 10 & 100 & 6821 & 1426 & 443 & -- & 76 & 1.61\\
128 & 256 & 100 & 10 & 1873 & 1006 & 721 & -- & 380 & 3.56\\
128 & 256 & 100 & 100 & 6821 & 1454.5 & 519 & -- & 336 & 2.43\\
\bottomrule
\end{tabular}

%% file: deferred_main.tex
\section{S-DANE accounting, numerical checks, and earlier
diagnostics}\label{app:deferred}

\paragraph{S-DANE transcription.} A literal S-DANE exchange occupies
three strict rounds under the strict interface. The selected \(s\)
clients first return gradients at the current point. They then return
local solutions after the cohort mean is broadcast, and finally return
gradients at those solutions. This transcription costs \(3s\) contacts
and \(6s\) vector transmissions per S-DANE iteration. It is feasible
accounting, not a lower bound on other implementations.
Table~\ref{tab:appendix-ledger} lists this row with the other ledgers.
S-DANE's subset-similarity and gradient-variance assumptions differ
from the anchor condition~(1).

\paragraph{S-DANE controls.} We ran source-faithful S-DANE
\citep[Algorithm~1]{jiang2024} under the strict ledger. The exact
quadratic reduction costs two rounds per iteration, since the mean
local point is the next center, and the literal transcription costs
three. The setup had 12 clients, dimension 6, cohorts \(\{1,3,12\}\),
5 seeds, and a 48-round cap: 60 paired cases and 635 runs. Its
parameter was the prespecified middle grid value, not a best-\(\lambda\)
choice. Table~\ref{tab:sdane} counts last iterates that reach relative
error \(10^{-4}\). The published weighted output reaches it in 18/45
(exact) and 16/45 (literal) quadratic cases. S-DANE never reaches the
target on the 30 partial-cohort quadratic cases and reaches it on all
15 full-cohort ones. These are finite-window observations under
assumptions S-DANE need not satisfy here (its subset similarity does not
follow from~(1)). They do not refute its theorems.

\begin{table}[h]
\centering\footnotesize
\caption{Last iterates reaching relative error \(10^{-4}\) within 48
strict rounds (5 seeds; 45 quadratic and 15 logistic cases).}
\label{tab:sdane}
\begin{tabular}{lrr}
\toprule
Method & Quadratic & Logistic\\
\midrule
BAPG & 36/45 & 10/15\\
Batched GD & 0/45 & 15/15\\
Strict SAGA & 0/45 & 10/15\\
Quadratic identification & 30/45 & n/a\\
S-DANE, exact two-round & 15/45 & n/a\\
S-DANE, literal three-round & 15/45 & 5/15\\
\bottomrule
\end{tabular}
\end{table}

\paragraph{Numerical checks of the theory.} Exact checks of the chain verify the geometric optimum,
the local-client split, support closure, the frontier process, and the
Chebyshev sandwich for representative finite instances. It also checks
the two-group instance of Lemma~C.7: the averaging identity, the anchor
deviation, local strong convexity, \(\bar\delta\leq\sqrt3\,\delta\),
closure of every reply type, and the gap of batched Acc-S-DANE on the
instance. Further checks cover the
finite-population identity, the parameter inequalities, the tight
constant of Lemma~C.1, Lemma~C.6 on chain instances, the case analysis
for \(\tau\), and the Theorem~6 bound of \citet{jiang2024} used in
Proposition~13. By Monte Carlo on quadratics with nonconvex clients and
exact or inexact anchor solves, it also checks both expected-potential
contractions and the expected round counts. These checks support but
do not replace the proofs.

\paragraph{Earlier diagnostics.} The variance enumeration has 20 cells
with 20,000 Monte Carlo draws per cell; each enumerated value equals the
finite-population identity. The charged optimization sweep has 1,926
checkpoints, with 30-seed quadratic and logistic sensitivity checks.
Appendix~\ref{computational-diagnostics-and-artifact-reconciliation}
contains the figures, uncertainty, and historical-oracle audit. The
30-seed calibration shows that a valid but loose public \(\mu\) makes
\(\delta/\mu\) plots poor tests of a theoretical constant. These
experiments check accounting and failure regimes, not a lower rate.

\paragraph{Sampling noise beyond Theorem~3.} Theorem~3 uses gradient
\emph{differences}, so client-specific linear terms cancel. For a fresh
sampled gradient at one point, the finite-population identity applies
instead to \(u_i=\nabla f_i(w)\), and~(1) does not control the resulting
gradient-offset variance. In the hidden-client family of Theorem~2 that
variance is exactly \((m-s)\mu^2\theta^2/s\), even at the optimum and
with \(\delta=0\). These witnesses rule out two shortcuts in a sampled
analysis; neither rules out a method that learns client deviations or
uses the exact local-prox oracle.

%% file: diagnostic_details.tex
\section{Computational diagnostics and historical
evidence}\label{computational-diagnostics-and-artifact-reconciliation}

\subsection{Exact variance checks}\label{exact-variance-checks}

We evaluated (8) at \(m=12,\mu=1,x-w=2\) with uniform cohorts
\(s\in\{1,2,4,8,11\}\). The zero-anchor grid crosses these five cohorts
with \(M\in\{1,3,9\}\), giving 15 cells. A second five-cell grid fixes
\(\delta=1,M=3\), so the anchor gap is \emph{positive and exact}. Each
cell has 20,000 Monte Carlo cohorts and an exact enumeration of all
subsets. Every enumerated second moment equals (8), and every Monte
Carlo estimate is within four reported standard errors of its exact
value; all five positive-\(\delta\) estimates are within two. The three
zero-\(\delta\) scales restart seed 17, so their draws are paired
rescalings rather than independent replications.

The finite-population identity predicts a precise participation
effect: for fixed client Hessians, increasing \(s\) reduces variance
by \((m-s)/[s(m-1)]\), while at fixed incomplete \(s\) changing
\(M\) changes the squared population deviation. The Monte Carlo
grid checks both axes. Its standard errors are calculated across
sampled cohorts, and exhaustive enumeration supplies the exact
benchmark. Agreement tests the code path for uniform sampling and
variance measurement; it does not test an optimization rate. The
positive-\(\delta\) cells matter because a zero-gap example could
otherwise appear to be a degenerate edge case, whereas the same
obstruction persists with an exactly prescribed nonzero anchor gap.

{\def\LTcaptype{none} 
\begin{table*}[t]
\centering\small
\caption{Exact and Monte Carlo sampled-difference variance at selected cohort sizes (20,000 draws per estimate).}
\label{tab:variance}
\begin{tabular}{rrrrr}
\toprule
Anchor $\delta$ & $M$ & $s$ & Exact variance & Monte Carlo mean (SE) \\
\midrule
0 & 3 & 1 & 6.0000 & 5.9814 (0.0948) \\
0 & 3 & 11 & 0.04959 & 0.04909 (0.00078) \\
1 & 3 & 1 & 4.6667 & 4.6468 (0.07365) \\
1 & 3 & 11 & 0.03857 & 0.03800 (0.00060) \\
\bottomrule
\end{tabular}
\end{table*}
}

\begin{figure}
\centering
\includegraphics[width=0.7\linewidth,height=\textheight,keepaspectratio,alt={Finite-population
gradient-difference variance at zero anchor dissimilarity. Lines are
exact values; points show 20,000-cohort means with unadjusted
1.96-standard-error Monte Carlo intervals. The exact
positive-dissimilarity check is reported in the
table.}]{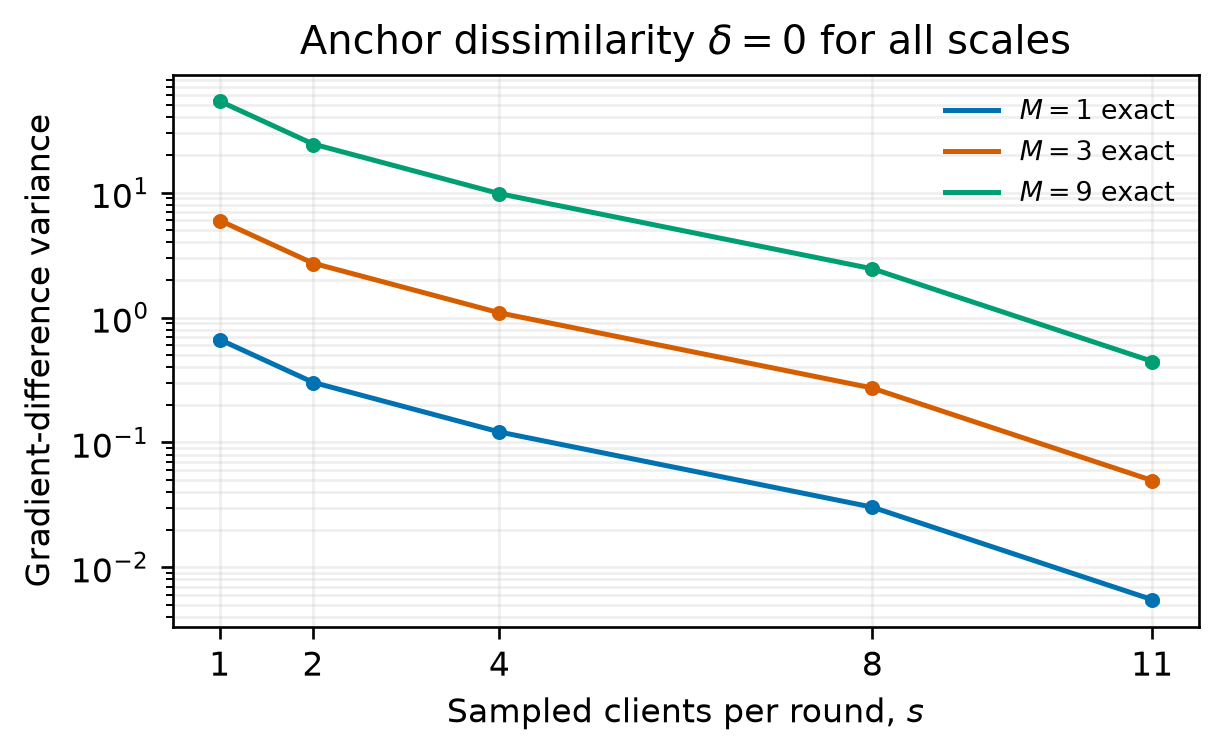}
\caption{Finite-population gradient-difference variance at zero anchor
dissimilarity. Lines are exact values; points show 20,000-cohort means
with unadjusted 1.96-standard-error Monte Carlo intervals. The exact
positive-dissimilarity check is reported in the
table.}\label{fig:variance}
\end{figure}

These cells check the finite-population calculation; they are not
empirical optimization lower bounds. At fixed \(s=1,\delta=0\), the
exact variance increases from \(2/3\) to \(6\) to \(54\) as \(M\) grows
from 1 to 3 to 9.

\subsection{Charged optimization and curvature
calibration}\label{charged-optimization-and-curvature-calibration}

The initial strict-oracle CPU sweep (seed 11, 96-round cap) saved 1,926
checkpoints across \(m=24,d=3\) quadratics and one \(m=12,d=3\)
regularized logistic instance. Every trace logs rounds, contacts, and
vectors. Batched GD receives a certified global smoothness bound; strict
SAGA receives a certified maximum local bound. These are oracle-aided
tuning inputs. SAGA's table initialization is charged. The diagnostic
objective evaluator computes gaps only for plots and never changes an
iterate or stopping rule. The seed-11 quadratic proximal steps use
floating-point linear solves; their logged zero model-gap bound treats
these well-conditioned solves as exact in the ideal oracle, rather than
certifying a general numerical residual. The seed-11 quadratics use a public
\(\mu=0.5\), while their \textbf{realized global minimum eigenvalue is
10.5} at every scale. At scales 0.4 and 6, their minimum client
eigenvalues are respectively 10.22 and 5.82. Therefore their observed
speed must not be read as a tight test of a \(\delta/0.5\) theorem
constant.

We repeated the quadratic design on seeds 100--129, pairing each seed's
family across scales 0.4 and 6, cohort sizes \(s=1,4\), and methods. The
target is a final saved gap at most \(10^{-4}\Delta_0\) under the same
96-round cap. The table calibrates the proof factor with the
\emph{largest valid common local strong-convexity constant} for each
realized instance. Its medians are over 30 seeds; recalibration changes
no trajectory.

{\def\LTcaptype{none} 
\begin{table*}[t]
\centering\small
\caption{Curvature calibration across 30 paired quadratic seeds; calibrated $\mu$ is the minimum local eigenvalue.}
\label{tab:curvature}
\begin{adjustbox}{max width=\textwidth}
\begin{tabular}{rrrrr}
\toprule
Scale & Median $\delta$ & Median min-client $\mu$ & Median $\delta/\mu$: public / calibrated & Median BAPG $q$: public / calibrated \\
\midrule
0.4 & 0.231 & 10.23 & 0.463 / 0.0226 & 0.481 / 0.0433 \\
6 & 3.471 & 5.63 & 6.942 / 0.622 & 0.933 / 0.554 \\
\bottomrule
\end{tabular}
\end{adjustbox}
\end{table*}
}

The calibrated number uses the minimum eigenvalue over \emph{all} local
Hessians; it is valid for the proof's clientwise assumption. The global
minimum remains 10.5. This sensitivity is large enough that the original
sweep cannot rank methods by the theorem's displayed \(\delta/\mu\)
constants.

The calibration illustrates a gap between an admissible proof input
and the geometry of a realized instance. At scale 0.4, the median
anchor gap is 0.231; the public \(\mu=0.5\) gives median ratio
\(\delta/\mu=0.463\), whereas the minimum local eigenvalue gives
0.0226. With \(\lambda=\delta\), the corresponding median displayed
BAPG factors are 0.481 and 0.0433. At scale 6, the same calculation
changes 0.933 to 0.554. These evaluate the proved expression on two
valid inputs for the same saved families; they change no trajectory
and do not supply the algorithm with free knowledge of the sharper
\(\mu\). The 30 paired seeds measure the frequency of this looseness
in the chosen generator, not a distribution-free improvement to (4).

The final-checkpoint success counts, out of 30 runs per cell, are

\[\begin{array}{cc|cccc}
\text{Scale} & s & \text{BAPG} & \text{GD} & \text{SAGA} & \text{Quadratic ID}\\
\hline
0.4 & 1 & 30/30 & 30/30 & 1/30 & 30/30\\
0.4 & 4 & 30/30 & 30/30 & 30/30 & 30/30\\
6 & 1 & 5/30 & 30/30 & 0/30 & 30/30\\
6 & 4 & 30/30 & 30/30 & 30/30 & 30/30
\end{array}\]

No cell is credited for an earlier transient crossing. For \(s=1\), BAPG
ends at round/contact 75/75, GD and SAGA at 96/96, and identification at
96/96. For \(s=4\), these are 91/325, 96/384, and 24/96 respectively.
Every contact transmits one query and one reply. Identification is a
class-specific negative control, not a general optimizer. The
large-scale \(s=1\) row shows that BAPG's theorem does not imply
empirical dominance over a full-gradient comparator. At this scale and
\(s=1\), the 95\% Wilson intervals for the seed success proportions are
\([0.073,0.336]\) for BAPG and \([0,0.114]\) for SAGA; for a 30/30 cell
the interval is \([0.886,1]\). These intervals describe this synthetic
seed distribution, not uncertainty about a minimax rate.

\begin{figure*}[t]
\centering
\includegraphics[width=0.8\textwidth,height=\textheight,keepaspectratio,alt={Across
30 paired quadratic seeds at s=1, fraction whose current iterate at each
round budget has relative gap at most 10\^{}\{-4\}. The latest saved
iterate is carried forward between method steps. Exact quadratic
identification solves at round 96; curves do not imply a
general-function lower bound.}]{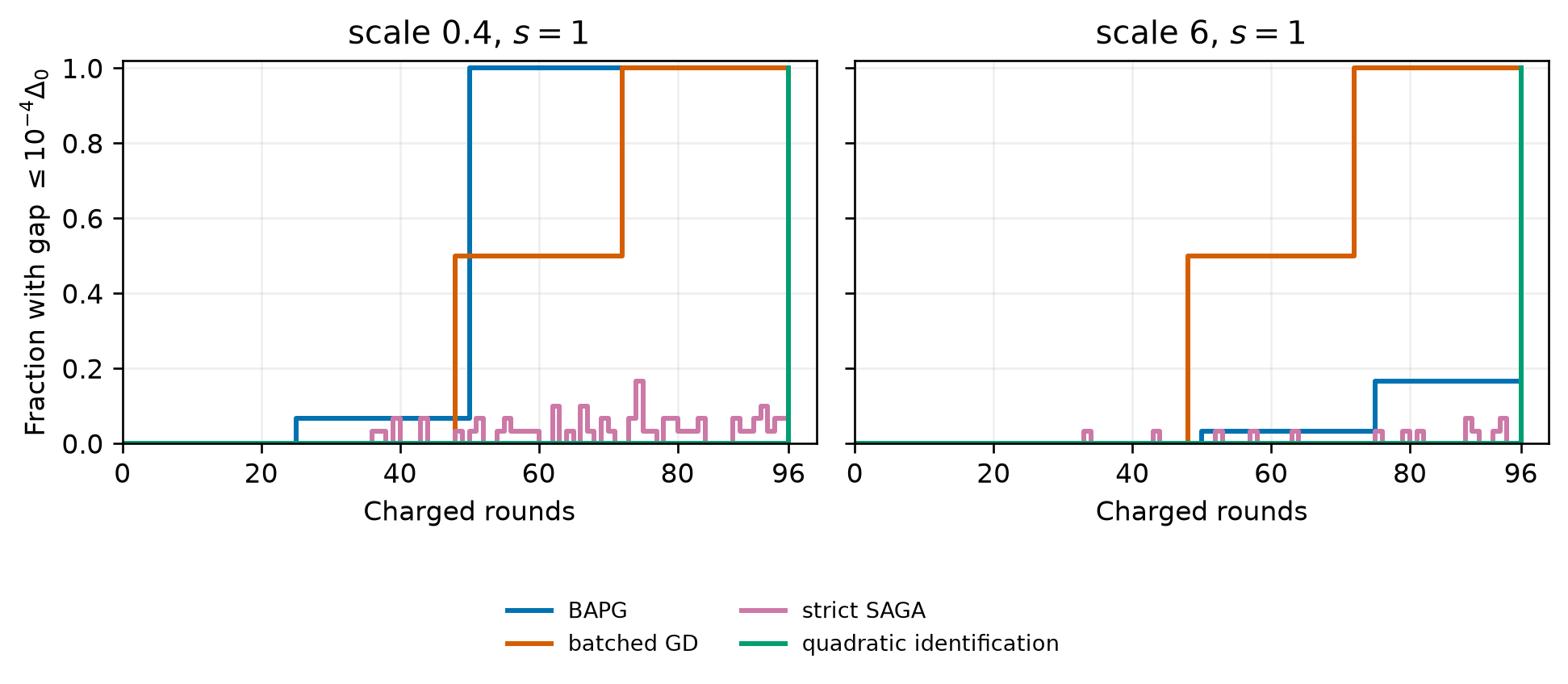}
\caption{Across 30 paired quadratic seeds at \(s=1\), fraction whose
\emph{current} iterate at each round budget has relative gap at most
\(10^{-4}\). The latest saved iterate is carried forward between method
steps. Exact quadratic identification solves at round 96; curves do not
imply a general-function lower bound.}\label{fig:replicates}
\end{figure*}

A separate 30-seed regularized-logistic check uses \(m=12,n=30\)
observations per client, \(d=3,s=2\), and the same round cap. The
regularization \(\mu=0.5\) is a sharp uniform lower curvature bound for
this logistic family, while the anchor Hessian-dissimilarity \(\delta\)
is a conservative analytic certificate for client zero (median 0.647).

To obtain that certificate, write \(G_i=X_i^\top X_i/n\). The Hessian
of client \(i\) is \(\mu I+X_i^\top W_i(x)X_i/n\), where each
diagonal logistic weight lies in \([0,1/4]\). The selected anchor
therefore satisfies the uniform, data-dependent bound
\[
\delta_{\mathrm{cert}}=\tfrac14
\left(\|G_0\|_{\mathrm{op}}+
\left\|m^{-1}\sum_iG_i\right\|_{\mathrm{op}}\right).
\tag{11}\]
It bounds the anchor-to-global Hessian gap for every \(x\), but
usually overestimates the exact supremum because it discards
cancellation and the actual weights. The certified \(\mu\) and
\(\delta_{\mathrm{cert}}\) are extra tuning inputs in this diagnostic,
not quantities learned through the strict one-vector optimization
transcript. Batched GD separately receives a global smoothness
certificate, and SAGA a maximum-local certificate; conflating the
two would misstate the comparison.
All three methods reach the \(10^{-4}\) endpoint target on all 30 seeds.
Median final relative gaps are \(3.34\times10^{-9}\) for BAPG (91
rounds, 169 contacts), numerical zero for batched GD (96 rounds, 192
contacts), and \(9.91\times10^{-11}\) for strict SAGA (96 rounds, 192
contacts). Their empirical interquartile ranges (Hazen quartiles) are
\([1.59\times10^{-9},1.74\times10^{-8}]\), zero at reported precision,
and \([1.99\times10^{-11},5.87\times10^{-10}]\), respectively. This
supports only a small nonlinear sensitivity check; objective differences
at numerical zero are below reliable ranking precision.

At \(\delta=0\), the seed-11 one-shot affine control has gap
\(1.56\times10^{-10}\) from \(7.3422\) after 25, 7, or 4 rounds for
\(s=1,4,8\), with 25 contacts in each case. Quadratic identification
reaches numerical zero after 96 contacts at \(s=1,4,8\), respectively
96, 24, and 12 rounds. The fixed-seed round and contact plots, and full
checkpoint data, are retained in the companion artifacts for audit.

\paragraph{Interpreting the charged diagnostics.}
The quadratic and logistic studies answer an accounting question,
not the same question as Theorem 3. The variance panel samples fresh
cohorts at fixed \(x,w\) and compares their second moment with an
exact finite-population value. The optimization panels run specific
algorithms under a round cap, record every query and reply, and
evaluate objective gap only for diagnosis. A success count of 30/30
describes the chosen generator and tuning inputs. It does not rule
out a harder family with the same \(\mu,\delta\), and a 0/30 cell
does not prove a lower bound for all strict-oracle methods.

Pairing the same 30 quadratic seeds across heterogeneity scales and
cohort sizes helps isolate changes in a method's charged trajectory;
the resulting cells are correlated. The Wilson intervals reported
above quantify seed variation within one cell, not uncertainty in
the theorem factor or a confidence band for the best optimizer.
The logistic check changes objective class and uses an analytic
curvature certificate, so it is a sensitivity control rather than
a direct replication of the quadratic curves. Keeping these roles
separate matters because exact quadratic identification can make a
difficult-looking rate plot trivial at fixed \(d\), while the
nonlinear logistic instance has no such finite-probe identity.

\subsection{Historical evidence under a different oracle}
\label{historical-run-inventory-and-exclusion-from-strict-oracle-evidence}

The original project saved quick CPU, full CPU, and large GPU
simulations and six figure families: rounds versus dissimilarity,
convergence, rounds versus participation, sampled floor, a quadratic
zero-chain, and logistic rounds versus dissimilarity. Their original
files remain intact; the companion experiment inventory indexes every
family and its data. The appendix retains the descriptive slopes
and their oracle audit; they are not strict-oracle rate estimates.

In the historical sliding code, each nominal iteration uses an
\textbf{exact full gradient} and an \textbf{uncharged anchor prox}; it
is not one strict round. Moreover, its control-variate participation run
sets the reference point equal to the current point, making the sampled
difference zero and leaving an exact full gradient. Its reported
control-variate iteration counts are consequently independent of \(s\)
(35.0 in full CPU and 42.36 in large GPU). The uncorrected sampled floor
is an observation about that synthetic generator, not a universal floor
theorem. The zero-chain curve is the best among a few named methods on
finite-dimensional quadratics, not a lower bound over all methods;
Proposition 5 gives a finite exact-identification counterpoint. The
historical logistic slopes retain the same full-gradient/uncharged-prox
convention. None of those plots supports a positive-\(\delta\)
strict-oracle minimax claim.